\documentclass{article}

\usepackage{iclr2027_conference,times}

\usepackage[utf8]{inputenc} 
\usepackage[T1]{fontenc}    
\usepackage{hyperref}       
\usepackage{url}            
\usepackage{booktabs}       
\usepackage{amsfonts}       
\usepackage{nicefrac}       
\usepackage{microtype}      
\usepackage{xcolor}         
\usepackage{microtype}
\usepackage{graphicx}
\usepackage{subcaption}
\usepackage{booktabs} 

\usepackage{hyperref}

\usepackage{amsmath}
\usepackage{amssymb}
\usepackage{mathtools}
\usepackage{amsthm}
\usepackage{tikz}
\usetikzlibrary{tikzmark,arrows.meta}
\usepackage{adjustbox}
\usepackage{graphicx}
\usepackage{multirow}
\usepackage{makecell}
\usepackage[all]{nowidow}
\usepackage[capitalize,noabbrev]{cleveref}
\usepackage{wrapfig}

\theoremstyle{plain}

\theoremstyle{definition}

\theoremstyle{remark}

\usepackage[textsize=tiny]{todonotes}

\definecolor{nicegreen}{rgb}{0.4, 0.615, 0.647}
\definecolor{niceyellow}{rgb}{0.95, 0.62, 0.3}
\definecolor{nicepurple}{rgb}{0.85, 0.5, 0.62}
\definecolor{myblue}{rgb}{0.298, 0.5, 0.9}
\definecolor{myred}{rgb}{1, 0.5, 0.5}
\definecolor{mygreen}{rgb}{0.298, 0.65, 0.298}

\newcommand{\add}[1]{{#1}}

\newcommand{\myparagraph}[1]{\vspace{0pt} \noindent \textbf{#1} \hspace{3pt}}

\usepackage[most]{tcolorbox}

\usepackage{tcolorbox}
\tcbuselibrary{listingsutf8}
\usepackage{listings}
\usepackage{xcolor}

\definecolor{codegray}{gray}{0.95}

\newtcblisting{pythonbox}{
  colback=codegray,
  colframe=black,
  listing only,
  listing options={
    language=Python,
    basicstyle=\ttfamily\scriptsize,
    keywordstyle=\color{blue},
    commentstyle=\color{gray},
    stringstyle=\color{red},
    showstringspaces=false,
    breaklines=true
  },
  boxsep=5pt,
  arc=4pt,
  outer arc=4pt,
  left=5pt,
  right=5pt,
  top=5pt,
  bottom=5pt
}

\tcbset{
  mygreybox/.style={
    colback=gray!10,
    colframe=gray!80,
    boxrule=0.5pt,
    arc=2pt,
    outer arc=2pt,
    boxsep=5pt,
    left=5pt,
    right=5pt,
    top=5pt,
    bottom=5pt,
    enhanced,
    breakable,
  }
}

\definecolor{myBlue}{RGB}{76,81,223}
\definecolor{niceRed}{RGB}{220,48,150}

\preprintcopy       

\title{
The Dual Mechanisms of Spatial Variable\\Binding in Vision–Language Models
}

\ifpreprint
\author{%
Kelly Cui$^{1}$\thanks{Indicates equal contribution. Address correspondence to: \texttt{prakash.nik@northeastern.edu, tamarott@mit.edu}} \quad Nikhil Prakash$^{2}$\footnotemark[1] \quad Shoval Messica$^{}$\footnotemark[1]\\ 
\textbf{Ayush Raina}$^3$
\quad \textbf{David Bau}$^2$ 
\quad \textbf{Antonio Torralba}$^1$ 
\quad \textbf{Tamar Rott Shaham}$^1$ 
\\
$^1$MIT CSAIL \quad $^2$Northeastern University \quad $^3$Sony Playstation\\
}
\else
  \author{Anonymous authors\\Paper under double-blind review}
\fi

\begin{document}
\maketitle
\begin{abstract}

Many multimodal tasks, such as image captioning and visual question answering, require vision–language models (VLMs) to bind objects with their properties
and spatial relations. Yet it remains unclear where and how such associations are computed within VLMs. In this work, we show that VLMs rely on two concurrent mechanisms to represent spatial variable binding. In the language model backbone, intermediate layers represent content-independent spatial relations on top of visual tokens corresponding to objects. However, this mechanism plays only a secondary role in shaping model predictions. Instead, the dominant source of spatial information originates in the vision encoder, whose representations encode the layout of objects and are directly exploited by the language model backbone. Notably, this spatial signal is distributed globally across visual tokens, extending beyond object regions into surrounding background areas. We validate the generalization of our findings to complex natural images from the COCO dataset, where globally amplifying the vision-derived spatial representations across all image tokens corrects spatial variable binding failures across models of various sizes. Together, our results clarify how spatial variable binding is computed within VLMs and highlight the central role of vision encoders in enabling it.

\end{abstract}

\section{Introduction}

\add{Spatial variable binding}, the ability to associate objects with their properties and their relations to other elements, is a fundamental component of 
multimodal reasoning~\citep{cheng2024spatialrgpt, chen2021geoqa, zheng2025multimodal, chen2024spatialvlm, goetting2024end, wen2025zero}. Tasks such as image captioning, visual question answering, and visual navigation require vision–language models (VLMs) to bind objects to spatial relations and reason about their relative arrangement within the scene.
Although recent advances have made VLMs highly capable across many tasks, \add{spatial variable binding} remains a significant challenge~\citep{campbell2024understanding, chen2025spatial, wen2024transformers,kamath2023whatsup}. 
In this work, we investigate the internal mechanisms underlying \add{spatial variable binding} in VLMs and show that the mechanisms identified in controlled settings also account for model behavior \add{in complex natural scenes}.

In language models (LMs), recent studies have shown that associating entities with their attributes (often referred to as variable binding) is done by forming symbolic representations for each entity in context. This representation encodes content-independent information about the order of entities in context, and allows the model to distinguish between entities and retrieve entity-specific information when needed~\citep{dai-etal-2024-representational, prakash2024fine, prakash2025languagemodelsuselookbacks}. 
Follow-up studies suggest that VLMs similarly rely on ordering information when reasoning about objects in images~\citep{assouel2025visual, kang2026linear}. However, while these findings establish the use of ordering-based representations during multimodal reasoning, they do not reveal their origin. In particular, it remains unclear whether such representations are constructed within the LM backbone, inherited from the vision encoder, or emerge from interactions between the two. Identifying the source of these representations is essential for understanding spatial variable binding failures in VLMs and potentially developing principled methods to diagnose and correct them.



We show that the ordering-based representations underlying spatial variable binding in VLMs arise from two concurrent sources: The vision encoder represents the global layout of objects in the image, encoding ordering information that is directly projected into the embedding space of the LM backbone. Then, the LM backbone can further augment these representations by forming ordering information over object-associated visual tokens. While the vision encoder provides the primary source of ordering information, the LM-side mechanism plays a secondary role, augmenting spatial ordering when that information in the vision embeddings is degraded or completely removed.
Notably, the ordering information provided by the vision encoder is distributed globally across visual tokens, extending beyond object regions into surrounding background areas. This is in contrast to the LM backbone, where information is formed locally over object-related tokens, similar to the process found in language processing~\citep{prakash2025languagemodelsuselookbacks, assouel2025visual}. 


We establish these results through a series of controlled interchange intervention experiments~\citep{vig2020investigating, meng2022locating, geiger2022inducing} using carefully constructed counterfactual samples across three synthetic datasets and one controlled naturalistic dataset~\citep{kamath2023whatsup}.
We validate our findings across five transformer-based VLMs from three model families (Qwen, Gemma, and Pixtral), showing that they rely on similar mechanisms to solve spatial variable binding.
Finally, we demonstrate that this mechanism extends to natural images: a simple global intervention that amplifies vision-derived ordering information across all image tokens corrects up to $55\%$ of previously incorrect predictions on complex, in-the-wild COCO scenes with diverse objects and spatial relations~\citep{lin2014microsoft,kamath2023whatsup} (see Fig.~\ref{fig:app:coco}).


Overall, our results elucidate the mechanisms underlying spatial association in VLMs, connect binding mechanisms identified in LMs to multimodal reasoning, and demonstrate the central role of vision encoders in supporting spatial variable binding in VLMs.
\ifpreprint
All code and data supporting this study are available at \url{https://spatial.baulab.info}.
\else
All code and data supporting this study will be made publicly available upon publication.
\fi

\section{Related Works}

\begin{figure}[t]
    \centering
    \setlength{\fboxsep}{0pt}      
    \setlength{\fboxrule}{1.5pt}   

    \begin{tabular}{ccccc}
        Squares & Shapes & Objects & What`sUp & COCO-spatial\\
        \fcolorbox{gray}{white}{\includegraphics[width=0.15\linewidth]{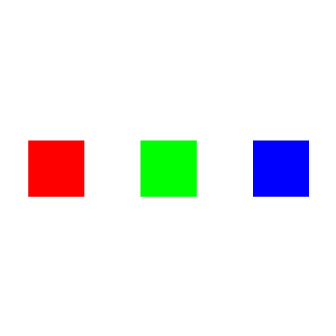}} &
        \fcolorbox{gray}{white}{\includegraphics[width=0.15\linewidth]{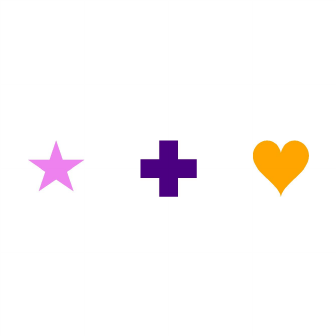}} & 
        \fcolorbox{gray}{white}{\includegraphics[width=0.15\linewidth]{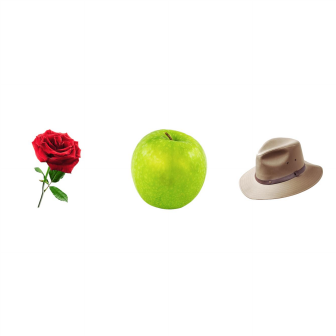}} &
        \fcolorbox{gray}{white}{\includegraphics[width=0.15\linewidth, height=.15\linewidth, trim=25 0 25 0,
    clip]{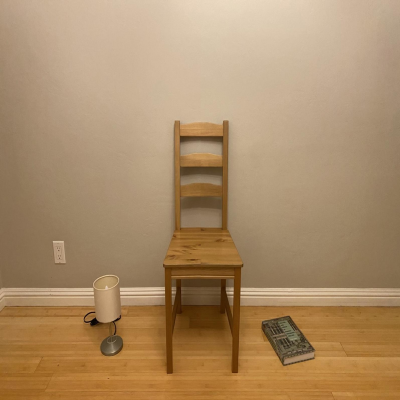}}
    &
    \fcolorbox{gray}{white}{\includegraphics[width=0.15\linewidth, height=.15\linewidth, trim=25 0 25 0,
    clip]{figures_main/coco.jpg}}
    \end{tabular}
    
    \vspace{0.5em}
    \small
    \textit{``The \texttt{<property>} of the \texttt{<object type>} to the \texttt{<direction>} of \texttt{<reference object>} is \textunderscore\textunderscore\textunderscore\textunderscore\textunderscore\textunderscore''}
    \caption{\textbf{Experimental Settings.} We study the internal mechanism responsible for spatial variable binding across three synthetic settings (Squares, Shapes, Objects) and one natural controlled settings (What'sUp). 
    We further show that our findings generalize to complex natural scenes from the COCO dataset with complex scenes and objects, where a simple intervention corrects spatial binding failures.}
    \label{fig:data}
    \vspace{-3mm}
\end{figure}

\myparagraph{Benchmarking Spatial Reasoning} Many benchmarks have been proposed to evaluate spatial reasoning in VLMs, including synthetic datasets designed to test compositional relations and naturalistic benchmarks derived from real images and human annotations~\citep{kamath2023whatsup,chen2024spatialvlm,gholami2025spatial,ma20253dsrbench}. These benchmarks have revealed that spatial reasoning remains challenging for current VLMs, particularly as relational complexity increases or when reasoning must generalize beyond object-centric cues. 

\myparagraph{Understanding Inner Working of VLMs}
A growing body of work has investigated the internal workings of VLMs, examining cross-modal attention patterns, token alignment, representational geometry, and information flow between vision encoders and LM backbones~\citep{nikankin2025same,qin2025vision,assouel2025visual,jiang2025vision,neo2025interpreting, jiang2024interpreting, Kaduri_2025_CVPR, shaham2024multimodal, cohen2026performancegapentityknowledge}. These studies have provided valuable insights into how visual and textual information is integrated and how multimodal representations emerge. Our work builds on this line of research by focusing specifically on the mechanisms underlying spatial association and reasoning. 

\myparagraph{Symbolic Representation in LMs  \& VLMs}
Several studies have shown that LMs implement variable binding by forming content-independent ordering identifiers, over important tokens \citep{gur2025mixing, prakash2025languagemodelsuselookbacks, dai-etal-2024-representational, prakash2024fine, feng2023language, davies2023discovering}. These mechanisms underlying variable binding are fundamental to many in-context reasoning tasks and have been presented as evidence of symbol-like processing in neural networks. 
Recent work has extended this line of inquiry to VLMs, investigating similar symbolic representations \citep{kang2026linear, assouel2025visual, saravanan2025investigating, hasani2025uncovering}. Our findings relate to this body of work while revealing an important distinction in the multimodal setting. We show that although VLMs form symbolic representations that encode ordering information within the LM backbone, they only act as a supporting mechanism. Instead, symbolic representations originating from the vision encoder play a dominant role.
\section{Experimental Setup}

\subsection{Data}
\label{subsec:data}

We study spatial variable binding across increasing visual complexity, from synthetic and controlled settings to complex natural images (Fig.~\ref{fig:data}). The controlled settings are necessary to enable causal analysis, which requires constructing original-counterfactual input pairs that differ in exactly one factor.
The naturalistic settings then provide evidence that the findings apply to real-world conditions.

\myparagraph{Synthetic Settings}
We consider three synthetic settings: \textit{Squares}, \textit{Shapes}, and \textit{Objects}. In all settings, three equal-sized items are placed at fixed, equal distances and arranged either horizontally or vertically. The settings differ only in visual content: \textit{Squares} uses colored squares, \textit{Shapes} uses colored geometric shapes, and \textit{Objects} uses real-world object images. All images are generated programmatically from predefined collections of colors, shapes, and objects (see full list at App.~\ref{app:data}), enabling precise control over attributes and spatial configuration. For each image, we query the VLM to predict an attribute (color, shape, or object name) of a target entity (square, shape, or object) based on its relative position to a reference entity. For our intervention experiments (see Sec.~\ref{sec:experiments}), we sample 100 clean--counterfactual image pairs from each setting and report 95\% bootstrap confidence intervals; we validate the statistical reliability of this sample size in App.~\ref{app:n100_stats}. Importantly, we further validate that our results generalize to images with real-world backgrounds (App.~\ref{app:probe_real_back}) and to diagonal spatial arrangements (App.~\ref{app:diagonal}).

\myparagraph{Naturalistic Setting}
For mechanistic analysis,
we use the control subset of What’sUp~\citep{kamath2023whatsup} 
which consists of images containing pairs of objects: a central reference object (e.g., a chair) and a secondary object positioned adjacent to it on either side. To adapt this dataset to our intervention experiments (Sec.~\ref{sec:experiments}), which require images containing three objects, we construct composite scenes by merging pairs of What’sUp images into a single image with a shared reference object and two surrounding objects, resulting in 1074 images. This construction supports only horizontal spatial relations.
In Sec.~\ref{sec:fixing}, we evaluate generalization to natural images from COCO-spatial~\citep{kamath2023whatsup,lin2014microsoft}, curated by filtering the COCO validation set for images that exhibit spatial relations. This results in 2687 images with complex scenes containing multiple objects, varied spatial arrangements, and realistic backgrounds. We also report generalization of these results to the on/under and front/behind relations using the original What'sUp images.







\subsection{Models}
\label{subsec:models}
The mechanistic study is performed
with five transformer-based VLMs from three model families: \texttt{Qwen2-VL-7B-Instruct}, \texttt{Qwen2-VL-2B-Instruct}, \texttt{Qwen3-VL-8B-Instruct}, \texttt{Gemma-3-4b-it}, and \texttt{Pixtral-12B}. These models differ in scale, training data, and vision-encoder design (in particular, the positional encoding of the vision encoder; see App.~\ref{app:strip_emergence}), allowing us to assess whether the identified mechanisms generalize across diverse VLM designs. The behavioral performance of the models on the spatial variable binding tasks is summarized in App.~\ref{app:models_accuracy} (Table~\ref{tab:behavioral_all_models}). In the main text we present results for \texttt{Qwen2-VL-7B-Instruct}; results for all other models are reported in App.~\ref{app:last_token} and App.~\ref{app:criss-cross}. For causal experiments, we use examples where the model answers correctly, ensuring the identified mechanisms reflect genuine behavior rather than noise. Nevertheless, we show generalization of the findings to failure cases in Sec.~\ref{sec:fixing}.


\subsection{Methods}
\label{subsec:tools_and_methods}

\paragraph{Representation Probing}
\looseness=-1 
A probe~\citep{belinkov2022probing} is typically a lightweight classifier trained to predict a target property $y$ from an internal representation $\mathbf{h} \in \mathbb{R}^d$ extracted from the model. In its simplest form, a linear probe takes the form $\hat{y} = \mathbf{W}\mathbf{h} + \mathbf{b}$, where $\mathbf{W}$ and $\mathbf{b}$ are learned parameters. Successful probing that generalizes to test cases indicates that the target information is linearly decodable from $\mathbf{h}$, suggesting that it is present in the model's representation.

\myparagraph{Causal Mediation Analysis}
Interchange intervention is a technique for testing causal relationships between a model’s internal representations and its behavior~\citep{vig2020investigating, meng2022locating}. Let $f(\cdot)$ denote the model, and let $\mathbf{h}_\ell(x)$ denote the internal representation at component or layer $\ell$ produced during a forward pass on input $x$. Given an \textit{original} input $x$ and a corresponding \textit{counterfactual} input $x'$, an interchange intervention replaces $\mathbf{h}_\ell(x)$ with $\mathbf{h}_\ell(x')$ while keeping the remainder of the computation unchanged (\emph{i.e.}, \textit{activation patching}). This yields an intervened output $\hat{y}_{\text{int}} = f\big(x;\, \mathbf{h}_\ell \leftarrow \mathbf{h}_\ell(x')\big)$. If the intervened output $\hat{y}_{\text{int}}$ matches the target outcome associated with the counterfactual input, this provides evidence that the intervened representation $\mathbf{h}_\ell$ plays a causal role in the model’s computation. We quantify this effect using \textit{interchange intervention accuracy} (IIA)~\citep{geiger2022inducing}. Rather than measuring binary agreement with the intervention target as in the original IIA, we average probability assigned to the target intervention outcome token.

\section{Preliminaries}

\subsection{Ordering Information in LMs}
\label{sec:llm_var_binding}
Variable binding is the process of associating attributes with their respective entities. For instance, given a red, green, and blue square, the process of assigning the color feature to each square involves variable binding. 
Recent works have investigated the underlying mechanisms that enable variable binding in LMs~\citep{prakash2025languagemodelsuselookbacks, dai-etal-2024-representational}. A common finding 
is the reliance of LMs on content-independent ordering representations to bind entities to their attributes. These representations encode the order of entities independently of their content, and align this ordering with a corresponding order of attributes in the context, enabling the model to retrieve entity-specific information when queried. For instance, given a prompt such as ``\textit{Box A contains an apple, box B contains a banana, and box C contains cherries. What is the color of the fruit in box B?}'', the model uses ordering information corresponding to the second entity (\textit{Box B}) to retrieve the attributes of the corresponding second fruit (\textit{banana}). Prior analyses show that such ordering information is formed in intermediate layers on top of entity-related token positions and is subsequently transferred to the final token position, 
to retrieve the relevant property 
(e.g., the color ``\textit{yellow}'' of the banana).


\subsection{Ordering Information in VLMs}
\label{sec:vlm_var_binding}

In VLMs, spatial variable binding is a fundamental capability central to many spatial reasoning tasks. 
Recently, \citet{assouel2025visual} demonstrates that when VLMs are queried about objects within images, they retrieve symbolic representations that encode their ordering information. While this work establishes the existence of such ordering representations for visual variable binding tasks, 
it does not characterize the source of it. Specifically, the work mainly shows that for variable binding, the model uses this symbolic representation at the last token position, and that it comes from visual tokens; however, it remains unclear \textit{where and how} VLMs create ordering information. In particular, it remains unclear whether ordering representations are constructed within the LM backbone, inherited directly from the vision encoder, or emerge through interactions between the two components.
In this work, we show that the source of this is two-fold: the vision encoder encodes positional information to capture spatial layout, while the LM backbone further enhances this information.

\section{Experiments}
\label{sec:experiments}


\subsection{VLMs Use Ordering Information for Spatial Variable Binding}\label{subsec:last_token}

\begin{wrapfigure}[]{r}{0.52\linewidth}    
    \centering
    \setlength{\fboxsep}{0pt}      
    \setlength{\fboxrule}{1.5pt}   
    \resizebox{\linewidth}{!}{%
    \begin{tabular}{cc}
        Clean & Counterfactual \\
        \fcolorbox{gray}{white}{\includegraphics[width=0.4\linewidth]{figures_main/squares.pdf}} &
        \fcolorbox{gray}{white}{\includegraphics[width=0.4\linewidth]{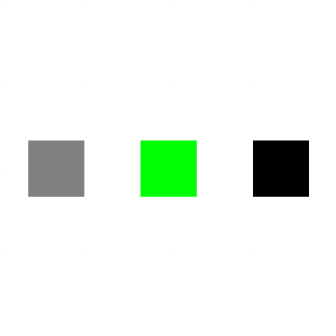}} \\
        \small
        \parbox{4cm}{\textit{The color of the square to the \textbf{left} of the green square\colorbox{yellow!60}{\tikzmark{isright} is }}}
        &
        \small
        \parbox{4cm}{\textit{The color of the square to the \textbf{right} of the green square\colorbox{yellow!60}{\tikzmark{isleft} is }}}
    \end{tabular}
    }
    \begin{tikzpicture}[overlay, remember picture]
    \draw[
    ->,
    ultra thick,
    color=black,
    line cap=round
    ]
    ([xshift=-35pt,yshift=5pt]pic cs:isleft)
    .. controls 
    ([xshift=-33pt,yshift=-0.3cm]pic cs:isleft) 
    and ([xshift=-3pt,yshift=-0.3cm]pic cs:isright)
.. ([xshift=-12pt,yshift=5pt]pic cs:isright);
    \end{tikzpicture}
    \vspace{5mm}
    \caption{\textbf{Last-Token Position Patching.}
    The residual stream at the final token (``is'') of the counterfactual run is patched into the clean run. Clean answer: \textbf{\textcolor{red}{Red}}. If ordering information is transferred, the output becomes \textbf{\textcolor{blue}{Blue}} (the clean square at the counterfactual's correct position); if attribute is transferred, it becomes \textbf{\textcolor{black}{Black}} (the counterfactual answer).}

    \label{fig:last_token_exp}
    \vspace{-3mm}
\end{wrapfigure}

We begin by establishing the presence of ordering representations that the model uses at the final token position, across a range of tasks, as described in Sec.~\ref{subsec:data}, by conducting interchange intervention experiments at the final token position. Specifically, we separately patch the residual stream vector of each layer at the last token position with that from a counterfactual sample (see Fig.~\ref{fig:last_token_exp}) featuring different square colors in the image and a different directional query. 

 \looseness=-1 If the model uses the ordinal position of the queried square to retrieve its color, we expect this ordering information to be transferred from the counterfactual run to the clean run. Concretely, in the counterfactual run corresponding to Fig.~\ref{fig:last_token_exp}, if the model encodes a representation indicating that the third object is the correct one, patching this representation into the clean run should cause the model to select the object with the same ordinal position in the clean image. As a result, the output should change from the color of the originally selected object (the leftmost square; \textcolor{red}{\textbf{Red}}) to the color of the square in the clean image corresponding to the third position (\textcolor{blue}{\textbf{Blue}}), which matches the correct position in the counterfactual run. In contrast, if the model directly transfers attribute information rather than ordering information, patching the final-token representation from the counterfactual run should cause the output to change directly to the correct answer 
of the counterfactual image itself (e.g., \textcolor{black}{\textbf{Black}} in Fig.~\ref{fig:last_token_exp}), independent of 
the clean image. 
This would indicate that color information and not order is being transferred by the intervention.

\begin{figure*}[t]
    \centering
    \includegraphics[width=1\linewidth]{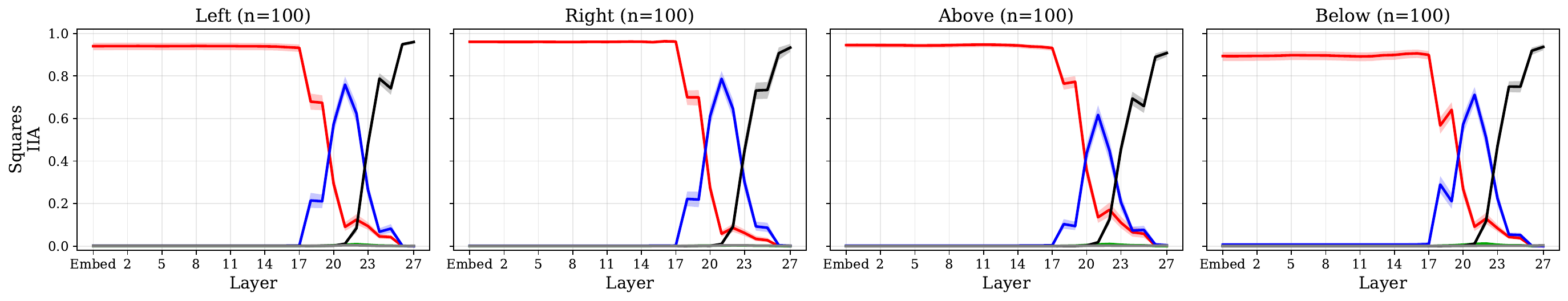}
    \vspace{-3mm}
    \caption{\textbf{Last Token Position Patching Results}: The output of the clean run remains unchanged up to layer 19. From layer 20 through layer 22, the output changes to the color of the clean square corresponding to the correct position of the counterfactual run, suggesting that these layers patch the order of the target square. Beyond layer 22, the model predicts the counterfactual answer, indicating that layers after 22 encode the final answer. 
    Shaded areas are 95\% bootstrap confidence intervals. 
    }
    \label{fig:last_token_results}
    \vspace{-3mm}
\end{figure*}

Figure~\ref{fig:last_token_results} shows that the final output initially corresponds to the color of the correct square of the clean input between layers \textcolor{red}{\textbf{0-19}}, indicating that the intervention on these layers has no effect. It then switches to selecting the color of the square in the clean image corresponding to the ordinal position of the correct square in the counterfactual image \textcolor{blue}{\textbf{20-22}}, indicating that ordering information is being transferred. At layers \textcolor{black}{\textbf{23–27}}, the model switches to the color of the counterfactual correct square, indicating that correct color attribute information is being transferred directly.

This pattern suggests that the model first computes an ordering representation of the correct square and then in later layers retrieves its corresponding color information. 
For clarity, we present results for the \textit{Squares} setting for the Qwen model. Consistent behavior is observed across all settings and models (App.~\ref{app:last_token}). Results are averaged over 100 clean--counterfactual pairs per setting, with 95\% bootstrap confidence intervals (App.~\ref{app:n100_stats}). The plots use the color scheme from the example in Fig.~\ref{fig:last_token_exp}; however, the specific colors, shapes, and object categories and queried directions vary across samples.


\vspace{-1mm}
\subsection{Is Vision Encoder the Source of Ordering Representations?}

While the previous subsection established the presence of ordering representations, it did not fully characterize them. In particular, a key open question concerns where these ordering representations are generated within VLMs: are they produced in the LM backbone, the vision encoder, or through interactions between the two components? Identifying the source of ordering representations is important not only for understanding how they arise, but also because it yields predictions that we test on natural images (Sec.~\ref{sec:fixing}).

\vspace{-2mm}
\subsubsection{Probing Visual Embeddings}
\label{subsec:probing}
\begin{wrapfigure}[]{r}{0.6\linewidth}
    \centering
    \vspace{-.5cm}
    \includegraphics[width=1\linewidth]{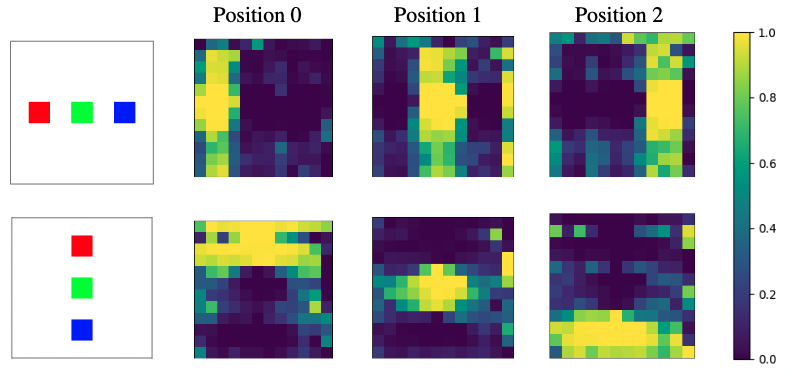}
    \caption{\textbf{Position information in visual embeddings extends beyond object regions.} We train linear probes on top of object tokens of the visual embedding to predict their order in the image. At test time, these probes generalize to background tokens, producing strips-like pattern aligned with object position, suggesting that background tokens encode order information of nearby objects. See App.~\ref{app:grid} for more complex arrangements.}
    \label{fig:probs}
    \vspace{-4mm}
\end{wrapfigure}
We begin by asking whether ordering representations originate in the vision encoder. To address this question, we train linear probes on visual token embeddings, extracted immediately after projection from the vision encoder, to predict the ordinal position of each object in the image. Successful probe performance indicates that ordering information is linearly decodable from the visual embedding.


For each image setting described in Section~\ref{subsec:data}, we train two linear probes, one for horizontal and one for vertical configurations, to classify embeddings as one of three spatial positions. Training data is constructed using token embeddings extracted from 90 images; the embeddings that correspond to the three objects in an image serve as positive examples for their respective positions. In the \textit{Squares}, \textit{Shapes}, and \textit{Objects} datasets, identifying the visual tokens associated with each object is straightforward due to their fixed spatial layouts. In contrast, objects in the \textit{What’sUp} dataset vary in size and are not aligned to fixed positions. To handle this, we preprocess \textit{What’sUp} images by extracting bounding boxes for each object and use their spatial coordinates to identify the corresponding visual tokens (see Fig.~\ref{app:whatsup_preprocessing}).


The resulting classifiers achieve nearly perfect accuracy on a test set of 30 images. In addition to classifying object-token representations, we also apply the probes to all other tokens in the image to assess whether ordering information representations are present in them.
Surprisingly, we find that the information about the object ordering is not confined to tokens corresponding to the objects themselves, but is instead distributed across multiple background tokens, as shown in Fig.~\ref{fig:probs}. Notably, probe predictions exhibit a strip-like spatial pattern aligned with object position, indicating that positional information is distributed coherently across background tokens rather than localized to object regions. In App.~\ref{app:strip_emergence} we trace at which vision-encoder layer these strips emerge and relate the pattern to the encoder's positional encoding.
This behavior contrasts with prior work on LMs \citep{prakash2024fine, prakash2025languagemodelsuselookbacks,dai-etal-2024-representational}, which shows that such representations are typically localized to a single token or a small set of adjacent tokens. In the following subsections, we demonstrate that the information contained in these strips is causally relevant for producing the correct final output.

\vspace{-1mm}
\subsubsection{Causal Intervention on Vision Tokens}
\label{subsec:vision_encoder}


Although the probing results demonstrate the presence of ordering information in visual representations, they do not by themselves establish a causal relationship with the model’s final output. We therefore perform interchange intervention experiments to test whether systematically manipulating these representations leads to corresponding changes in the model’s predictions.

\begin{wrapfigure}[]{r}{0.59\linewidth}    
    \centering
    \setlength{\fboxsep}{0pt}
    \setlength{\fboxrule}{1.5pt}
    \resizebox{\linewidth}{!}{%
    \begin{tabular}{cc}
        Clean & Counterfactual \\
        \tikzmarknode{box}{\fcolorbox{gray}{white}{\includegraphics[width=0.3\linewidth]{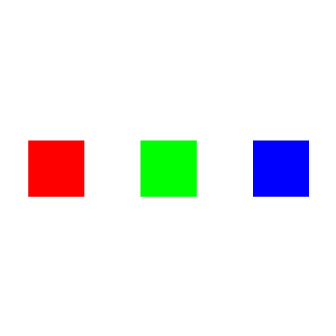}}} &
        \fcolorbox{gray}{white}{\reflectbox{\includegraphics[width=0.3\linewidth]{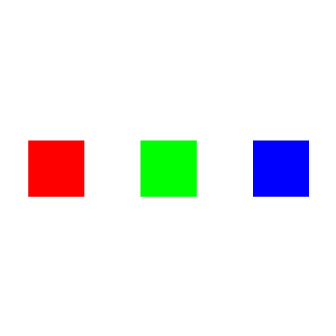}}} \\
        \tikzmarknode{box2}{\fcolorbox{gray}{white}{\includegraphics[width=0.3\linewidth]{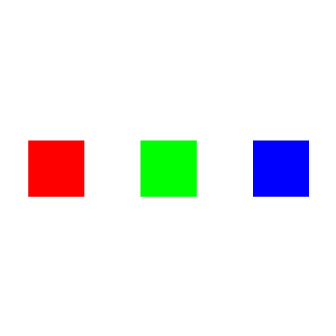}}} &
        \fcolorbox{gray}{white}{\reflectbox{\includegraphics[width=0.3\linewidth]{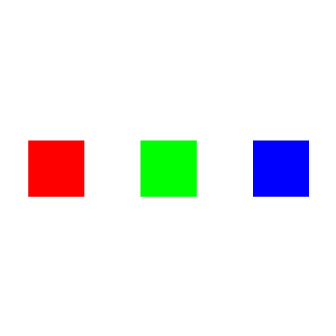}}} \\
        \small
        
        \parbox{4cm}{\centering\textit{ \vspace{1mm} The color of the square to the \textbf{left} of the green square is}}
        &
        \small
        \parbox{4cm}{\centering\textit{\vspace{1mm}The color of the square to the \textbf{right} of the green square is}}
    \end{tabular}
    }
    \begin{tikzpicture}[overlay, remember picture]
        \node[anchor=east, rotate=90, font=\scriptsize] at ([xshift=-0.4cm, yshift=1.1cm]box.west)
        {\textbf{{(a) Object Patching}}};
        
        \node[anchor=east, rotate=90, font=\scriptsize] at ([xshift=-0.4cm, yshift=1.15cm]box2.west)
        {\textbf{{(b) Strip Patching}}};

        \draw[
            line width=3pt,
            color=yellow!80
        ]
        ([xshift=1.65cm,yshift=-1.2cm]box.north west)
        rectangle
        ([xshift=2.15cm,yshift=-1.7cm]box.north west);

        \draw[
            line width=3pt,
            color=yellow!80
        ]
        ([xshift=5.75cm,yshift=-1.2cm]box.north west)
        rectangle
        ([xshift=6.25cm,yshift=-1.7cm]box.north west);

        \draw[
            line width=3pt,
            color=yellow!80
        ]
        ([xshift=.15cm,yshift=-1.2cm]box.north west)
        rectangle
        ([xshift=.65cm,yshift=-1.7cm]box.north west);
        
        \draw[
            line width=3pt,
            color=yellow!80
        ]
        ([xshift=4.2cm,yshift=-1.2cm]box.north west)
        rectangle
        ([xshift=4.7cm,yshift=-1.7cm]box.north west);
        
        \draw[
            ->,
            ultra thick,
            color=black,
            line cap=round,
            arrows={-Stealth[round]}
        ]
        ([xshift=4.4cm,yshift=-1.8cm]box.north west)
        .. controls
            ([xshift=4.1cm,yshift=-2.2cm]box.north west)
            and
            ([xshift=2.2cm,yshift=-2.2cm]box.north west)
        .. ([xshift=1.9cm,yshift=-1.8cm]box.north west);

        \draw[
            ->,
            ultra thick,
            color=black,
            line cap=round,
            arrows={-Stealth[round]}
        ]
        ([xshift=6cm,yshift=-1.1cm]box.north west)
        .. controls
            ([xshift=5.6cm,yshift=-.5cm]box.north west)
            and
            ([xshift=1cm,yshift=-.5cm]box.north west)
        .. ([xshift=0.4cm,yshift=-1.1cm]box.north west);

        \draw[
            line width=3pt,
            color=yellow!80
        ]
        ([xshift=1.6cm,yshift=-0.14cm]box2.north west)
        rectangle
        ([xshift=2.2cm,yshift=-2.3cm]box2.north west);

        \draw[
            line width=3pt,
            color=yellow!80
        ]
        ([xshift=5.7cm,yshift=-.14cm]box2.north west)
        rectangle
        ([xshift=6.3cm,yshift=-2.3cm]box2.north west);

        \draw[
            line width=3pt,
            color=yellow!80
        ]
        ([xshift=.05cm,yshift=-.14cm]box2.north west)
        rectangle
        ([xshift=.65cm,yshift=-2.3cm]box2.north west);
        
        \draw[
            line width=3pt,
            color=yellow!80
        ]
        ([xshift=4.15cm,yshift=-.14cm]box2.north west)
        rectangle
        ([xshift=4.75cm,yshift=-2.3cm]box2.north west);
        
        \draw[
            ->,
            ultra thick,
            color=black,
            line cap=round,
            arrows={-Stealth[round]}
        ]
        ([xshift=4.3cm,yshift=-1.8cm]box2.north west)
        .. controls
            ([xshift=5.1cm,yshift=-1.8cm]box2.north west)
            and
            ([xshift=2.8cm,yshift=-1.8cm]box2.north west)
        .. ([xshift=1.85cm,yshift=-1.8cm]box2.north west);

        \draw[
            ->,
            ultra thick,
            color=black,
            line cap=round,
            arrows={-Stealth[round]}
        ]
        ([xshift=5.7cm,yshift=-0.7cm]box2.north west)
        .. controls
            ([xshift=7.3cm,yshift=-.7cm]box2.north west)
            and
            ([xshift=0.65cm,yshift=-.7cm]box2.north west)
        .. ([xshift=0.3cm,yshift=-.7cm]box2.north west);

    \end{tikzpicture}

    \caption{\textbf{Causal intervention on visual token embeddings.} The visual tokens of the left and right squares are swapped (left$\leftrightarrow$right) from the counterfactual run into the clean run, either (a) for the square tokens only or (b) together with their background strips. If the patched tokens carry ordering information, the output switches to the square on the opposite side of the queried reference.}
    \vspace{-2mm}
    \label{fig:criss-cross}
\end{wrapfigure}

As illustrated in Fig.~\ref{fig:criss-cross}, we intervene on visual token embeddings by swapping the left and right strips of the clean image with those from a counterfactual image. Specifically, we replace the embeddings of the left strip in the clean image with the right-strip embeddings from the counterfactual image, and vice versa. The intervention is designed such that both patched regions contain squares of the same color, ensuring that color values are preserved and that only ordering information is altered. In addition to intervening on the visual token embeddings, we also separately patch the residual stream vectors of each layer of the LM backbone.

\begin{figure}[t]
    \centering
    \centering
    \includegraphics[width=1\linewidth]{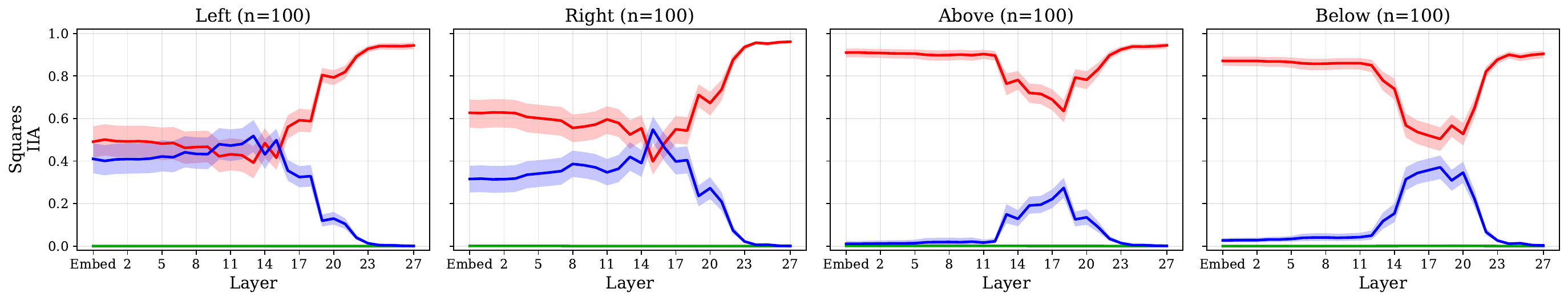}
    \vspace{-4mm}
    \caption{\textbf{Object patching}. 
    Interchange interventions applied only to visual tokens corresponding to the squares. Patching square tokens at any layer does not reliably change the model’s final output relative to the clean run, indicating that square-localized tokens alone do not carry sufficient ordering information to determine the model’s prediction. }
    \label{fig:obj_patching}
    \vspace{3mm}
    \includegraphics[width=1\linewidth]{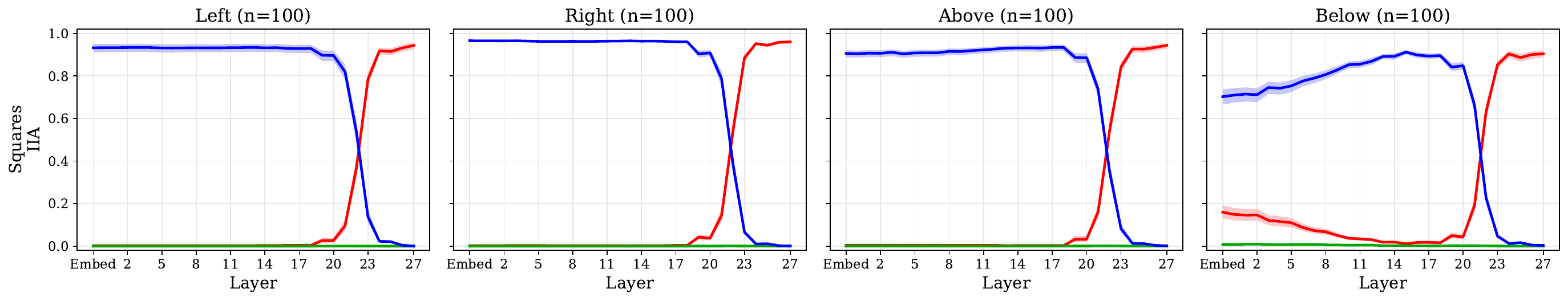}
    \vspace{-3mm}
    \caption{\textbf{Strip patching}. Interchange interventions applied to both square tokens and their background strips immediately switch the model’s output to the incorrect color in the clean image, consistent with the transferred ordering information. This demonstrates that ordering information is distributed across strip-aligned visual tokens and is causally relevant for reasoning. 
    } \label{fig:strip_patching}
    \vspace{-4mm}
\end{figure}

If ordering information encoded by the vision encoder is causally relevant, this intervention should cause the model’s final output on the clean image to switch from the left square (\textcolor{red}{red}) to the right square (\textcolor{blue}{blue}), reflecting the swapped ordering information. As shown in Fig.~\ref{fig:strip_patching}, we observe the expected change in the final output when either the visual token embeddings or the residual stream vectors up to layer 24 are patched. In contrast, patching only the square-localized visual tokens, without the surrounding strip, does not reliably induce this change (Fig.~\ref{fig:obj_patching}). These indicate that ordering information encoded by the vision encoder is not localized to object tokens, but is distributed across background visual tokens.


We observe similar behavior in other settings, including Shapes, Objects, and What’sUp, which also generalizes to all models, as described in App.~\ref{app:criss-cross}. All results are averaged over 100 clean--counterfactual pairs (shaded: 95\% bootstrap CIs). Moreover, the original correct square's logit overtakes the intervened one around layer 23, indicating that by then all required information, both ordering and color, has been transferred to the last token (Sec.~\ref{sec:vlm_var_binding}).

\vspace{-1mm}
\subsection{LM Backbone Enhances Ordering Information}
\label{subsec:sec_mech}

In the previous sections, we showed that visual token embeddings produced by the vision encoder and projector module already encode object ordering information. We now ask what role, if any, the LM backbone plays in spatial variable binding. Namely, does the LM backbone merely consume ordering information from the vision components, or can it generate such information when needed?

\vspace{-1mm}
\subsubsection{Removing Ordering Information from Vision Embeddings}\label{subsec:removing_info}

\begin{wrapfigure}{r}{0.52\linewidth}
    \centering
    \vspace{-3mm}
    \includegraphics[width=.95\linewidth]{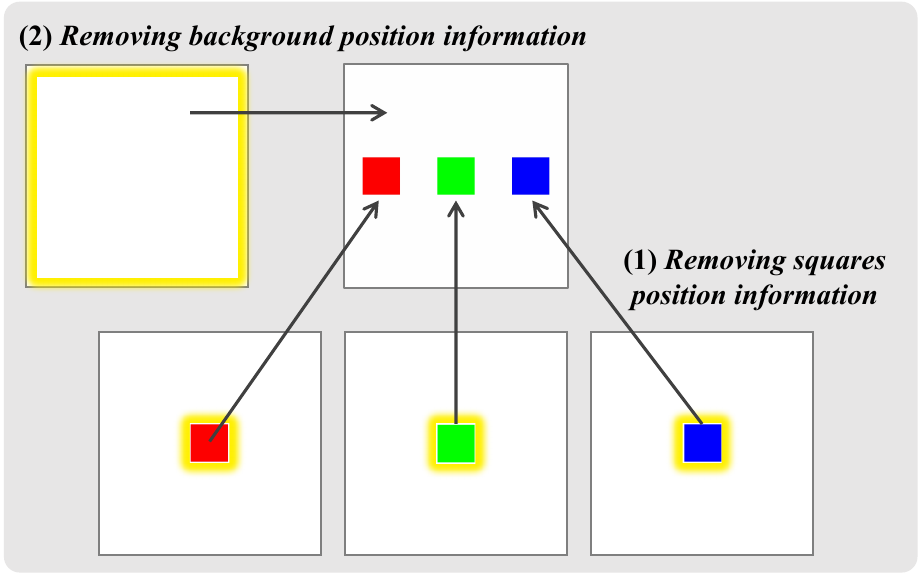}
    \caption{\textbf{Ordering information is ablated from vision embeddings} by replacing square embeddings with same-colored isolated middle squares and background embeddings with those from an empty image, preserving color while removing position cues.} \label{fig:secondary_settings}
    \vspace{-5mm}
\end{wrapfigure}

To isolate the contribution of the LM backbone, we first ablate ordering information from the visual token embeddings using the interventions illustrated in Fig.~\ref{fig:secondary_settings}. Specifically, we apply two modifications:  
(1) we replace the embedding of each square with an embedding of the same-colored square taken from an image in which that square appears alone in the middle of the image, thereby preserving color information while removing relative ordering; and  
(2) we replace all background-token embeddings with those from an empty image, eliminating ordering information encoded in background regions. Together, these interventions effectively remove ordering information originating from the vision encoder, while preserving object colors. We do not apply this intervention to the What’sUp dataset, as constructing a precise intervention is challenging for natural images.

As shown in Table~\ref{tab:behavioral_inter} (App.~\ref{app:removed_ordering_info}), removing ordering information from the vision embeddings leads to a substantial drop in performance across datasets, confirming the central role of vision-derived ordering information. However, performance remains above chance ($33.3\%$), 
suggesting that the LM backbone can compensate for the missing ordering signals.

\vspace{-1mm}
\subsubsection{Do LMs Backup Ordering Information?}
\label{sec:lm_backbone_oi}
To test whether the LM backbone generates its own ordering information, we perform an interchange intervention experiment analogous to that described in Section~\ref{subsec:vision_encoder}, but under the ablated-vision setting described above (e.g., the LM gets images with no ordering information from the vision encoder). In contrast to earlier experiments, here we patch only the square-associated visual tokens, rather than the entire strips.



\begin{figure*}[t]
    \centering
    \includegraphics[width=1\linewidth]{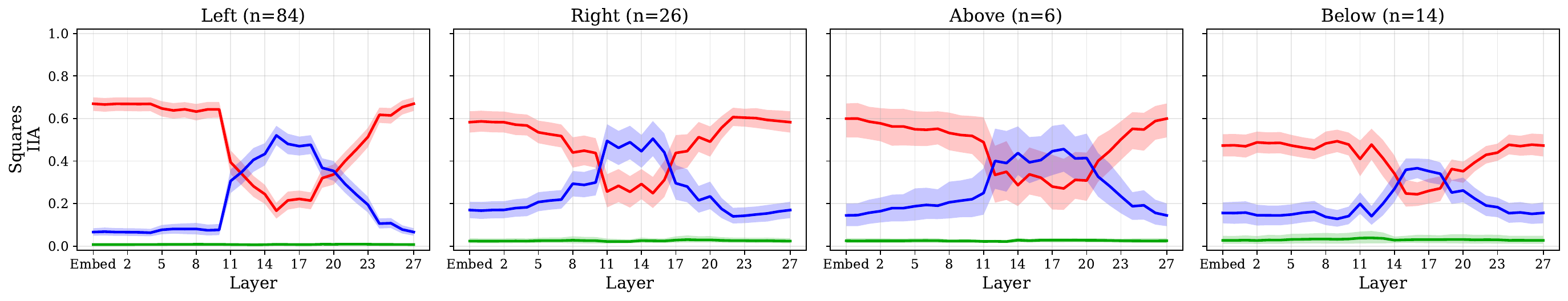}
    \vspace{-3mm}
    \caption{\textbf{Patching results after removing ordering information from vision encoder.} The model’s output switches only in intermediate layers within the range of 11-20, indicating that the LM backbone generates ordering information even when vision-derived ordering representations are absent. Panel titles give the per-direction $n$, which is capped by the accuracy of the ablated model.} \label{fig:secondary_mech}
    \vspace{-3mm}
\end{figure*}

If the LM backbone generates ordering information independently, then patching square tokens from a counterfactual run should transfer this order information and induce a change in the final output. If not, the model’s prediction should remain unchanged. The results in Fig.~\ref{fig:secondary_mech} show a flip in model prediction according to order information at middle layers. This indicates that, when vision-derived ordering information is removed, the LM backbone forms its own ordering representations in middle layers, and that patching these representations switches the prediction according to the patched order. See App.~\ref{app:criss-cross} for other settings and models.


Together, these results indicate that VLMs rely on two sources of ordering information: a primary signal originating from the vision encoder and projector module, and a secondary signal generated within the LM backbone. While the vision-derived signal dominates when present, the LM backbone can partially reconstruct ordering information and act as a backup mechanism. Supporting evidence for this interpretation is provided in Figs.~\ref{fig:strip_patching},~\ref{fig:app_strip_gemma},~\ref{fig:app_strip_qwen}, where we show that vertically arranged objects strengthen ordering signals in the same intermediate LM layers, indicating the LM backbone enhances the ordering information provided by the vision embedding in these cases. We further find that the LM-backbone ordering information is predominantly encoded by relative position, whereas the vision encoder encodes both absolute and relative position (App.~\ref{app:abs_vs_ord}).


\vspace{-1mm}
\subsection{Validation on Natural Images}
\label{sec:fixing}

Sections~\ref{subsec:last_token}--\ref{subsec:sec_mech} characterize the mechanism on controlled synthetic images. In this subsection we check whether it generalizes to natural images. Specifically, we test three hypotheses that follow from the synthetic-setting results: (i) the vision encoder's ordering representations should be decodable from natural images globally; (ii) strengthening the vision-encoder ordering information
should correct spatial binding failures; and (iii) if the ordering information is content-independent, it should transfer across object types and spatial relations.

\begin{wraptable}{r}{0.52\linewidth}    \centering
    \caption{Amplifying ordering representation corrects failures.}
    \resizebox{\linewidth}{!}{%
    \begin{tabular}{lcc cc}
        \toprule
        & \multicolumn{2}{c}{\textbf{Gemma}} & \multicolumn{2}{c}{\textbf{Qwen}}\\
        \cmidrule(lr){2-3} \cmidrule(lr){4-5}
         \textbf{Intervention} 
        & \textbf{Acc.} & \textbf{\% Corr. Fail.}
        & \textbf{Acc.} & \textbf{\% Corr. Fail.} \\
        \midrule
        None 
        & 0.53 & -- 
        & 0.60 & -- \\
        Random amp.
        & 0.57 & 9.6\% & 0.65 & 13.1\% \\
        Ordering amp.  
        & \textbf{0.72} & \textbf{40.2}\% & \textbf{0.82} & \textbf{54.5}\% \\
        \bottomrule
    \end{tabular}
    }
    \vspace{-4mm}
    \label{tab:fixing}
\end{wraptable}
We test this on COCO-spatial (Sec.~\ref{subsec:data}; examples in Fig.~\ref{fig:app:coco}), where the model states the relation (left/right or above/below) between a query and a reference object. Since the position probes of Sec.~\ref{subsec:probing} are trained to capture ordering information, adding their directions to the visual embeddings should enhance the vision-encoder ordering signal. Concretely, for each visual token $t$ we set $\mathrm{emb}_t \leftarrow \mathrm{emb}_t + \alpha_i \cdot \mathrm{probe}_i$, where $\mathrm{probe}_i$ is the probe for direction $i$ and $\alpha_i \in [1,15]$. The intervention is applied to all visual tokens and probe directions, so neither object locations nor the queried direction is needed at test time; probes are trained once on annotated object tokens (App.~\ref{app:fix-setup}). 

Table~\ref{tab:fixing} reports \texttt{Gemma-3-4b-it} and \texttt{Qwen2-VL-2B-Instruct} (the \texttt{7B} variant is near ceiling here), using a per-item $\alpha$; fixed-$\alpha$ results and all other models are in App.~\ref{app:fixing}. Ordering directions correct $40\%$ and $54\%$ of failures versus $10\%$ and $13\%$ for random directions,
so the gains are specific to ordering information, which confirms that vision-encoder ordering information also dominates in complex natural images. This experiment aims to test generalization to natural settings, where causal interventions are hard to design, not to reach state-of-the-art accuracy. Still, chain-of-thought (CoT) prompting is no substitute: it raises \texttt{Qwen2-VL-2B}'s accuracy but lowers \texttt{Gemma-3-4b}'s, and breaks about three times as many previously correct items as amplification (App.~\ref{app:cot}). Amplification costs at most $5$ points on seven non-spatial VQA benchmarks (App.~\ref{app:side_effects}).

We conduct additional experiments to check the generalization of these results to further natural-image settings.
(i)~\emph{Transferability}: position probes trained on object images, as in Sec.~\ref{subsec:probing}, recover the strip-like ordering patterns on What'sUp images (App.~\ref{app:prob_whatsup}) and on synthetic objects over real backgrounds (App.~\ref{app:probe_real_back}). This indicates that the ordering information captured by probe directions is transferable across object and image settings.
(ii)~\emph{Correcting more failures}: amplification also corrects $43$--$55\%$ of failures on the What'sUp composites
(App.~\ref{app:fixing_whatsup}) and $53$--$68\%$ for additional object relations such as ``on/under'' and ``in front/behind'' on the original, unstitched What'sUp images (examples in Fig.~\ref{fig:app:other_relations}; App.~\ref{app:other_relations}). This indicates that the findings also hold for more complex object relations, including depth. The gains persist when $\alpha$ is selected on a validation split and evaluated on held-out failures (App.~\ref{app:alpha_split}).
(iii)~\emph{Content independence}: a probe trained on What'sUp corrects COCO-spatial failures as well as an in-domain probe ($48\%$ vs.\ $47\%$), and a horizontal probe corrects vertical failures nearly as well as a vertical one ($33\%$ vs.\ $35\%$; App.~\ref{app:probe_transfer}), indicating a shared ordering code rather than dataset- or axis-specific features.
Together, these results indicate that the mechanism identified in the controlled synthetic settings also governs spatial variable binding in complex natural scenes.

\vspace{-2mm}
\section{Conclusions}
We show that spatial variable binding in VLMs relies on ordering representations arising from two complementary sources: a primary signal encoded by the vision encoder and a secondary mechanism formed within the LM backbone. By isolating and intervening on these representations, we show that the same mechanism governs natural images, where amplifying them corrects binding failures. Our results highlight that mechanistic understanding can improve model performance.

\ifpreprint 
\section{Acknowledgement}
This research was supported by Sony PlayStation, ARL, MIT-IBM Watson AI Lab, and Hyundai Motor Company. NP and DB were supported by Coefficient Giving.
\fi

\bibliographystyle{iclr2027_conference} \bibliography{main}

\newpage
\newpage
\appendix
\textbf{{\LARGE{Appendix}}}

We bring here additional results and experimental details.

\section{Data}
\label{app:data}

\subsection{Models' Accuracy}
\label{app:models_accuracy}
Table~\ref{tab:behavioral_all_models} reports the accuracy of all five models on the studied tasks; Table~\ref{tab:behavioral_all} breaks the synthetic settings down by queried direction.
\begin{table}[h]
    \centering
    \caption{Accuracy of all five models on the studied tasks.}
    \begin{tabular}{lcccc}
        \toprule
        & Squares & Shapes & Objects & What'sUp \\
        \midrule
        Qwen2-VL-7B-Instruct & 1.00 & 1.00 & 1.00 & 0.73 \\
        Gemma-3-4b-it        & 0.98 & 0.99 & 0.99 & 0.62 \\
        Pixtral-12B          & 0.55 & 0.55 & 0.63 & 0.44 \\
        Qwen2-VL-2B-Instruct & 1.00 & 1.00 & 0.96 & 0.66 \\
        Qwen3-VL-8B-Instruct & 1.00 & 1.00 & 1.00 & 0.88 \\
        \bottomrule
    \end{tabular}
    \label{tab:behavioral_all_models}
\end{table}

\begin{table}[h]
    \resizebox{\linewidth}{!}{%
    \centering
    \begin{tabular}{lcccccccccccc}
        & \multicolumn{4}{c}{Squares} 
        & \multicolumn{4}{c}{Shapes} 
        & \multicolumn{4}{c}{Objects} \\
        \cmidrule(lr){2-5}
        \cmidrule(lr){6-9}
        \cmidrule(lr){10-13}
        & L & R & A & B
        & L & R & A & B
        & L & R & A & B \\
        \midrule
        Qwen2-VL-7B-Instruct & 1.00 & 1.00 & 1.00 & 0.99 & 1.00 & 1.00 & 1.00 & 1.00 & 1.00 & 1.00 & 1.00 & 1.00 \\
        Qwen2-VL-2B-Instruct & 1.00 & 1.00 & 1.00 & 1.00 & 1.00 & 1.00 & 1.00 & 1.00 & 0.96 & 0.99 & 0.95 & 0.92 \\
        Qwen3-VL-8B-Instruct & 1.00 & 1.00 & 1.00 & 1.00 & 1.00 & 1.00 & 1.00 & 1.00 & 1.00 & 1.00 & 1.00 & 1.00 \\
        Gemma-3-4b-it        & 1.00 & 1.00 & 0.93 & 0.97 & 1.00 & 1.00 & 0.97 & 0.97 & 1.00 & 1.00 & 0.95 & 1.00 \\
        Pixtral-12B          & 0.59 & 0.57 & 0.50 & 0.54 & 0.25 & 0.57 & 0.68 & 0.69 & 0.61 & 0.57 & 0.67 & 0.67 \\
    \end{tabular}
    }
    \vspace{2mm}
    \caption{Full behavioral performance, according to queried directions (Left, Right, Above, Below)}
    \label{tab:behavioral_all}
\end{table}

\subsection{Image Settings}
\label{app:image_settings}
Each of our settings consists of images constructed from the following possible objects: 

\begin{quote}
\textbf{Squares:} \\
\texttt{red}, \texttt{green}, \texttt{blue}, \texttt{black}, \texttt{brown}, \texttt{gray}

\vspace{0.5em}
\textbf{Shapes:} \\
\texttt{red square}, \texttt{green circle}, \texttt{blue triangle}, \texttt{black star}, \texttt{brown cross}, \texttt{gray heart}

\vspace{0.5em}
\textbf{Objects:} \\
\texttt{orange}, \texttt{apple}, \texttt{pot}, \texttt{bell}, \texttt{hat}, \texttt{rose}, \texttt{key}, \texttt{bomb}
\end{quote}

\begin{figure}[h!]
    \centering
    \setlength{\fboxsep}{0pt}
    \setlength{\fboxrule}{1.5pt}
    \begin{subfigure}[b]{0.32\linewidth}
        \centering
         \fcolorbox{gray}{white}{\reflectbox{\includegraphics[width=\linewidth]{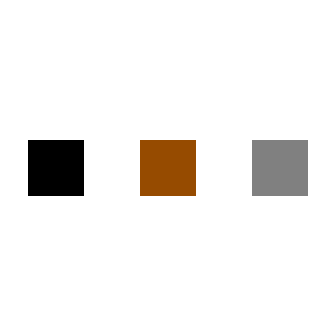}}}
        \caption{Square setting}
        \label{fig:square_img}
    \end{subfigure}
    \hfill
    \begin{subfigure}[b]{0.32\linewidth}
        \centering
         \fcolorbox{gray}{white}{\reflectbox{\includegraphics[width=\linewidth]{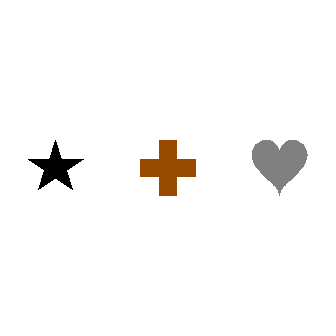}}}
        \caption{Shapes setting}
        \label{fig:shape_img}
    \end{subfigure}
    \hfill
    \begin{subfigure}[b]{0.32\linewidth}
        \centering
         \fcolorbox{gray}{white}{\reflectbox{\includegraphics[width=\linewidth]{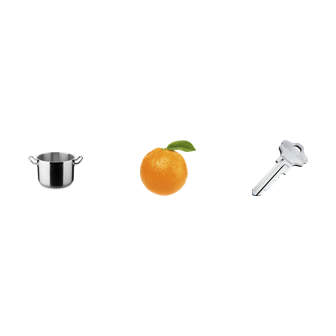}}}
        \caption{Objects setting}
        \label{fig:objects_img}
    \end{subfigure}
    
    \caption{Example images of the Square, Shapes, and Objects settings.}
    \label{fig:all_settings}
\end{figure}

For each model, we report the following parameters in Table \ref{tab:image_params}.
\begin{quote}
\begin{itemize}
    \item \textbf{\# Image tokens}: total number of image tokens in an image, as a square grid of \texttt{num tokens} $\times$ \texttt{num tokens} tokens.
    \item \textbf{\# Pixels / Token}: number of image pixels per token, as $\texttt{num pixels} \times \texttt{num pixels}$ pixels per token.
    \item \textbf{\# Tokens / Object}: the number of tokens each object in the image occupies.
    \item \textbf{Object spacing}: the horizontal/vertical distance, in tokens, between adjacent objects in an image.
    \item \textbf{Strip width}: the strip width, in tokens, used for interventions such as those in Fig.~\ref{fig:strip_patching}.
\end{itemize}
\end{quote}

\begin{table}[h]
\resizebox{\linewidth}{!}{%
\centering
\begin{tabular}{lccccc}
\toprule
\textbf{Model} &
\textbf{\# Image tokens} &
\textbf{\# Pixels / Token} &
\textbf{\# Tokens / Object} &
\textbf{Object spacing} &
\textbf{Strip width} \\
\midrule
Qwen2-VL-7B-Instruct & $12 \times 12$ & $28 \times 28$ & $2 \times 2$ & $2$ & $4$\\
Qwen2-VL-2B-Instruct & $12 \times 12$ & $28 \times 28$ & $2 \times 2$ & $2$ & $4$\\
Qwen3-VL-8B-Instruct & $12 \times 12$ & $32 \times 32$ & $2 \times 2$ & $2$ & $4$\\
Gemma-3-4b-it       & $16 \times 16$ & $56 \times 56$ & $3 \times 3$ & $3$ & $3$\\
Pixtral-12B         & $15 \times 15$ & $16 \times 16$ & $2 \times 2$ & $2$ & $2$\\
\bottomrule
\end{tabular}
}
\caption{Image tokenization and spatial layout parameters for each model. Qwen2-VL-2B shares the 7B vision encoder (14-px patches merged $2\times2$), and Qwen3-VL-8B uses 16-px patches merged $2\times2$, so both produce the same $12\times12$ grid as Qwen2-VL-7B. Pixtral's encoder has a fixed 16-px patch stride, so a 240-px canvas yields a $15\times15$ grid.}\label{tab:image_params}
\end{table}




\newpage
\subsection{Prompts}
\label{app:prompts}
Generation prompt tokens are added to the end of each prompt to signal the beginning of the assistant response.

\begin{quote}
\textbf{System prompt (all settings):} \\
\texttt{You are a helpful assistant. You respond in one token.}

\vspace{0.5em}
\textbf{User prompt templates:} \\
Each \texttt{<direction>} $\in$ \{\texttt{left}, \texttt{right}, \texttt{above}, \texttt{below}\} is mapped to a preposition: 
\begin{center}
\texttt{\{left $\mapsto$ ``to the left of'', right $\mapsto$ ``to the right of'',} \\
\texttt{\ \ above $\mapsto$ ``above'', below $\mapsto$ ``below''\}}
\end{center}

\vspace{0.5em}
\textbf{Squares:} \\
\texttt{The color of the square <preposition> the <middle color> square is}

\textbf{Shapes:} \\
\texttt{The color of the shape <preposition> the <middle color> <middle shape> is}

\textbf{Objects:} \\
\texttt{The object <preposition> the <middle object> is a(n)}

\textbf{What’s Up (Objects):} \\
\texttt{The object on the floor <preposition> the <middle object> is a(n)}

\vspace{0.5em}
\textbf{Two-choice prompt templates:}
For the templates below, \texttt{<option 1>} and \texttt{<option 2>} denote the two answer options: object labels for What’sUp and relation labels for COCO. Each example is evaluated under both option orderings:
\texttt{<option 1> or <option 2>} and \texttt{<option 2> or <option 1>}.

\textbf{What’sUp (Objects):} \\
\texttt{Which object is <preposition> the <middle object>: <option 1> or <option 2>? Answer with one word.}



\textbf{COCO-spatial, horizontal (Gemma):} \\
\texttt{Is the <subject object> to the <option 1> or to the <option 2> of the <reference object>? Answer with one word.}

\textbf{COCO-spatial, vertical (Gemma):} \\
\texttt{Is the <subject object> <option 1> or <option 2> the <reference object>? Answer with one word.}
 
\end{quote}

\clearpage
\subsection{Natural-Image Validation -- Experimental setup}
\label{app:fix-setup}

Following the dataset construction described in Sec.~\ref{subsec:data}, we evaluate the ordering-amplification intervention from Sec.~\ref{sec:fixing} on the What’sUp composites and COCO-spatial examples. 

\begin{figure}[h]
    \centering
    \setlength{\fboxsep}{0pt}      
    \setlength{\fboxrule}{1.5pt}   

    \begin{tabular}{ccc}
        \includegraphics[width=0.25\linewidth]{figures_main/whatsup.pdf} &
        \includegraphics[width=0.25\linewidth]{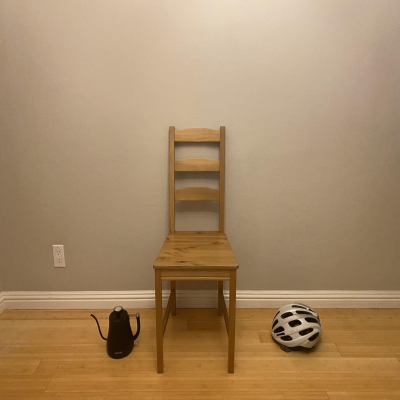} & 
        {\includegraphics[width=0.25\linewidth]{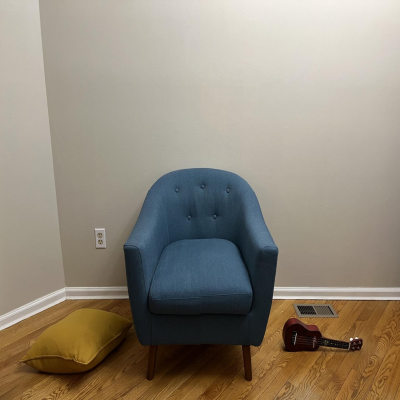}} 
    \end{tabular}
    \vspace{0.5em}
    \small
    \caption{\textbf{Examples for images from What'sUp control~\citep{kamath2023whatsup}}}
    \label{fig:app:whatsup}
    \vspace{-3mm}
\end{figure}

For What’sUp, each composite image contains 3 objects, yielding two two-option forced-choice questions; querying the object to the \texttt{left} and to the \texttt{right} of the middle reference object. For example, a possible query for the leftmost image in Fig~\ref{fig:app:whatsup} is \textit{Which object is to the left of the chair: a book or a lamp? Answer with one word.} and the other query is {Which object is to the right of the chair: a book or a lamp? Answer with one word.}

For COCO-spatial, each pair of objects serves as subject--reference for querying, and is converted into a binary spatial-relation question: \texttt{left}/\texttt{right} for horizontal relations and \texttt{above}/\texttt{below} for vertical relations. This formulation reduces dependence on open-ended object naming or recognition. For example, a possible query for the leftmost image in Fig~\ref{fig:app:coco} is \textit{Is the dog to the left or to the right of the newspaper? Answer with one word.} and the other query is {Is the newspaper to the left or to the right of the dog? Answer with one word.}

\begin{figure}[h]
    \centering
    \setlength{\fboxsep}{0pt}      
    \setlength{\fboxrule}{1.5pt}   

    \begin{tabular}{cccc}
        \includegraphics[width=0.2\linewidth]{figures_main/coco4.jpg} &
        \includegraphics[width=0.2\linewidth]{figures_main/coco2.jpg} &
        \includegraphics[width=0.2\linewidth]{figures_main/coco3.jpg} & 
        {\includegraphics[width=0.2\linewidth]{figures_main/coco5.jpg}}
    \end{tabular}
    \vspace{0.5em}
    \small
    \caption{\textbf{Examples for images from COCO-spatial~\citep{kamath2023whatsup,lin2014microsoft}}}
    \label{fig:app:coco}
    \vspace{-3mm}
\end{figure}

To control for answer-option order bias, each query is evaluated twice, once with the answer options A and B in the query, and once with the reverse order: option B and A (for example \textit{Is <subject> to the \textbf{left} or to the \textbf{right} of <reference>} and then again with reverse answer options: \textit{Is <subject> to the \textbf{right} or to the \textbf{left} of <reference>})
An example is counted as correct only if both queries are answered correctly; otherwise, it is treated as a failure. Under random guessing, this strict criterion yields a \(25\%\) chance-level accuracy.

Amplification directions are obtained from the position probes described in Sec.~\ref{subsec:probing}. For What’sUp, baseline-correct data consists of composite images for which both directional examples, \texttt{left} and \texttt{right}, are correct under the strict two-ordering protocol. We train a three-class \texttt{left}/\texttt{middle}/\texttt{right} position probe using visual-token embeddings from the left side-object, central reference-object, and right side-object strips. For COCO-spatial, we train separate two-class probes for horizontal and vertical relations using baseline-correct subject--reference examples. The original COCO bounding boxes of the subject and reference objects are mapped to visual tokens, whose embeddings are mean-pooled within each box to obtain object-level representations for probe training. Probes are trained as linear classifiers with AdamW (lr=$10^{-3}$, weight\_decay=$10^{-4}$) for 200 epochs.

As described in Sec.~\ref{sec:fixing}, the intervention amplifies all the learned ordering directions in the visual embeddings. For each failure example, we test only the queries that were incorrect before amplification (e.g., asking if the subject is \textit{to the left or to the right} of the reference, and not about whether it is \textit{to the right or to the left} if the second was answered correctly). 
The example is counted as corrected only if all previously failed query(s) become correct under the intervention. If the model originally failed under both queries, the same amplification coefficient \(\alpha\) must correct both.

We use consistent notation across all amplification tables in Sec.~\ref{app:fixing}. \textsc{Probe} and \textsc{Random} denote the fixed-\(\alpha\) setting, where a single coefficient is selected across all examples. \textsc{Probe*} and \textsc{Random*} denote the per-example setting, where each example may use the coefficient that yields its best correction outcome, serving as an upper bound on intervention performance. We report post-intervention accuracy and corrected-failure rate, i.e., the fraction of baseline failures corrected by the intervention.

\paragraph{Held-out selection of the amplification coefficient.}
\label{app:alpha_split}
The fixed-$\alpha$ results above select the single best $\alpha$ on the same failures on which it is evaluated. To verify that this does not inflate the reported fix rates, we repeat the experiment with a validation/test split of the baseline failures (seed 0; approximately 70\%/30\%): $\alpha$ is chosen only on the validation failures and the fix rate is then measured on the held-out failures. Probe training is unaffected, as probes are trained only on baseline-correct examples. Table~\ref{tab:alpha_split} shows that the held-out fix rates closely track the all-failures results of Tables~\ref{tab:amplification_full} and~\ref{tab:amplification_whatsup} and remain far above the random-direction control, whose held-out fix rate drops to $0$--$19\%$.

\begin{table}[h]
\centering
\small
\caption{Held-out generalization of the amplification coefficient. $\alpha$ is selected on validation failures only (seed 0, $\approx$70/30 split) and evaluated on the held-out failures; ``fix'' is the fraction of held-out failures corrected. Gemma-3 COCO-spatial rows come from a run on a 394-item subset of COCO-spatial (baseline $185/394=47.0\%$), so their failure counts differ from Table~\ref{tab:amplification_full}.}
\label{tab:alpha_split}
\vspace{1mm}
\resizebox{\linewidth}{!}{%
\begin{tabular}{llcc cc cc}
\toprule
& & & & \multicolumn{2}{c}{Probe} & \multicolumn{2}{c}{Random} \\
\cmidrule(lr){5-6} \cmidrule(lr){7-8}
Dataset & Model & Axis & $n_{\text{val}}/n_{\text{test}}$ & $\alpha$ & Held-out fix & $\alpha$ & Held-out fix \\
\midrule
COCO-spatial & Qwen2-VL-2B & horizontal & 75/31  & 8  & 32.3\% (10/31)  & 15 & 16.1\% (5/31) \\
COCO-spatial & Qwen2-VL-2B & vertical   & 49/21  & 7  & 38.1\% (8/21)   & 12 & 19.0\% (4/21) \\
COCO-spatial & Gemma-3-4b  & horizontal & 97/41  & 14 & 22.0\% (9/41)   & 15 & 17.1\% (7/41) \\
COCO-spatial & Gemma-3-4b  & vertical   & 50/21  & 15 & 28.6\% (6/21)   & 10 & 0.0\% (0/21) \\
What'sUp     & Qwen2-VL-2B & horizontal & 78/33  & 13 & 45.5\% (15/33)  & 15 & 0.0\% (0/33) \\
What'sUp     & Gemma-3-4b  & horizontal & 310/134 & 7 & 58.2\% (78/134) & 13 & 12.7\% (17/134) \\
\bottomrule
\end{tabular}
}
\end{table}

\clearpage
\section{Additional Discussions}

\subsection{Limitations.}
\label{sec:app:limitations}
This work focuses on a small set of spatial relations as a minimal setting that enables rigorous causal analysis. The mechanistic analysis covers left/right and above/below, and extends to diagonal arrangements in App.~\ref{app:diagonal}; the amplification experiment additionally covers the on/under and front/behind relations of What'sUp (App.~\ref{app:other_relations}). More complex configurations, such as ``between'' or hierarchical relations, and a mechanistic account of depth relations such as front/behind, are left for future work. The mechanistic analysis is further limited to controlled and composite settings, as causal interchange interventions require paired inputs that differ in exactly one factor --- a condition that cannot currently be satisfied with unmodified natural images. Nevertheless, the amplification experiment of Sec.~\ref{sec:fixing} extends to complex natural images from COCO-spatial. Several of the additional analyses are run on a single model: the probe-transfer experiments (App.~\ref{app:probe_transfer}) on \texttt{Qwen2-VL-2B} and the diagonal analysis on \texttt{Qwen2-VL-7B}. Finally, while we validate our findings across five open-weight VLMs, results on closed-source models or architectures with fundamentally different vision encoders remain unexplored. We view these as important directions for future work.

\subsection{Broader Impact Statement}
\label{sec:app:impact}
This work underscores the critical role of the vision encoder in multimodal systems, challenging the prevailing focus on the language backbone. By demonstrating that the dominant symbolic representations for spatial variable binding originate in the visual components, we highlight that future advances in VLMs must prioritize the training of robust vision encoders to ensure accurate spatial layout encoding. Furthermore, we show that the identified mechanisms account for model behavior on natural images, where amplifying vision-derived spatial signals corrects spatial binding failures without additional training or fine-tuning.

Methodologically, our discovery that spatial information is diffused across multiple visual tokens, extending even into background regions, signals a necessary paradigm shift for interpretability research. Standard analysis techniques that focus on single tokens are insufficient for capturing these distributed representations. As the field moves toward reasoning models that generate increasingly long sequences of tokens, developing interpretability methods capable of analyzing information distributed across multiple tokens will be essential for diagnosing and enhancing model behaviors.

\subsection{Compute Resources}
\label{sec:app:compute}
All the experiments were carried out on two 80GB NVIDIA A100 GPUs.

\clearpage
\section{Additional results}

\subsection{Last Token Position Patching} 
\label{app:last_token}
\begin{figure}[h]
    \centering
    \includegraphics[width=1\linewidth]{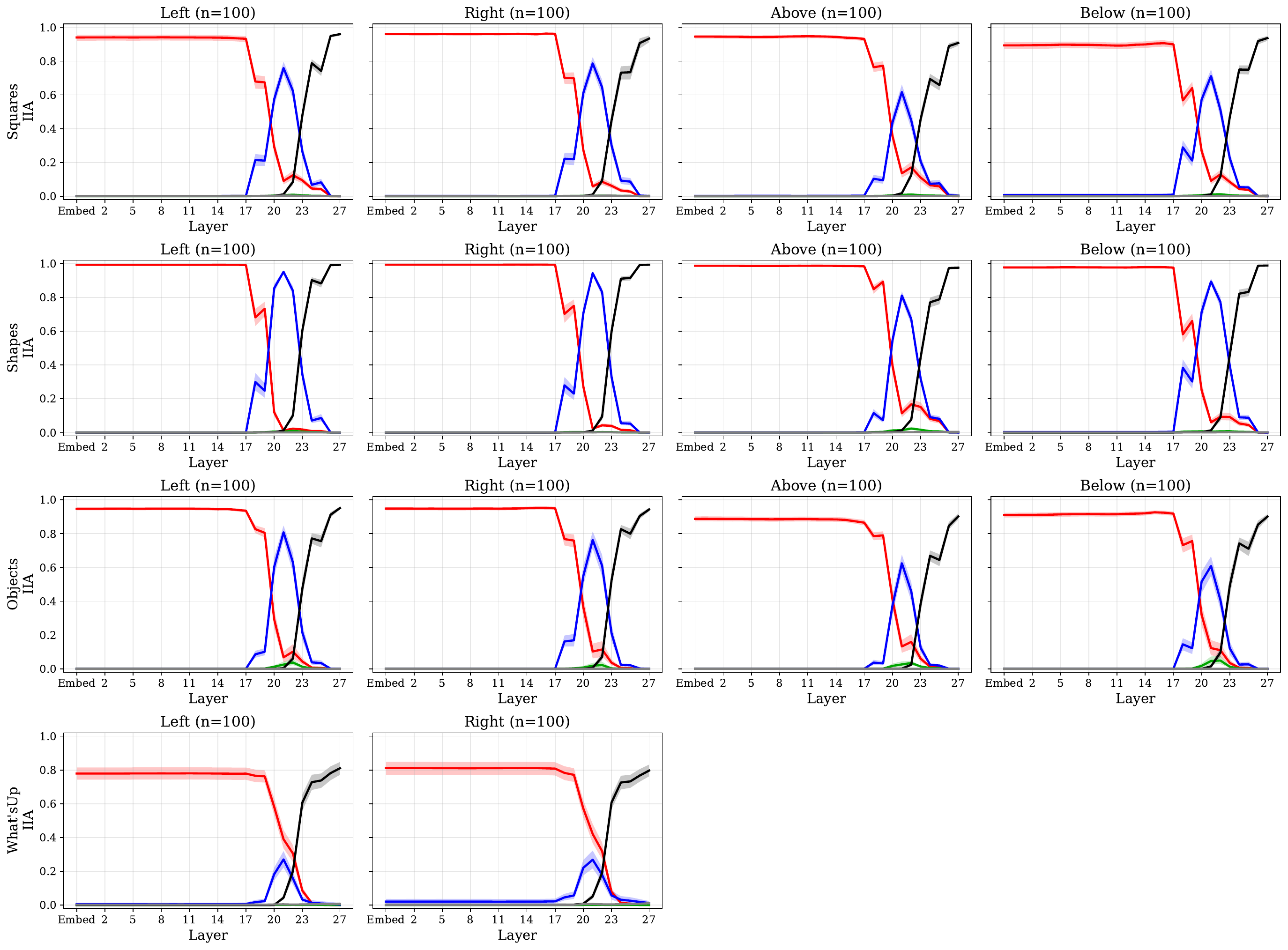}
    \caption{Final token interchange intervention experiments with the \texttt{Qwen2-VL-7B-Instruct} model.}
    \label{fig:last_token_qwen}
\end{figure}

\begin{figure}[h]
    \centering
    \includegraphics[width=1\linewidth]{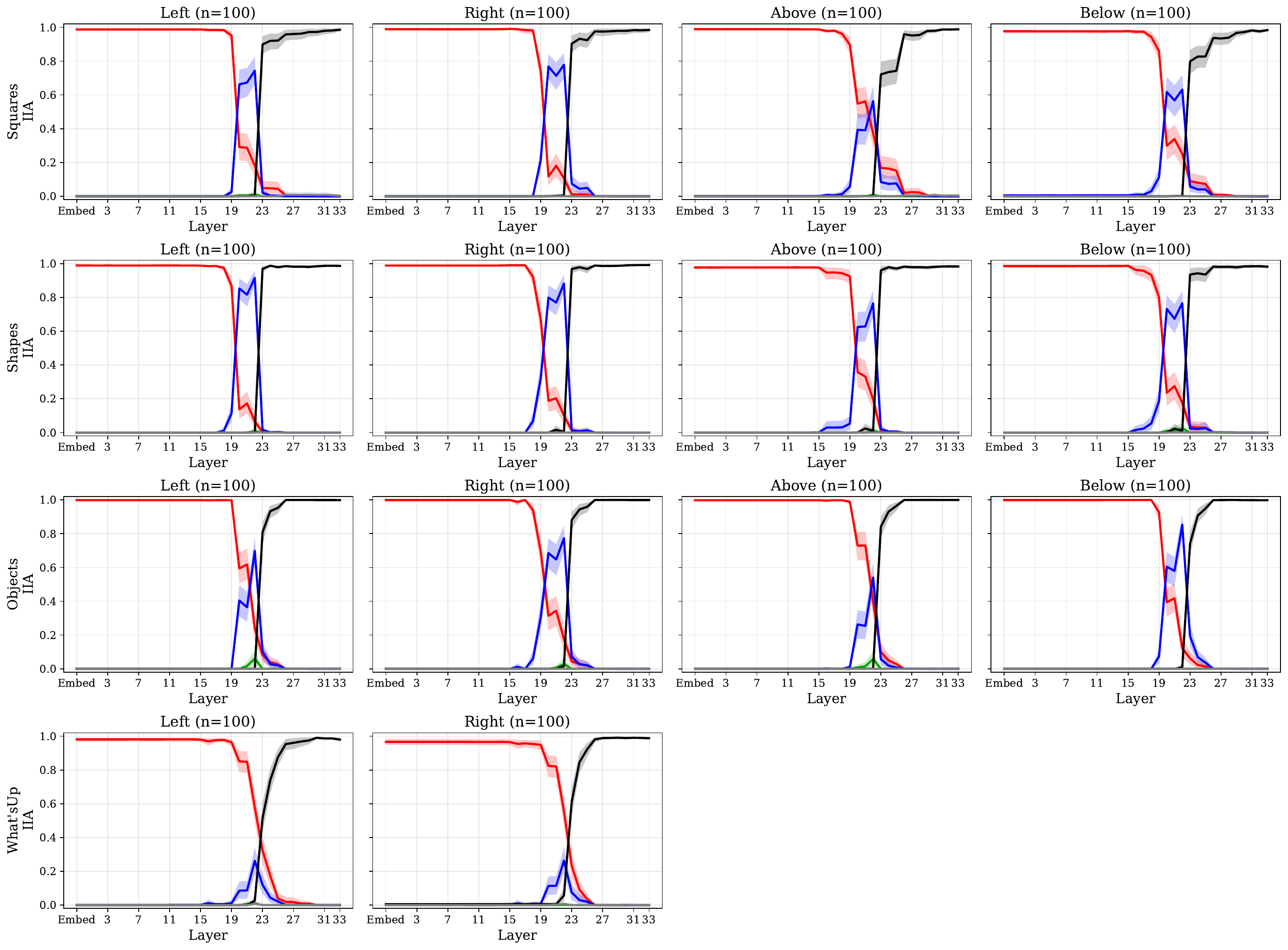}
    \caption{Final token interchange intervention experiments with the \texttt{Gemma-3-4b-it} model.}
    \label{fig:last_token_gemma}
\end{figure}

\begin{figure}[h]
    \centering
    \includegraphics[width=1\linewidth]{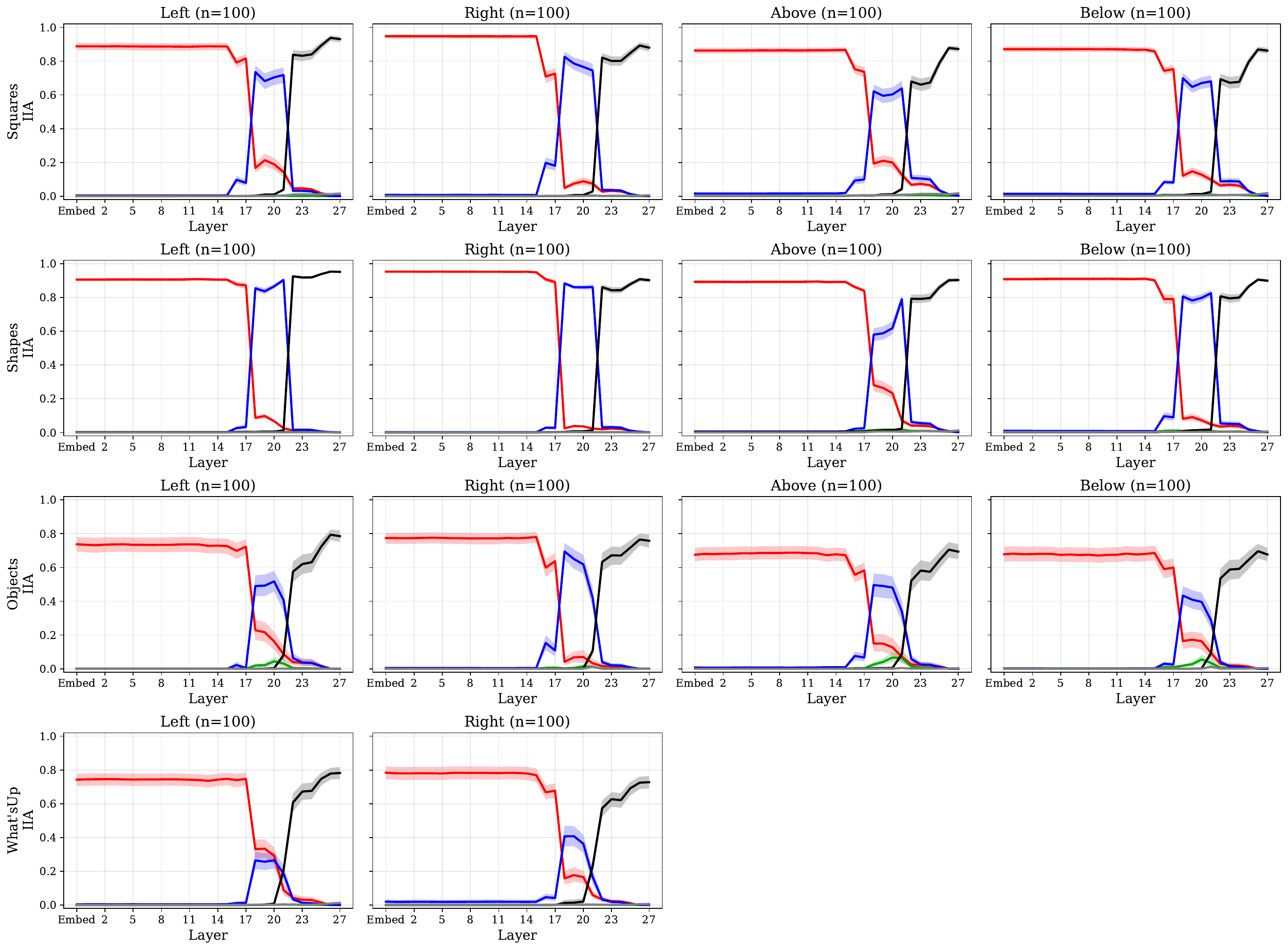}
    \caption{Final token interchange intervention experiments with the \texttt{Qwen2-VL-2B-Instruct} model.}
    \label{fig:last_token_qwen2vl2b}
\end{figure}

\begin{figure}[h]
    \centering
    \includegraphics[width=1\linewidth]{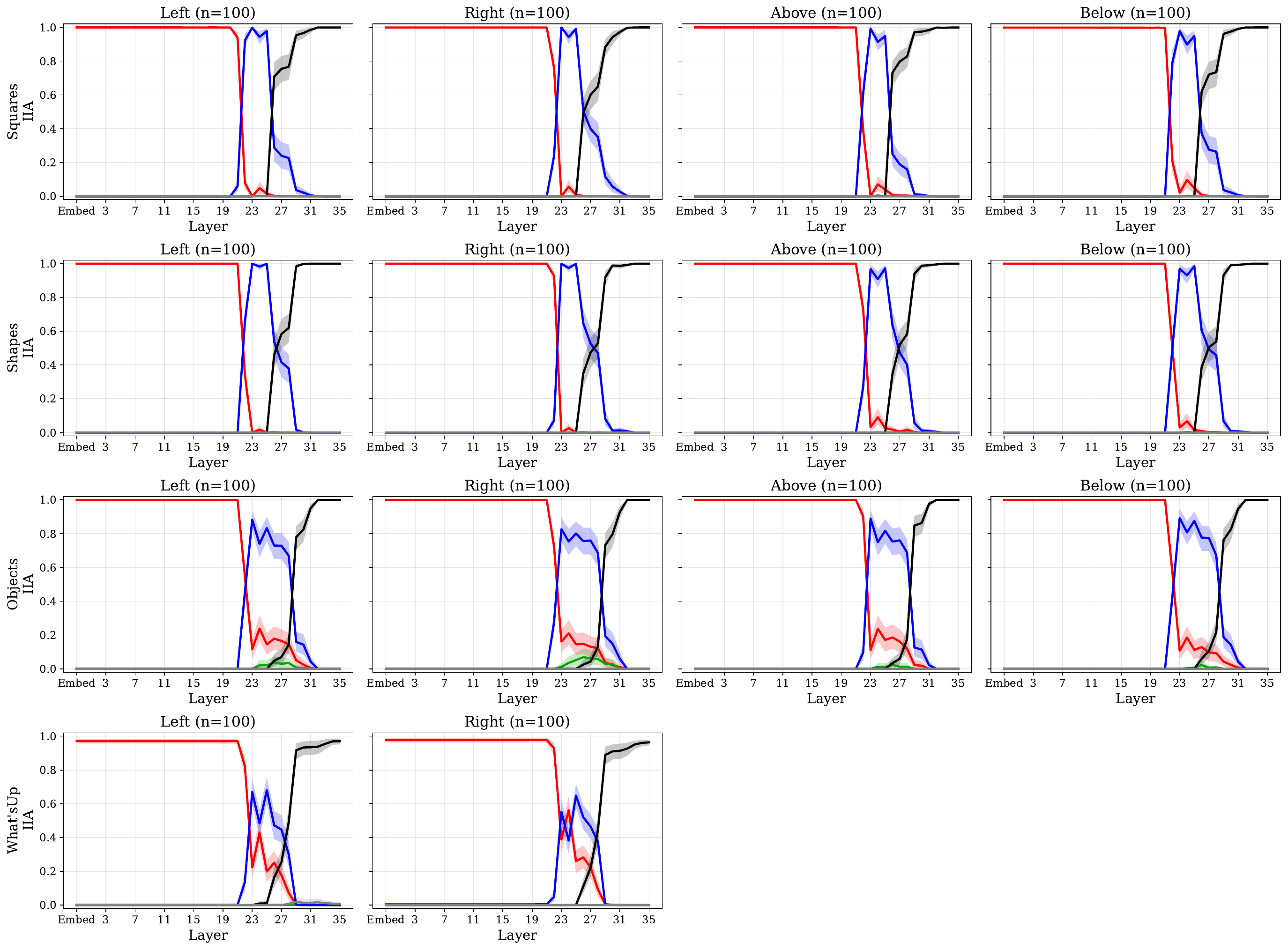}
    \caption{Final token interchange intervention experiments with the \texttt{Qwen3-VL-8B-Instruct} model.}
    \label{fig:last_token_qwen3vl}
\end{figure}

\begin{figure}[h]
    \centering
    \includegraphics[width=1\linewidth]{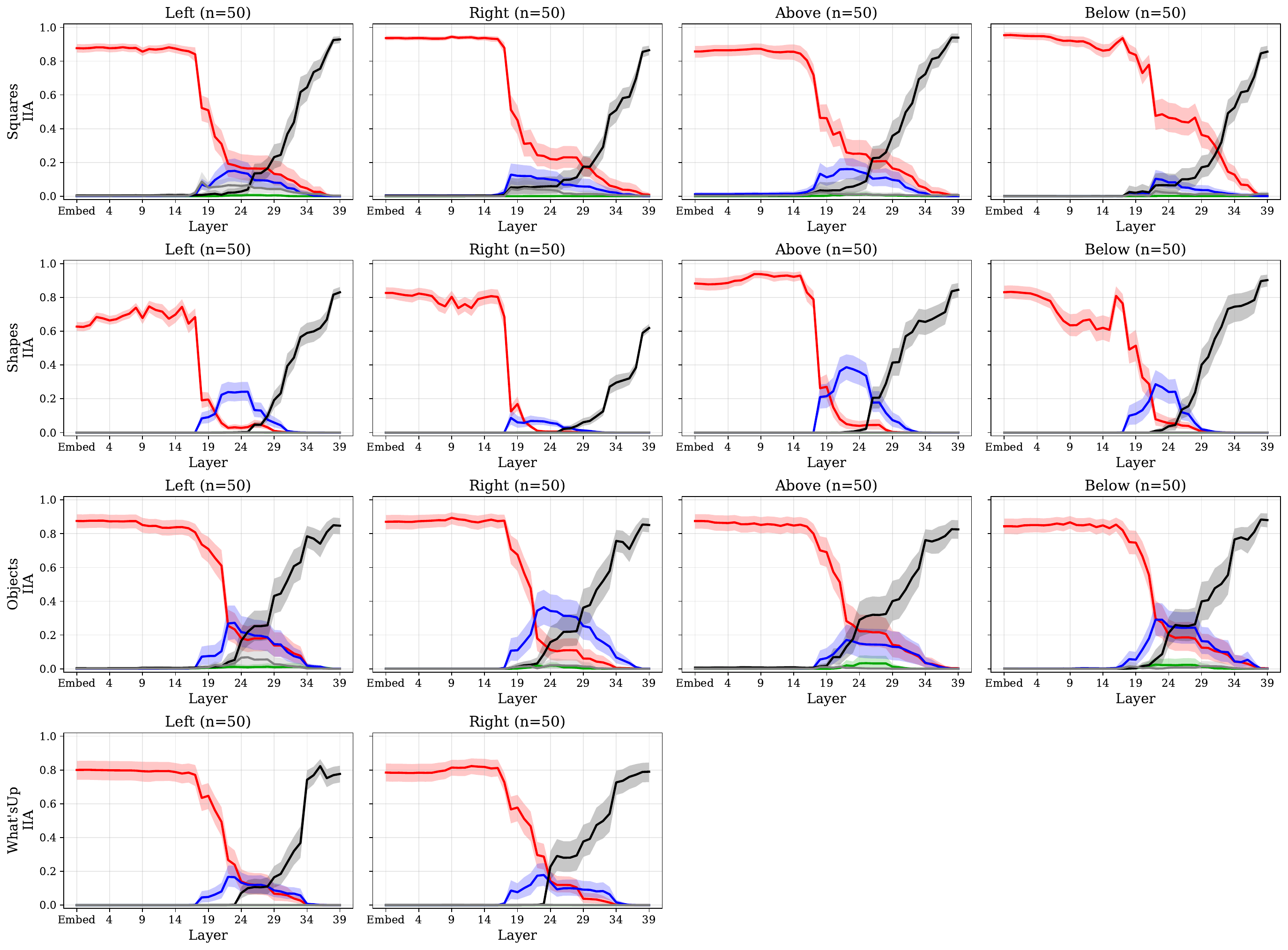}
    \caption{Final token interchange intervention experiments with the \texttt{Pixtral-12B} model ($n{=}50$ pairs; see text).}
    \label{fig:last_token_pixtral}
\end{figure}

In addition to the results described in Section~\ref{subsec:last_token}, we report the results of the last-token interchange intervention experiment on all settings (Squares, Shapes, Objects, and What'sUp) and all models: \texttt{Qwen2-VL-7B-Instruct} (Fig.~\ref{fig:last_token_qwen}), \texttt{Gemma-3-4b-it} (Fig.~\ref{fig:last_token_gemma}), \texttt{Qwen2-VL-2B-Instruct} (Fig.~\ref{fig:last_token_qwen2vl2b}), \texttt{Qwen3-VL-8B-Instruct} (Fig.~\ref{fig:last_token_qwen3vl}), and \texttt{Pixtral-12B} (Fig.~\ref{fig:last_token_pixtral}). All curves are means over $n{=}100$ clean--counterfactual pairs with 95\% bootstrap confidence bands, except Pixtral, for which the original $n{=}50$ runs are shown (its lower task accuracy, Table~\ref{tab:behavioral_all_models}, does not admit 100 pairs on which the model is correct for both the clean and the counterfactual prompt). The same three-phase pattern --- clean answer, then the answer at the counterfactual's ordinal position, then the counterfactual answer --- appears in every model and setting, with the transition shifted to the model's own late layers (e.g., layers 18--21 of 28 for Qwen2-VL-2B, layers 22--25 of 36 for Qwen3-VL-8B).



\clearpage
\subsection{Removing Ordering-Information from Visual Embeddings Impact on Behavioral Performance}
\label{app:removed_ordering_info}
We report the behavioral performance of the VLMs after removing the ordering information encoded by the vision encoder. We observe a significant drop, demonstrating its vital role in the spatial reasoning capabilities of VLMs.
\begin{table}[h]
    \centering
    \caption{Behavioral performance after removing spatial information from vision embeddings (chance is $0.33$). Pixtral is evaluated in bf16.}
    \begin{tabular}{lcccc}
        \toprule
          & Squares & Shapes & Objects \\
        \midrule
        Qwen2-VL-7B-Instruct & 0.60 & 0.62 & 0.43 \\
        Qwen2-VL-2B-Instruct & 0.62 & 0.51 & 0.19 \\
        Gemma-3-4b-it        & 0.57 & 0.64 & 0.54 \\
        Pixtral-12B          & 0.41 & 0.78 & 0.42 \\
        \bottomrule
    \end{tabular}
    \vspace{-3mm}
    \label{tab:behavioral_inter}
\end{table}


\subsection{Vision Token Patching} 
\label{app:criss-cross}
\begin{figure}[h]
    \centering
    \includegraphics[width=1\linewidth]{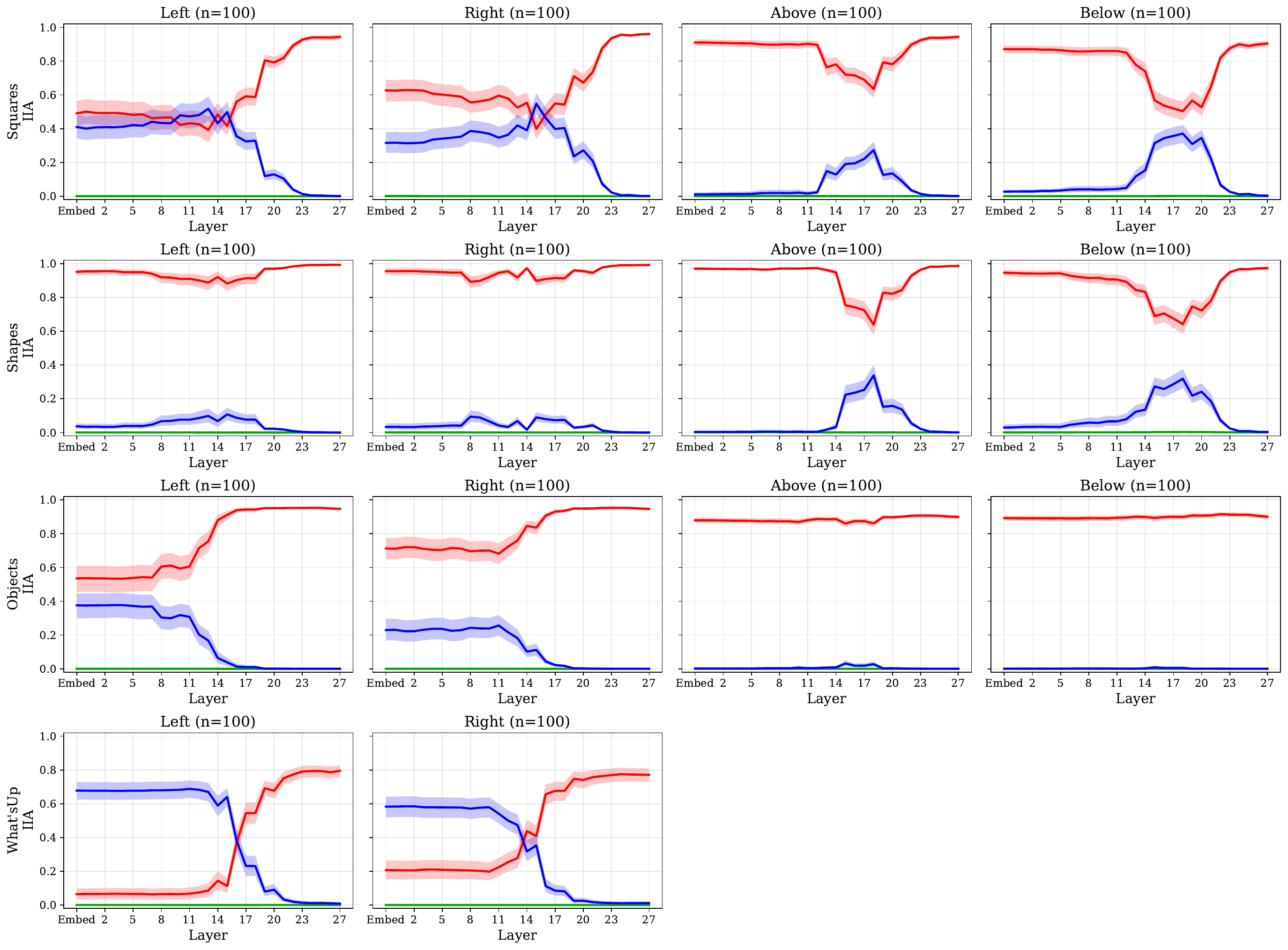}
    \caption{Visual token embeddings of square tokens interchange intervention experiment on \texttt{Qwen2-VL-7B-Instruct} model.}
    \label{fig:criss_cross_square_qwen}
\end{figure}

\begin{figure}
    \centering
    \includegraphics[width=1\linewidth]{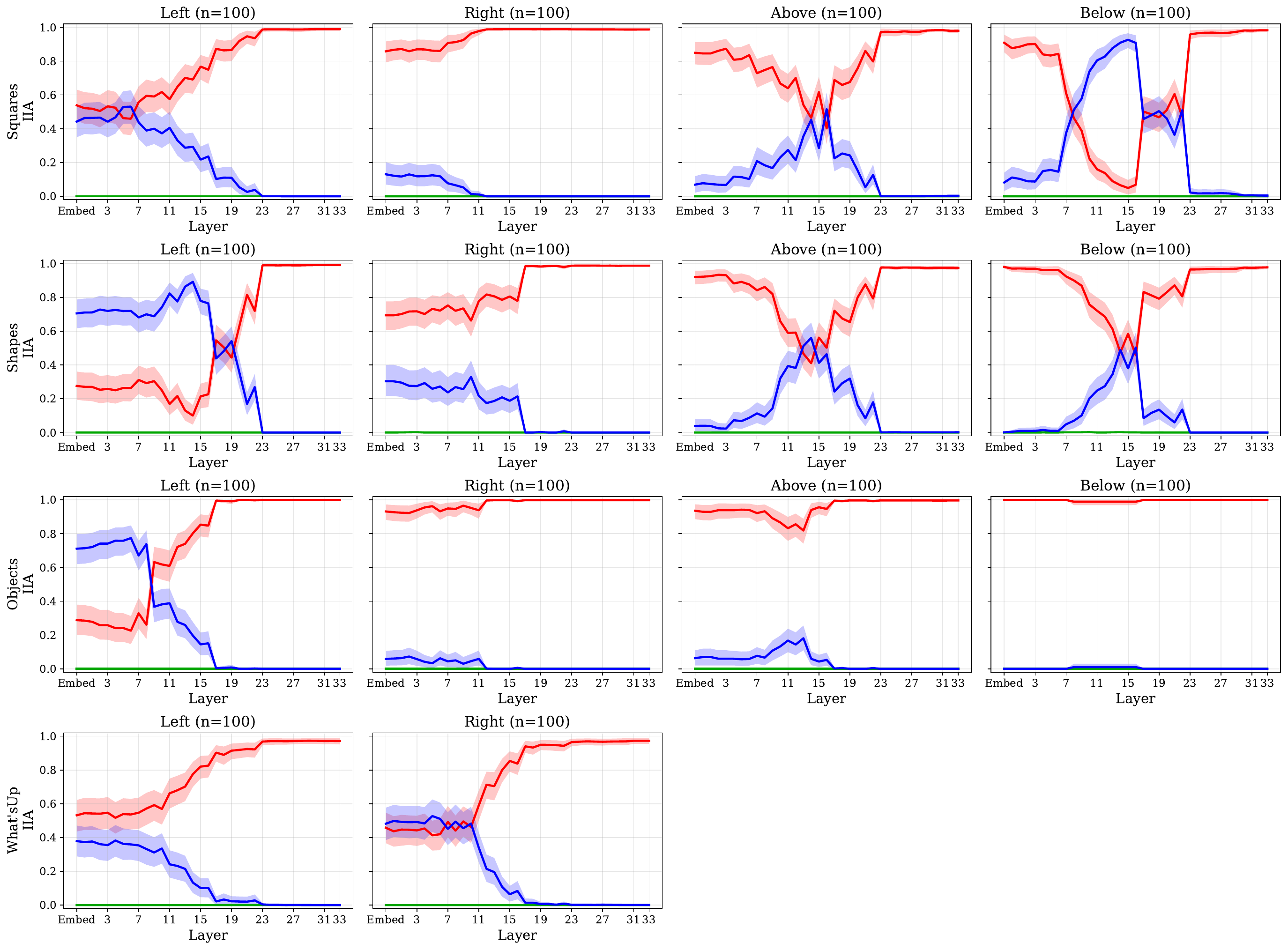}
    \caption{Visual token embeddings of square tokens interchange intervention experiment on \texttt{Gemma-3-4b-it} model.}
    \label{fig:criss_cross_gemma}
\end{figure}

\begin{figure}
    \centering
    \includegraphics[width=1\linewidth]{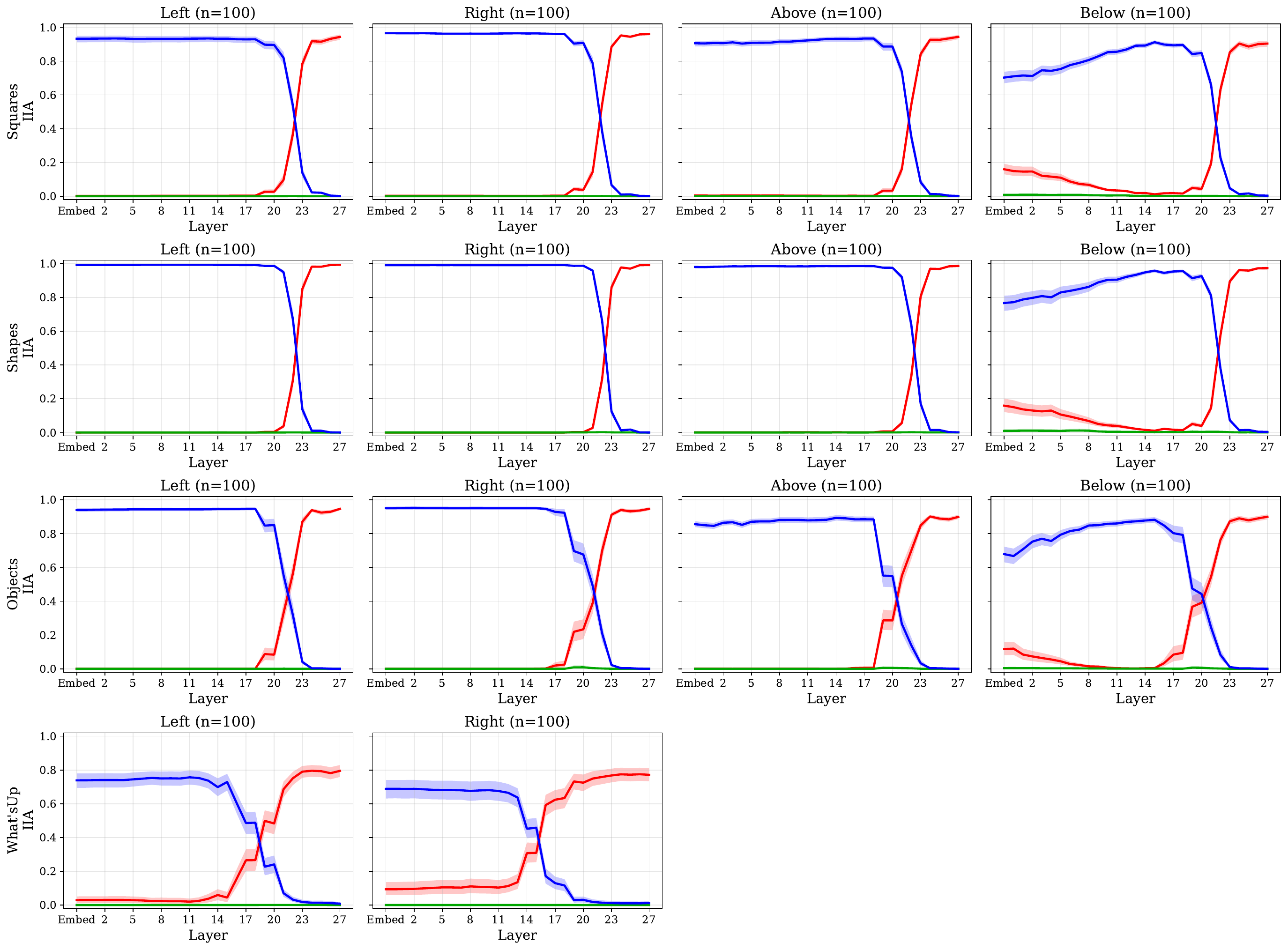}
    \caption{Visual token embeddings of strip tokens interchange intervention experiment on \texttt{Qwen2-VL-7B-Instruct} model.}
    \label{fig:criss_cross_strip_qwen}
\end{figure}

\begin{figure}
    \centering
    \includegraphics[width=1\linewidth]{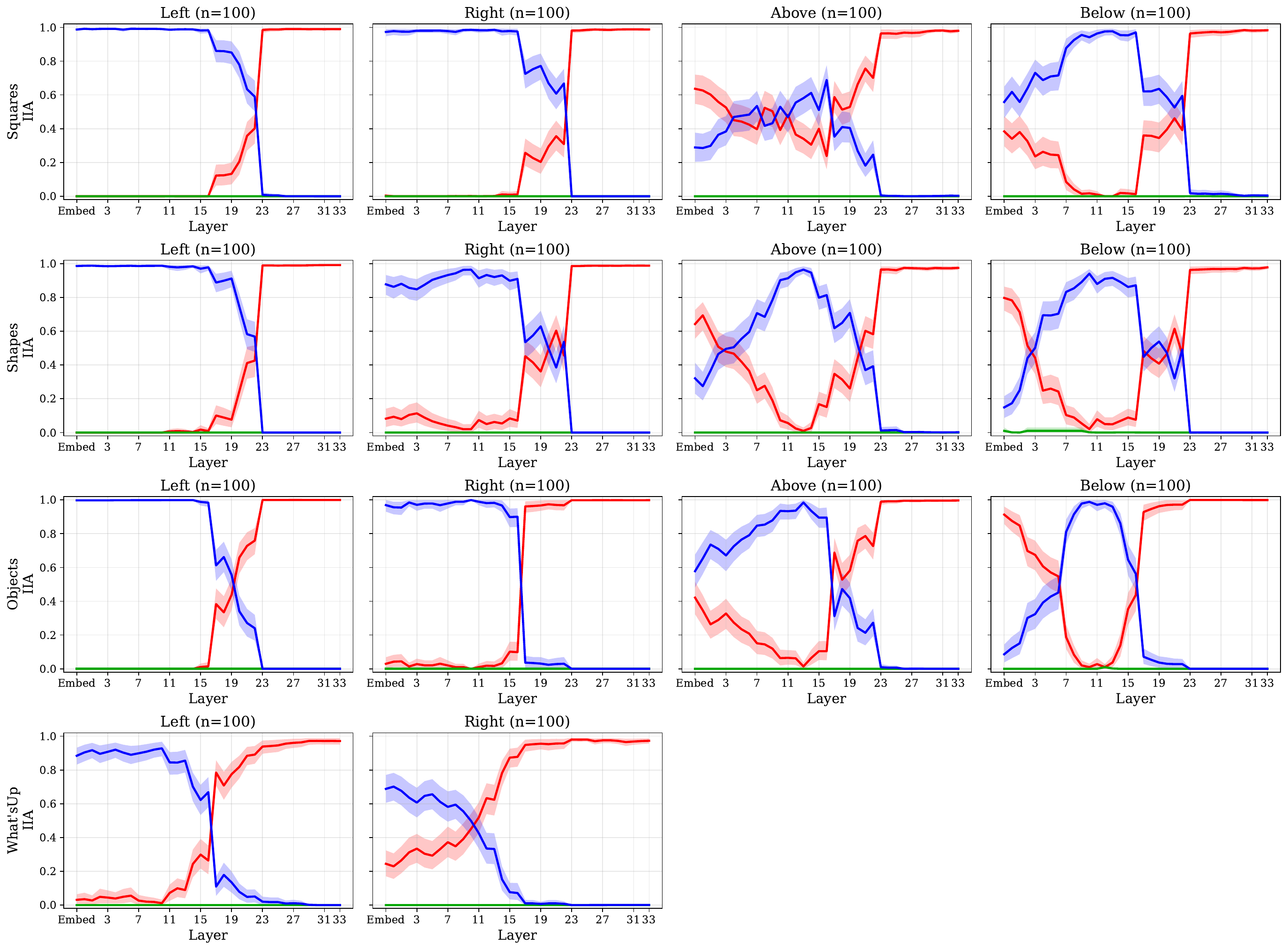}
    \caption{Visual token embeddings of strip tokens interchange intervention experiment on \texttt{Gemma-3-4b-it} model.}
    \label{fig:criss_cross_strip_gemma}
\end{figure}

\begin{figure}[h]
    \centering
    \includegraphics[width=1\linewidth]{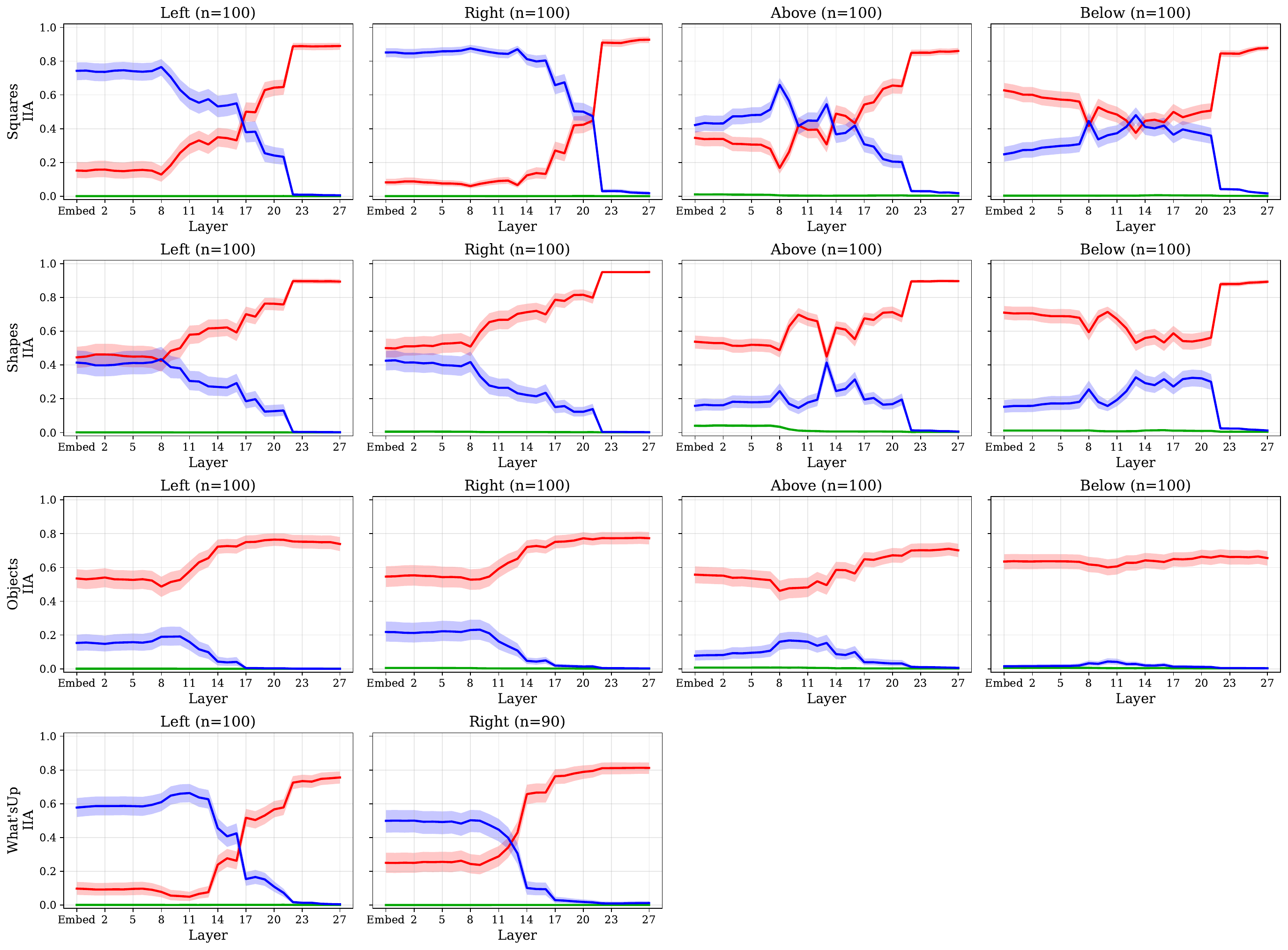}
    \caption{Visual token embeddings of square (object) tokens interchange intervention experiment on the \texttt{Qwen2-VL-2B-Instruct} model.}
    \label{fig:criss_cross_qwen2vl2b}
\end{figure}

\begin{figure}[h]
    \centering
    \includegraphics[width=1\linewidth]{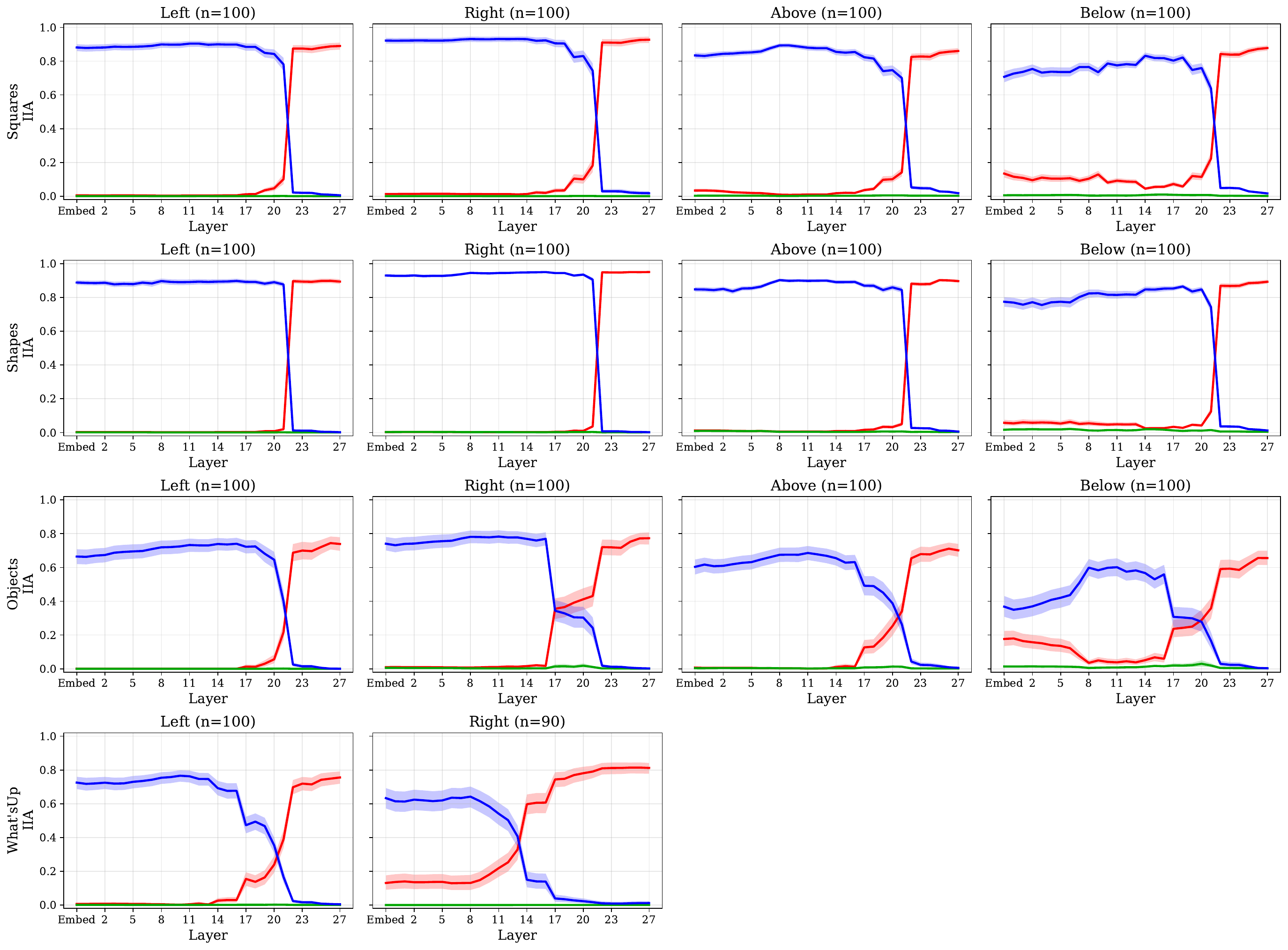}
    \caption{Visual token embeddings of strip tokens interchange intervention experiment on the \texttt{Qwen2-VL-2B-Instruct} model.}
    \label{fig:criss_cross_strip_qwen2vl2b}
\end{figure}

\begin{figure}[h]
    \centering
    \includegraphics[width=1\linewidth]{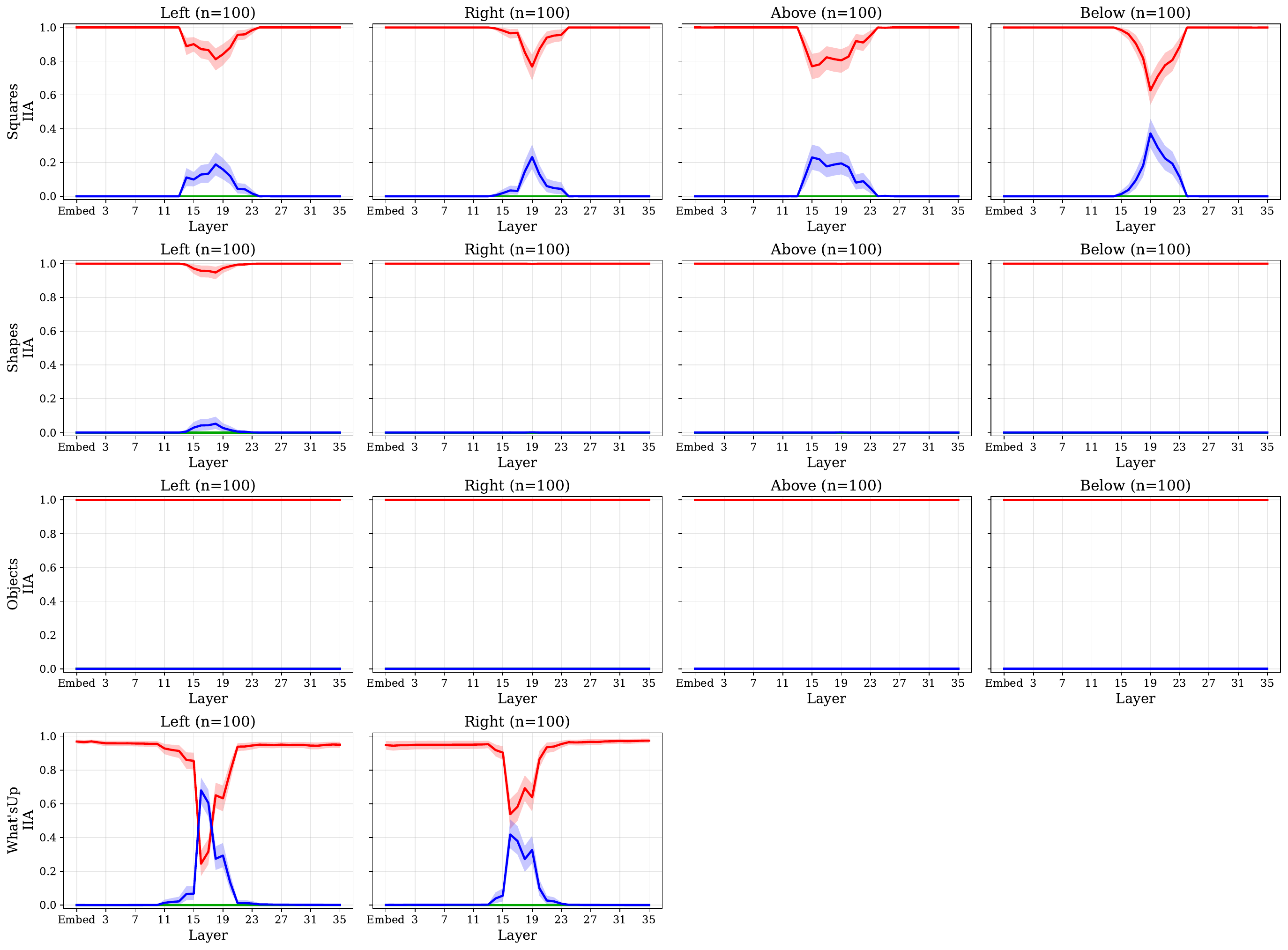}
    \caption{Visual token embeddings of square (object) tokens interchange intervention experiment on the \texttt{Qwen3-VL-8B-Instruct} model.}
    \label{fig:criss_cross_qwen3vl}
\end{figure}

\begin{figure}[h]
    \centering
    \includegraphics[width=1\linewidth]{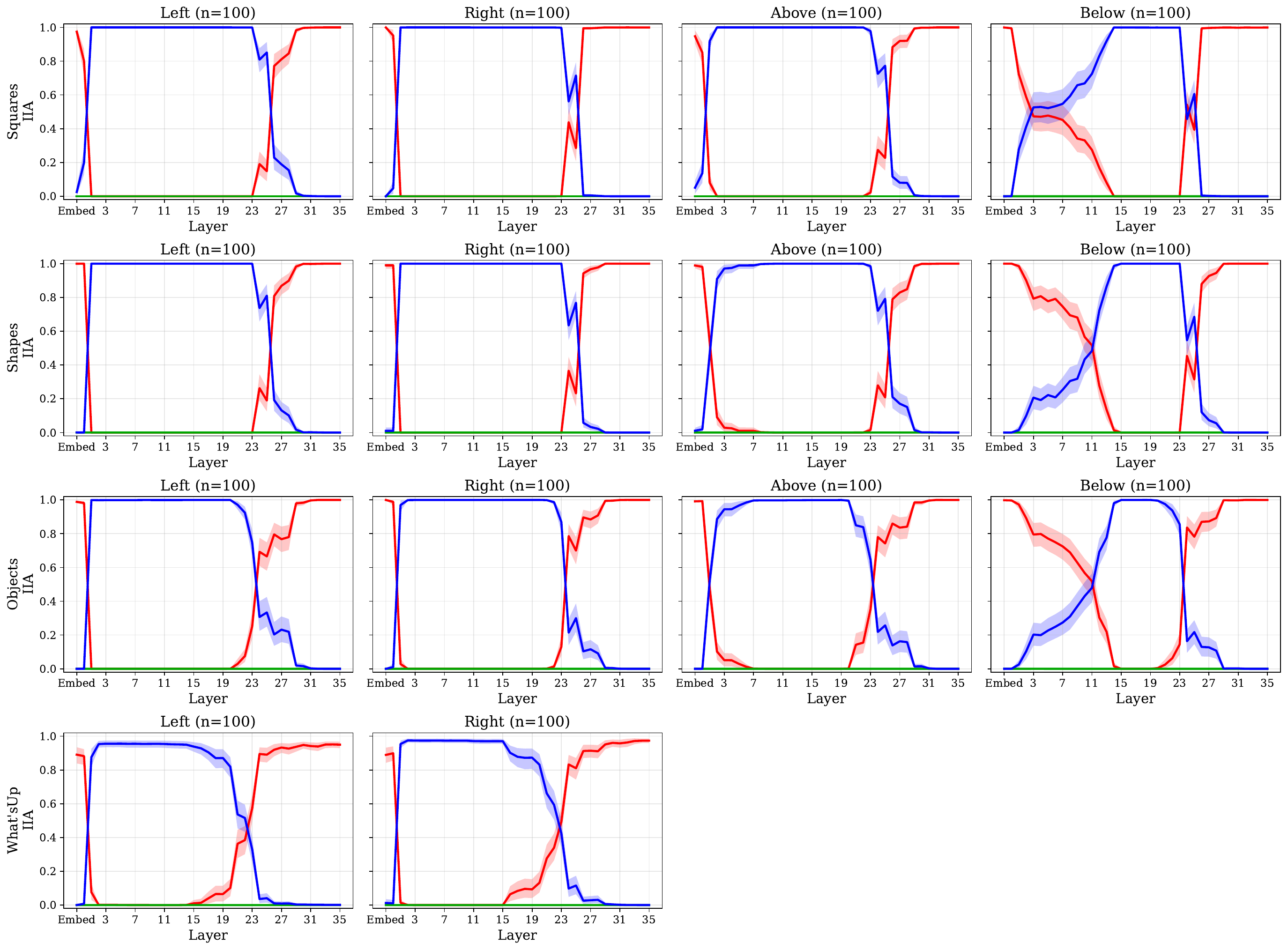}
    \caption{Visual token embeddings of strip tokens interchange intervention experiment on the \texttt{Qwen3-VL-8B-Instruct} model.}
    \label{fig:criss_cross_strip_qwen3vl}
\end{figure}

\begin{figure}[h]
    \centering
    \includegraphics[width=1\linewidth]{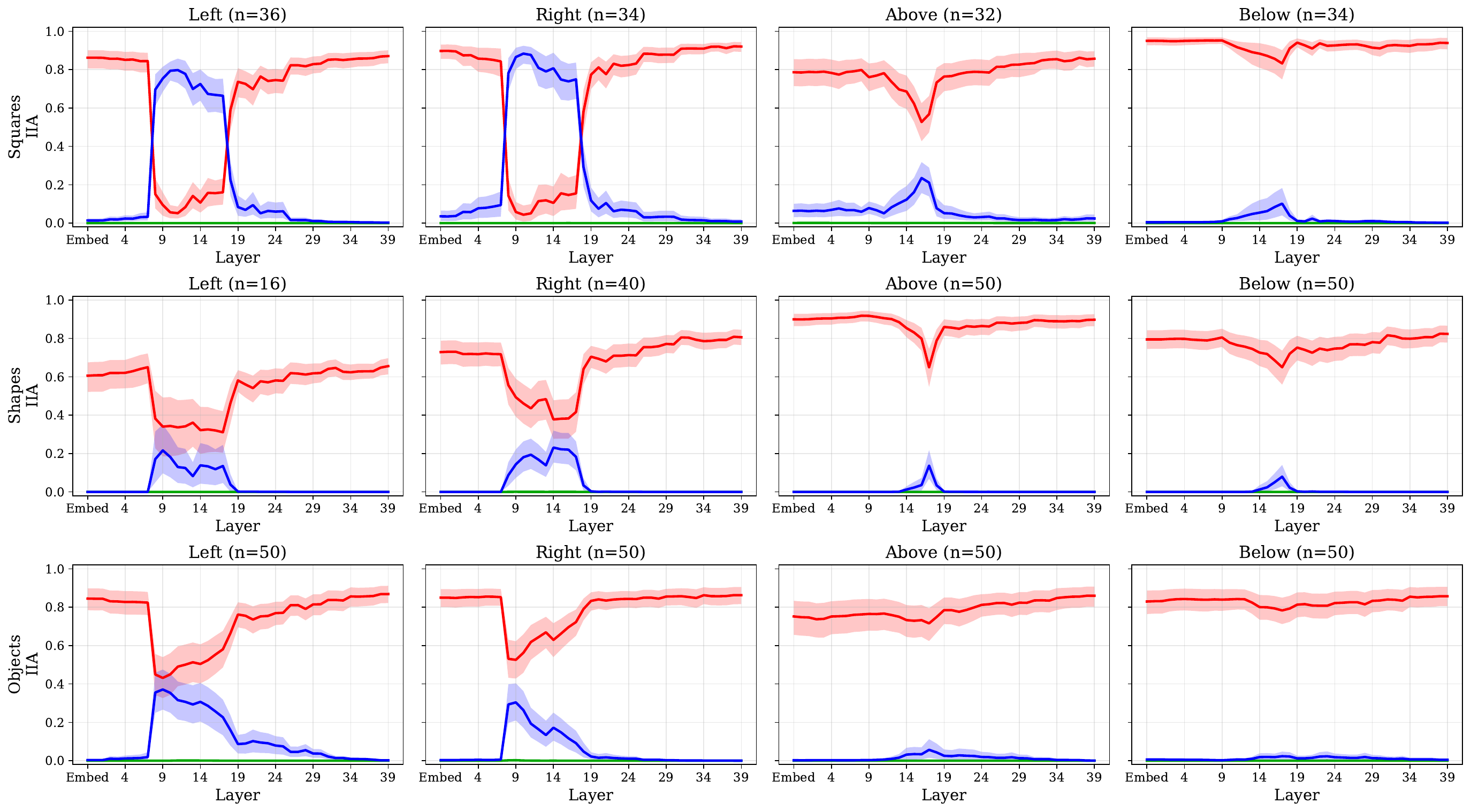}
    \caption{Visual token embeddings of square (object) tokens interchange intervention experiment on the \texttt{Pixtral-12B} model ($n{=}50$).}
    \label{fig:criss_cross_pixtral}
\end{figure}

\begin{figure}[h]
    \centering
    \includegraphics[width=1\linewidth]{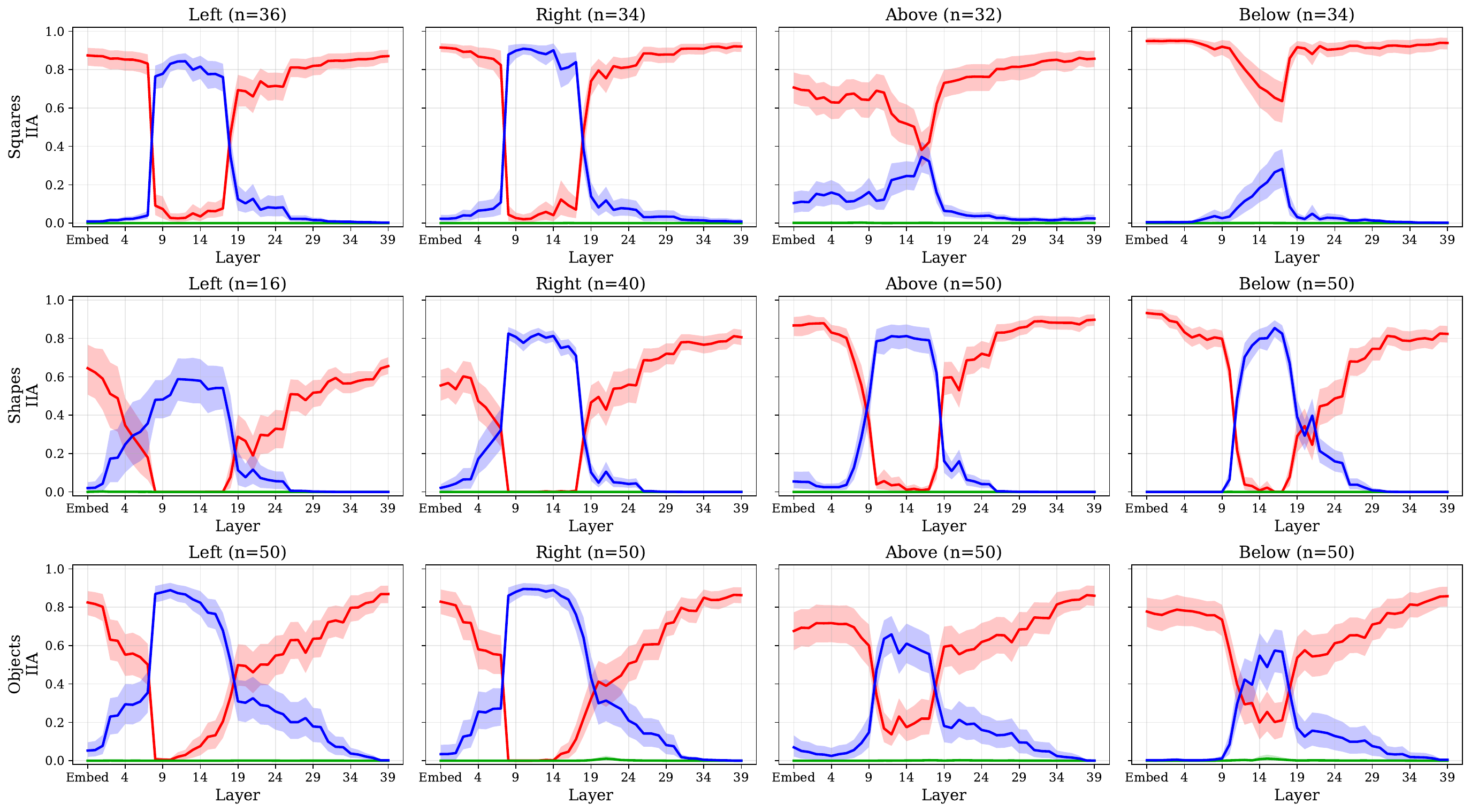}
    \caption{Visual token embeddings of strip tokens interchange intervention experiment on the \texttt{Pixtral-12B} model ($n{=}50$).}
    \label{fig:criss_cross_strip_pixtral}
\end{figure}

In this subsection, we present the results of the interchange intervention experiment described in Section~\ref{subsec:vision_encoder}. We report results for interventions that patch both the square and strip visual tokens for each model. The findings are shown in Fig.~\ref{fig:criss_cross_gemma}, Fig.~\ref{fig:criss_cross_strip_gemma}, Fig.~\ref{fig:criss_cross_square_qwen}, and Fig.~\ref{fig:criss_cross_strip_qwen}. In both cases, the results suggest that ordering information is not sufficiently encoded in the square visual token embeddings. Figs.~\ref{fig:criss_cross_qwen2vl2b}--\ref{fig:criss_cross_strip_pixtral} report the same two interventions for \texttt{Qwen2-VL-2B-Instruct}, \texttt{Qwen3-VL-8B-Instruct}, and \texttt{Pixtral-12B}. Strip patching flips the prediction in all three models while square-only patching does not, but the layer at which the flip appears differs across models, and comparing it with the vision-ablated experiment of Sec.~\ref{sec:lm_backbone_oi} distinguishes two cases.

\begin{figure}[h]
    \centering
    \includegraphics[width=1\linewidth]{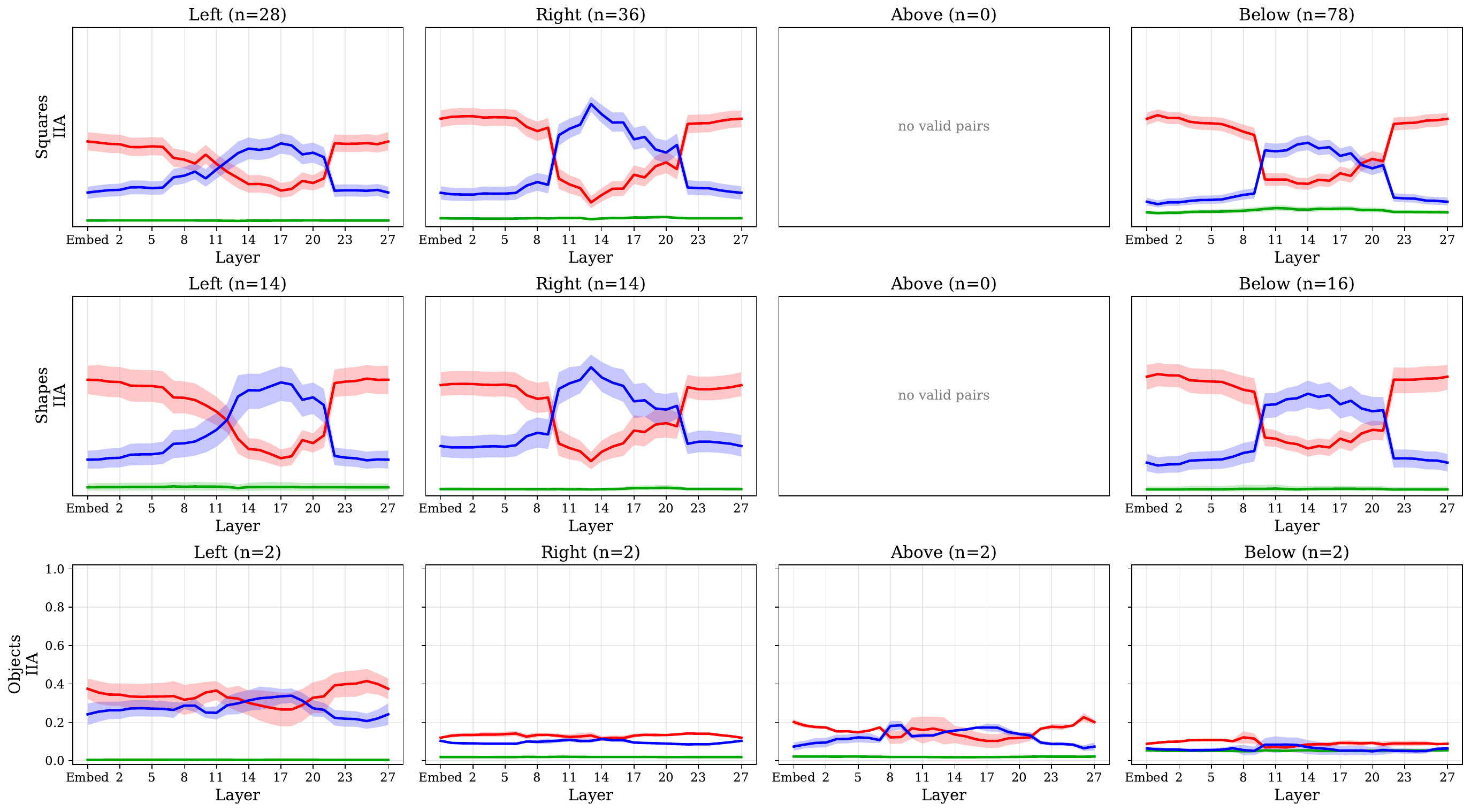}
    \vspace{1mm}
    \includegraphics[width=1\linewidth]{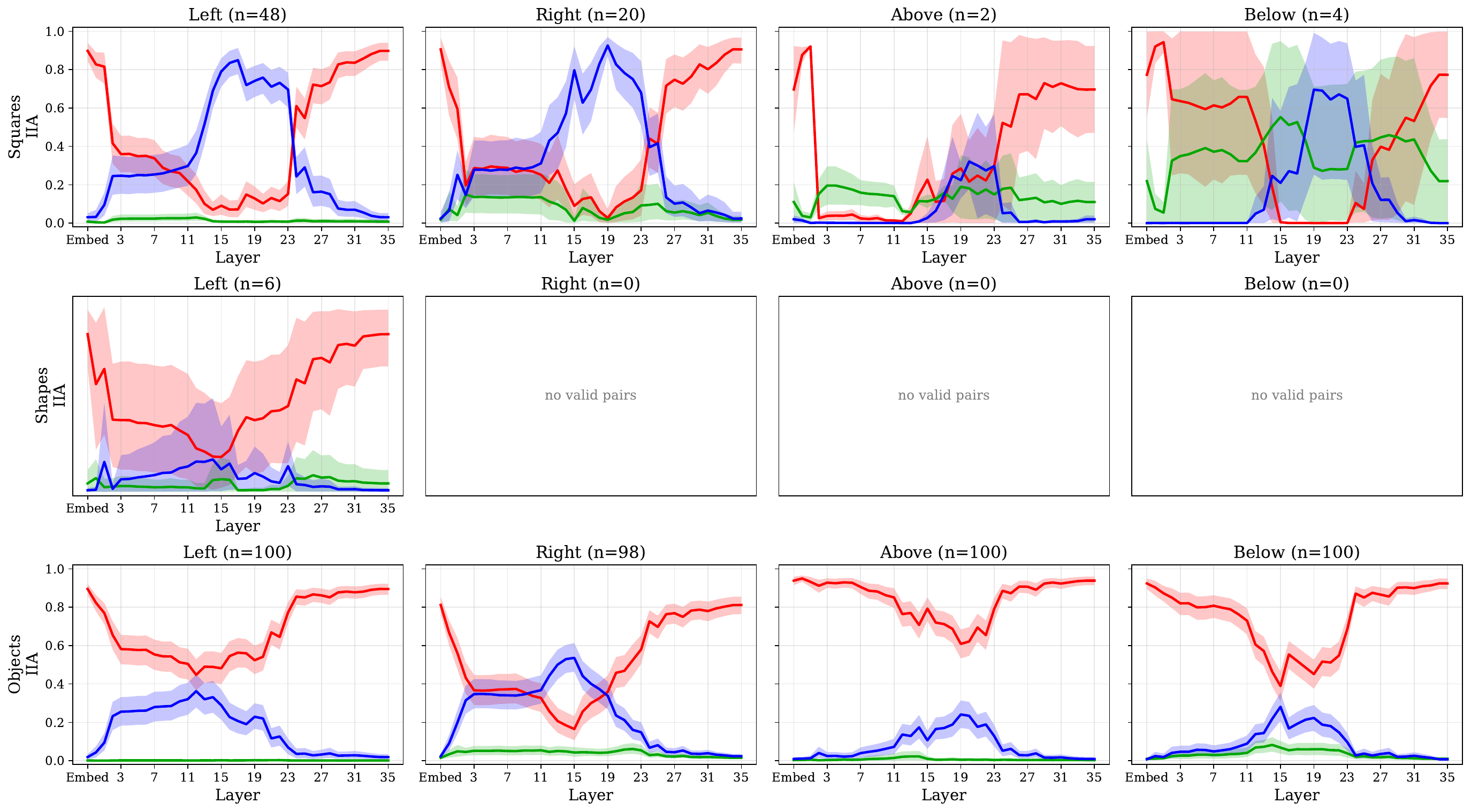}
    \caption{Square-token patching after removing vision-derived ordering information for \texttt{Qwen2-VL-2B-Instruct} (top three rows) and \texttt{Qwen3-VL-8B-Instruct} (bottom three rows). Empty panels are directions for which the ablated model answers no clean--counterfactual pair correctly.}
    \label{fig:app_strip_others}
\end{figure}

In \texttt{Qwen3-VL-8B-Instruct}, patching the projected visual embeddings alone has no effect, and the flip appears from LM layer 1 onward (Fig.~\ref{fig:criss_cross_strip_qwen3vl}). This does not indicate that the LM backbone constructs the ordering information itself: when the vision-derived ordering information is ablated (Fig.~\ref{fig:app_strip_others}, bottom rows), the LM backbone's own ordering representation flips the prediction only in layers 12--23, well after the layer-1 onset of the strip-patching flip. The ordering information therefore still originates in the vision encoder; it enters the LM backbone through its first layers rather than through the input embeddings alone, consistent with Qwen3-VL's DeepStack design, which injects intermediate vision-encoder features into the early LM layers in addition to the projected embeddings.

\begin{figure}[h]
    \centering
    \includegraphics[width=1\linewidth]{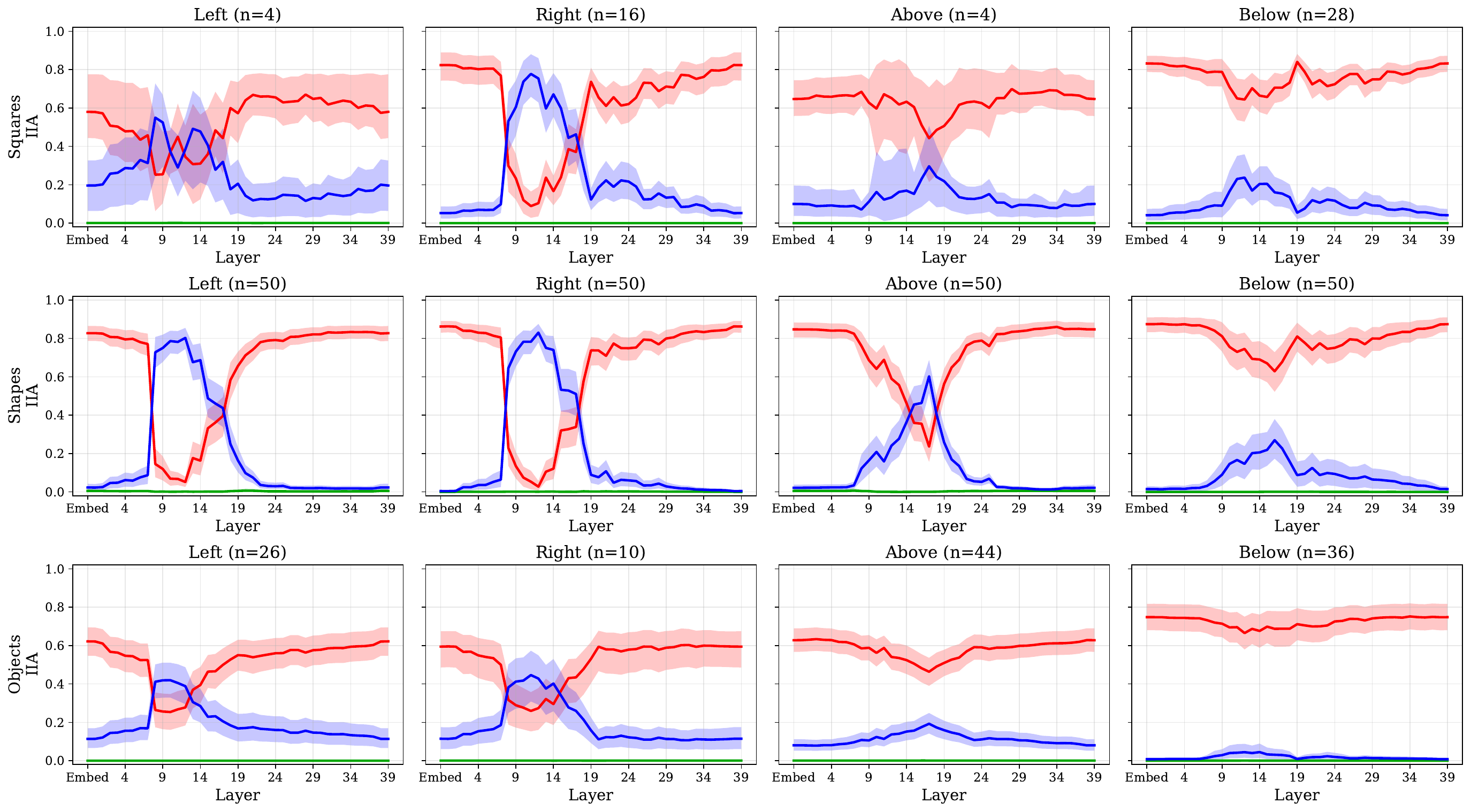}
    \caption{Square-token patching after removing vision-derived ordering information, \texttt{Pixtral-12B} ($n{=}50$ runs; the per-direction $n$ in the panel titles is capped by the accuracy of the ablated model and is very small for several cells). The flip occurs in LM layers 8--17, the same window as the strip-patching flip in Fig.~\ref{fig:criss_cross_strip_pixtral}.}
    \label{fig:app_strip_pixtral}
\end{figure}

In \texttt{Pixtral-12B}, by contrast, neither the projected embeddings nor the first seven LM layers carry a transferable ordering signal: strip patching flips the prediction only in LM layers 8--17 (Fig.~\ref{fig:criss_cross_strip_pixtral}; cleanly for the horizontal arrangements and for Shapes, weakly for the remaining vertical arrangements), and this is the same window in which square-token patching flips the prediction after the vision-side ablation (Fig.~\ref{fig:app_strip_pixtral}). In Pixtral the LM backbone thus appears to construct the ordering information itself, with little contribution from the vision encoder. This may also explain Pixtral's low accuracy on the synthetic tasks (Table~\ref{tab:behavioral_all_models}): without a vision-derived ordering signal the model has to rely on the weaker LM-side mechanism, and ablating the vision-side ordering information does not consistently reduce its accuracy (Table~\ref{tab:behavioral_inter}).

All curves are means over $n{=}100$ pairs with 95\% bootstrap confidence bands (Pixtral: original $n{=}50$ runs).

\begin{figure}
    \centering
    \includegraphics[width=1\linewidth]{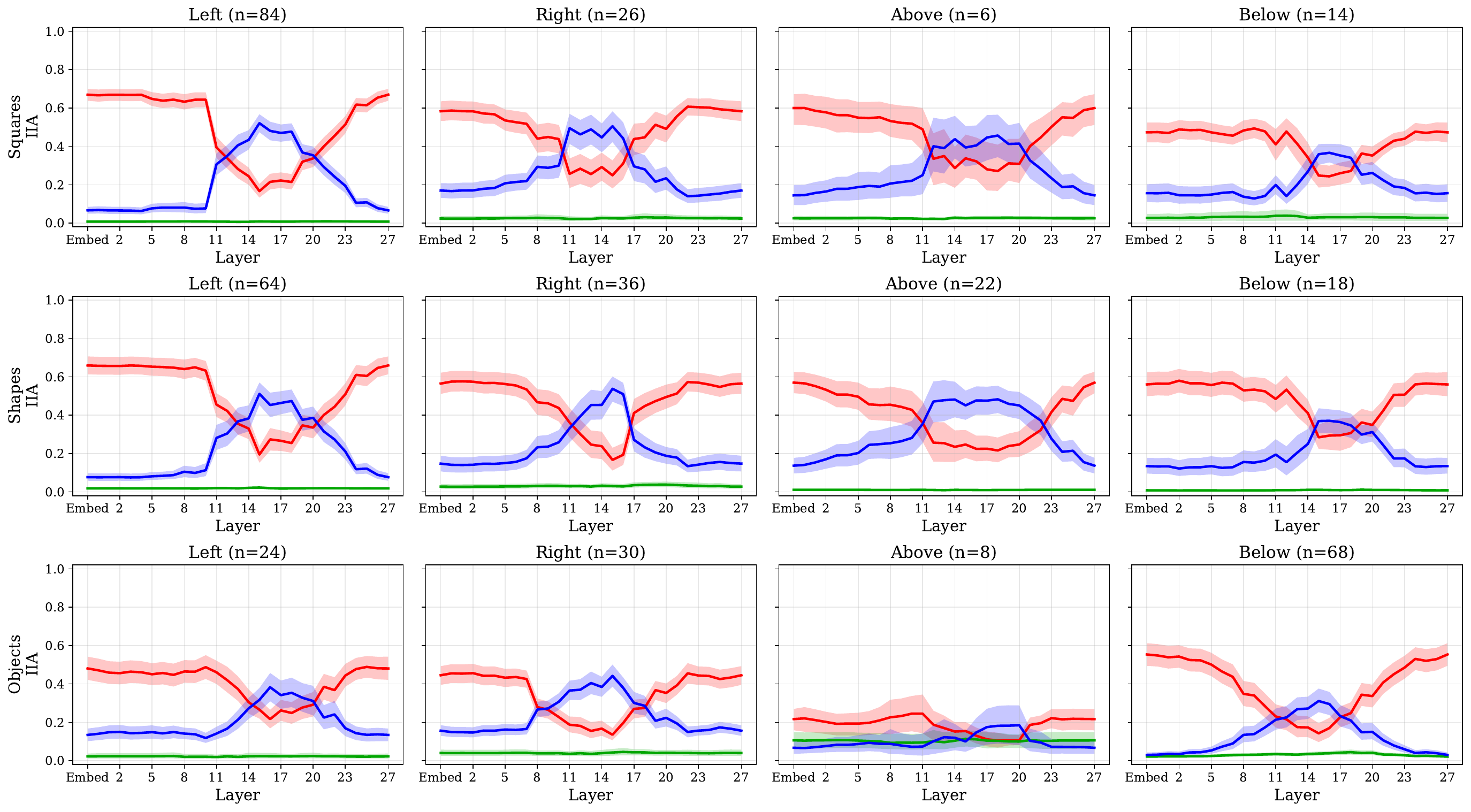}
    \caption{Square-token patching after removing vision-derived ordering information, \texttt{Qwen2-VL-7B-Instruct}.}
    \label{fig:app_strip_qwen}
\end{figure}

\begin{figure}
    \centering
    \includegraphics[width=1\linewidth]{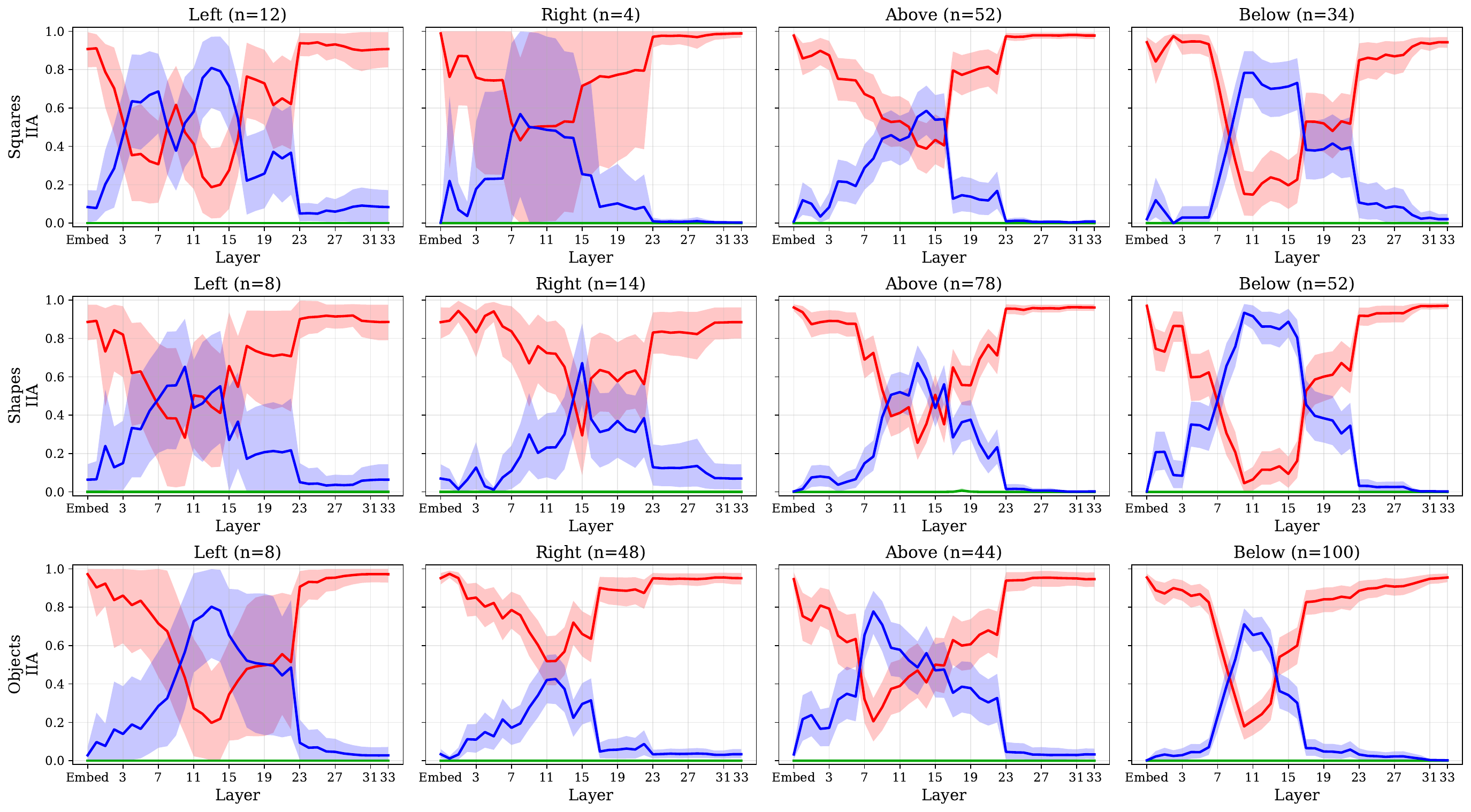}
    \caption{Square-token patching after removing vision-derived ordering information, \texttt{Gemma-3-4b-it}.}
    \label{fig:app_strip_gemma}
\end{figure}

In Fig. ~\ref{fig:app_strip_gemma} and Fig.~\ref{fig:app_strip_qwen}, we present additional patching results after removing vision encoder-generated ordering information from visual embeddings using the procedure from Section~\ref{subsec:removing_info}. Output switching occurs in intermediate layers, corroborating our finding that the LM backbone produces its own ordering information. Figs.~\ref{fig:app_strip_others} and~\ref{fig:app_strip_pixtral} show the same experiment for the remaining models. Under the ablation, the per-direction $n$ (panel titles) is capped by the accuracy of the ablated model, so several cells have $n \ll 100$ and correspondingly wide confidence bands; cells with no valid pairs are left empty.   

Results are omitted when sufficient clean/counterfactual input pairs cannot be generated. 

\clearpage

\clearpage
\clearpage
\clearpage
\clearpage

\clearpage

\subsection{Sample Size and Statistical Significance}
\label{app:n100_stats}
All interchange-intervention curves in this paper use $n{=}100$ clean--counterfactual pairs per direction and setting, with 95\% percentile-bootstrap confidence bands ($B{=}2000$ resamples over pairs). An earlier version of these experiments used $n{=}50$ pairs. Because pairs are kept only when the model answers \emph{both} the clean and the counterfactual prompt correctly, the achievable $n$ is bounded by model accuracy; the ordering-ablation experiment of Sec.~\ref{sec:lm_backbone_oi} and several Pixtral cells therefore have $n<100$ (reported in panel titles).

To check that the sample size does not change the conclusions, we compared the $n{=}50$ and $n{=}100$ estimates cell by cell. For each (experiment, model, setting, direction) we take the peak-layer effect, $P(\text{counterfactual answer}) - P(\text{clean answer})$ at the layer where it is maximal, and test whether it differs between the two sample sizes with a two-sided permutation test ($10^4$ shuffles). Where the $n{=}100$ pairs are a superset of the $n{=}50$ pairs, we test the original 50 against the 50 added pairs; otherwise the two samples are treated as independent draws. Table~\ref{tab:n50_vs_n100} summarizes the 132 direction cells (119 testable; 13 ablation cells admitted no new pairs). The median absolute change in the peak effect is $0.010$; 5 of 119 tests reach $p<0.05$, matching the $\approx 6$ expected by chance, none survives Benjamini--Hochberg correction, and 129 of 132 cells have overlapping 95\% confidence intervals. The $n{=}100$ results thus reproduce the $n{=}50$ results and merely tighten the intervals.

\begin{table}[h]
\centering
\small
\caption{$n{=}50$ vs.\ $n{=}100$ peak-layer effects. $|\Delta|$: absolute change in the peak effect; $p<.05$: cells with an uncorrected permutation $p<0.05$; $q<.05$: cells significant after Benjamini--Hochberg correction across all 119 tests; overlap: cells whose 95\% CIs overlap.}
\label{tab:n50_vs_n100}
\vspace{1mm}
\resizebox{\linewidth}{!}{%
\begin{tabular}{llccccc}
\toprule
Experiment & Model & Cells (tested) & Median $|\Delta|$ & Max $|\Delta|$ & $p<.05$ / $q<.05$ & CI overlap \\
\midrule
Last-token patching (Fig.~\ref{fig:last_token_results}) & Qwen2-VL-7B & 14 (14) & 0.009 & 0.018 & 1 / 0 & 14/14 \\
                                                          & Gemma-3-4b  & 14 (14) & 0.003 & 0.013 & 0 / 0 & 13/14 \\
Object patching (Fig.~\ref{fig:obj_patching})            & Qwen2-VL-7B & 14 (14) & 0.044 & 0.190 & 0 / 0 & 14/14 \\
                                                          & Gemma-3-4b  & 14 (14) & 0.061 & 0.234 & 2 / 0 & 13/14 \\
Strip patching (Fig.~\ref{fig:strip_patching})           & Qwen2-VL-7B & 14 (14) & 0.005 & 0.041 & 1 / 0 & 13/14 \\
                                                          & Gemma-3-4b  & 14 (14) & 0.012 & 0.119 & 0 / 0 & 14/14 \\
Ablated-vision patching (Fig.~\ref{fig:secondary_mech})  & Qwen2-VL-7B & 12 (5)  & 0.000 & 0.027 & 0 / 0 & 12/12 \\
                                                          & Gemma-3-4b  & 12 (6)  & 0.001 & 0.140 & 1 / 0 & 12/12 \\
Absolute vs.\ relative (App.~\ref{app:abs_vs_ord})       & Qwen2-VL-7B & 12 (12) & 0.007 & 0.079 & 0 / 0 & 12/12 \\
                                                          & Gemma-3-4b  & 12 (12) & 0.032 & 0.117 & 0 / 0 & 12/12 \\
\midrule
All & & 132 (119) & 0.010 & 0.234 & 5 / 0 & 129/132 \\
\bottomrule
\end{tabular}
}
\end{table}
\clearpage

\subsection{Diagonal Arrangements}
\label{app:diagonal}
\begin{figure}[h]
    \centering
    \includegraphics[width=1\linewidth]{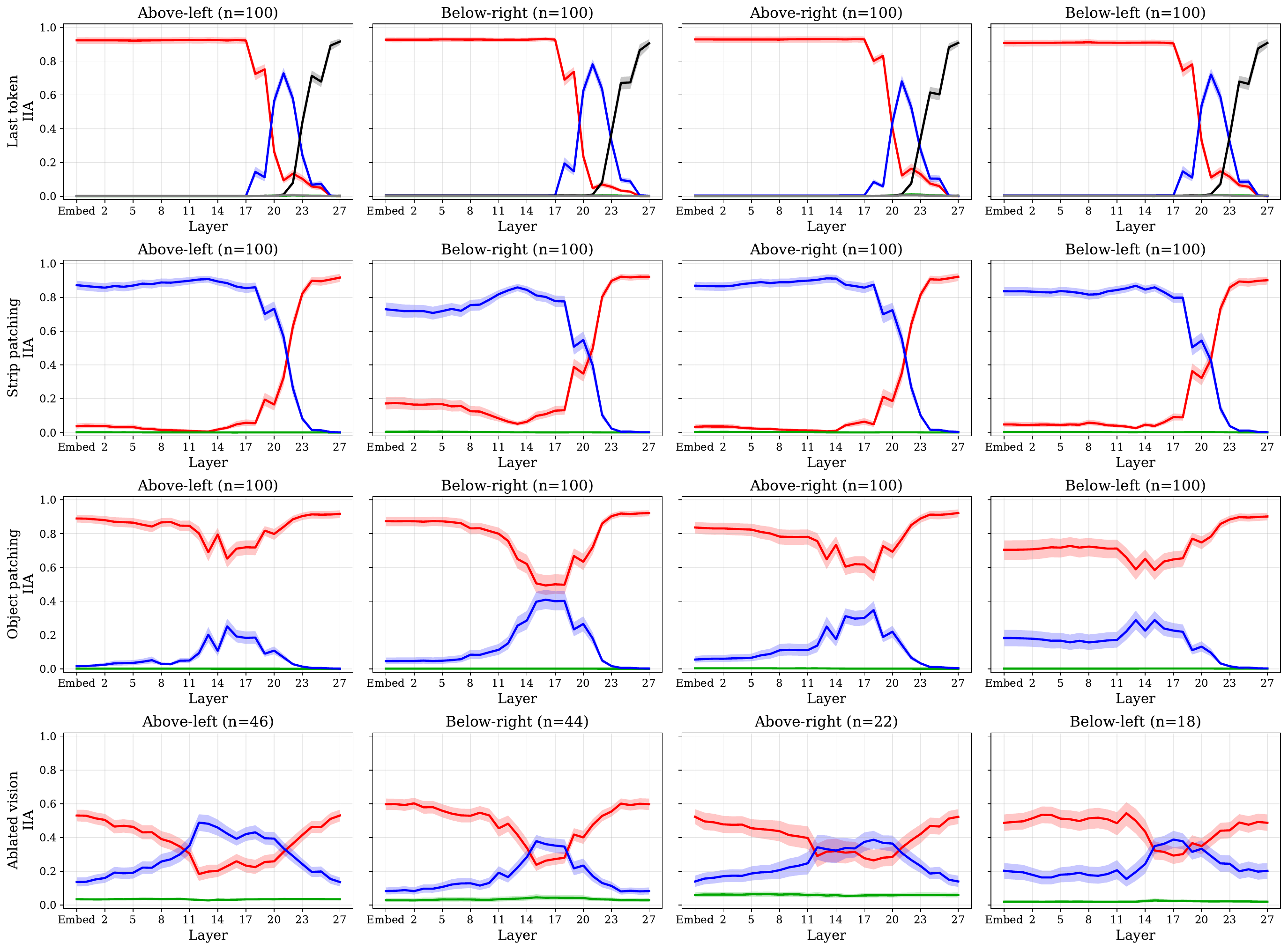}
    \caption{\textbf{Diagonal arrangements, \texttt{Qwen2-VL-7B-Instruct}} ($n{=}100$ pairs, 95\% bootstrap CIs). Rows: last-token patching (colors as in Fig.~\ref{fig:last_token_results}), strip patching, square patching, and square patching after the vision-ordering ablation (colors as in Fig.~\ref{fig:strip_patching}). Columns: the two principal-diagonal relations followed by the two secondary-diagonal relations.}
    \label{fig:diagonal}
\end{figure}

\begin{figure}[h]
    \centering
    \includegraphics[width=0.49\linewidth]{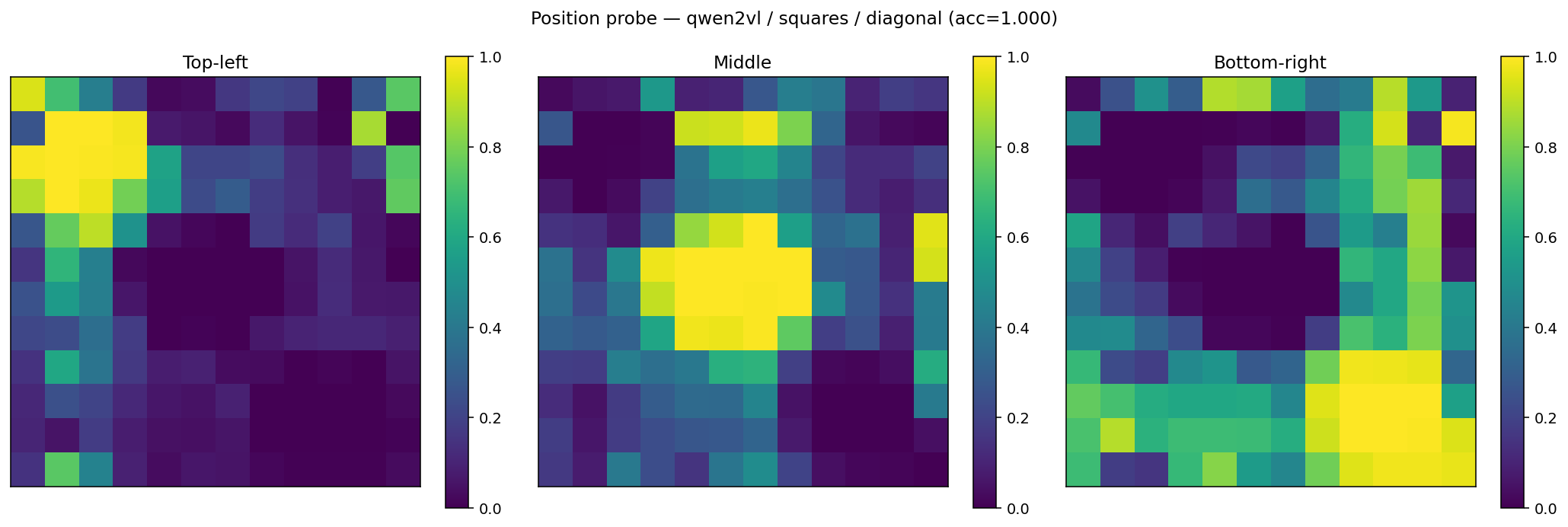}
    \includegraphics[width=0.49\linewidth]{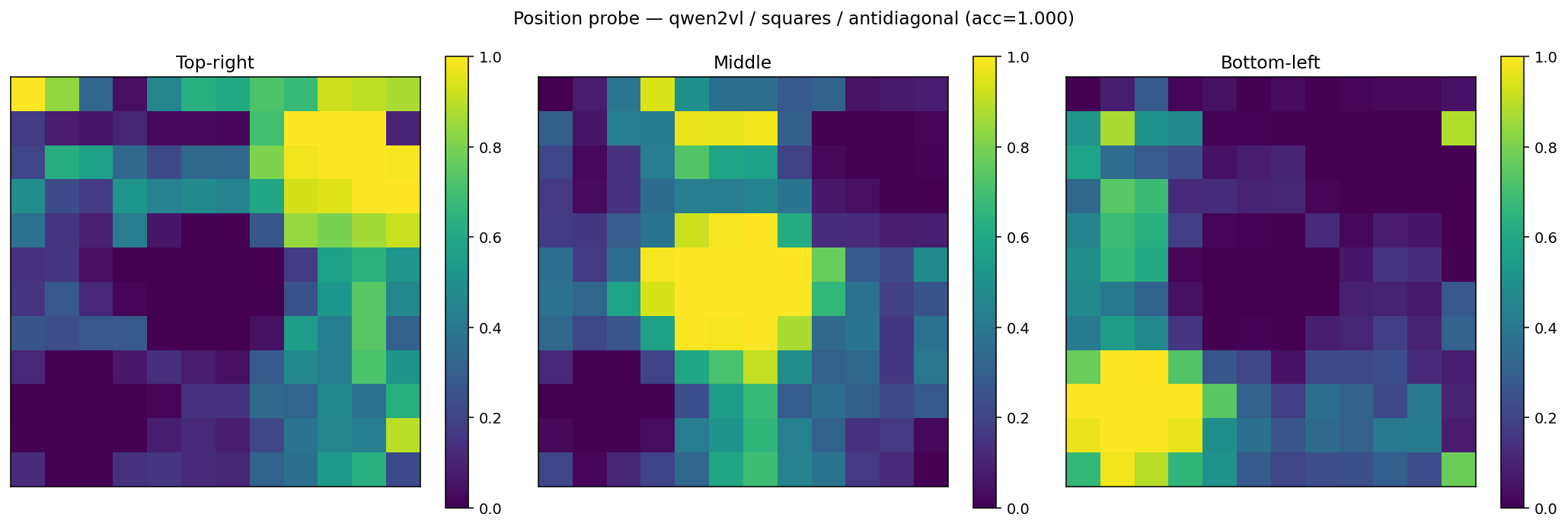}
    \caption{Position-probe predictions on the diagonal (left) and anti-diagonal (right) Squares settings for \texttt{Qwen2-VL-7B-Instruct}. As in Fig.~\ref{fig:probs}, probes trained only on square tokens generalize to background tokens along diagonal strips.}
    \label{fig:diagonal_probes}
\end{figure}

The main text studies axis-aligned relations (left/right, above/below). To test whether the same mechanism handles relations that are not aligned with the token grid, we construct two additional Squares settings in which the three squares lie on the principal (top-left to bottom-right) or the secondary (top-right to bottom-left) diagonal of the image, and query them with compound relations (``above and to the left of'', ``below and to the right of'', ``above and to the right of'', ``below and to the left of''). Strips are defined analogously to the axis-aligned case as diagonal bands (the set of tokens with a fixed value of row$+$column, or row$-$column, that pass through a square). We run all four causal experiments on \texttt{Qwen2-VL-7B-Instruct} with $n{=}100$ pairs (Fig.~\ref{fig:diagonal}) and train the position probes of Sec.~\ref{subsec:probing} on the diagonal settings (Fig.~\ref{fig:diagonal_probes}).

The results mirror the axis-aligned settings. Last-token patching shows the same three-phase pattern with an ordering phase around layers 20--22; strip patching flips the prediction from the vision embedding through layer $\approx$20; square-only patching has a weak effect, as it does for the axis-aligned Squares setting on this model (Fig.~\ref{fig:obj_patching}); and, after ablating the vision-derived ordering information, square-token patching flips the prediction in middle layers, with $n$ reduced to 18--46 by the accuracy of the ablated model. The diagonal probes reach $100\%$ test accuracy, and their predictions again spread along diagonal strips through the squares. The ordering mechanism is therefore not tied to the horizontal and vertical axes of the token grid.

\clearpage
\subsection{How Are Spatial Positions Represented?}
\label{sec:relative_vs_absolute}

\subsubsection{Experimental Settings}

Building on the synthetic Squares setting described in Sec.~\ref{subsec:data}, we construct two layouts, each containing three equal-sized colored squares with fixed spacing. The two layouts differ only by a shift: one layout is shifted left and the other is shifted right (or up and down), as illustrated in Fig.~\ref{fig:relative_absolute_probe}. The layouts are designed such that the rightmost square in the left-shifted layout occupies the same absolute image location as the leftmost square in the right-shifted layout. We refer to the square occupying this shared location as the \emph{overlapping square}.

We label the squares according to their ordinal position within the layout: left ($0$), middle ($1$), and right ($2$). Under this labeling scheme, the overlapping square receives label $2$ in the left-shifted layout and label $0$ in the right-shifted layout. Thus, the same absolute image location is associated with different ordinal labels depending on the layout.

This construction allows us to distinguish ordinal ordering information from absolute-position information. In particular, because the overlapping square appears at the same absolute image location in both layouts but receives different ordinal labels, its label cannot be determined from absolute position alone. The overlapping square therefore serves as the critical test case for whether vision-encoder representations encode ordinal position within the layout, rather than only fixed image coordinates.

\subsubsection{Absolute Position vs. Ordinal Ordering in the Vision Encoder}
\label{app:abs_vs_ord}

\paragraph{Probing Vision-Encoder Layers.}
Following the probing procedure described in Sec.~\ref{subsec:probing}, we train a separate linear probe on the hidden representations produced at each vision-encoder layer in \texttt{Gemma-3-4b-it} and \texttt{Qwen2-VL-7B-Instruct}. Each probe receives the hidden representation of a visual token corresponding to one of the squares and predicts the square's ordinal position: left, middle, or right. In the extracted vision-encoder hidden states, each square corresponds to a $7 \times 7$ token region for Gemma and a $4 \times 4$ token region for Qwen.

\paragraph{Probing shows the presence of ordinal ordering information.}

\begin{figure}[!htbp]
    \centering
    \includegraphics[width=0.72\textwidth]
    {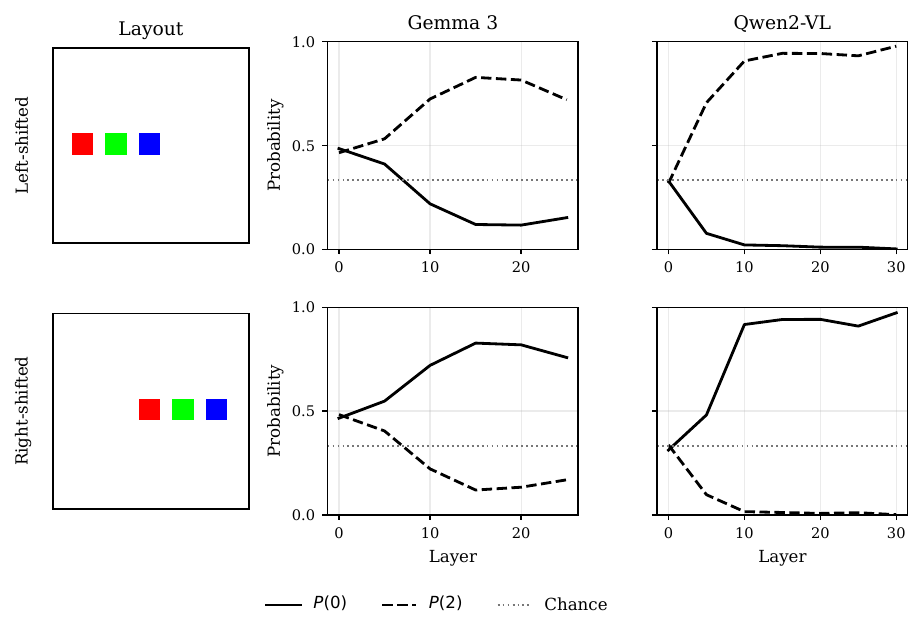}
    \caption{\textbf{Training a single probe for both layouts learns ordinal order information.}
    The overlapping square appears at the same absolute image location in both layouts, while its ordinal label changes from $2$ in the left-shifted layout to $0$ in the right-shifted layout. We plot the probabilities assigned to the two candidate labels, $P(0)$ (solid) and $P(2)$ (dashed). For this square only, chance is shown as a dotted line. The correct curve is dashed in the top row and solid in the bottom row. Although both squares occupy the same absolute image location, the probe correctly distinguishes their ordinal positions, demonstrating that ordinal information is linearly decodable.}
    \label{fig:relative_absolute_probe}
\end{figure}

We use 90 images for training and 30 held-out images for evaluation, generated from the square-color assignments described in Sec.~\ref{subsec:data}. Each color assignment is rendered in both layouts, and a single three-way linear probe is trained jointly on visual-token representations from both layouts. This joint training ensures that the shared absolute location receives conflicting ordinal labels across the two layouts. For instance, consider the left and right-shifted configurations of the squares depicted in Fig.~\ref{fig:relative_absolute_probe}. 
Because the overlapping location is assigned two different ordinal labels during training, any classifier that relies solely on absolute image coordinates must fail. Successful prediction therefore requires information beyond absolute position.

We evaluate this distinction by reporting, for the overlapping square, the probabilities assigned to its two possible ordinal labels, $P(0)$ and $P(2)$, separately for each layout. The probabilities are averaged over the visual tokens corresponding to the overlapping square and across all held-out images. For completeness, we also report overall probe accuracies over all visual tokens corresponding to each of the squares in Table~\ref{tab:overlap_probe_accuracy}.

As shown in Fig.~\ref{fig:relative_absolute_probe}, the probability assigned to the correct layout-dependent label increases across vision-encoder layers in both models. 
These results show that ordinal ordering information
is linearly decodable from the vision-encoder representations of both models.

\begin{table}[!htbp]
    \centering
    \small
    \setlength{\tabcolsep}{5pt}
    \caption{\textbf{Overall probe accuracy by layout.}
    Test accuracy is computed over all visual tokens corresponding to the squares. Mean averages the two layouts.}
    \label{tab:overlap_probe_accuracy}
    \vspace{2mm}
    \begin{tabular}{c ccc ccc}
        \toprule
        & \multicolumn{3}{c}{\textbf{Gemma 3}}
        & \multicolumn{3}{c}{\textbf{Qwen2-VL}} \\
        \cmidrule(lr){2-4}\cmidrule(lr){5-7}
        Layer & Left & Right & Mean
              & Left & Right & Mean \\
        \midrule
        0  & 84.4 & 77.3 & 80.8 & 25.8  & 19.5 & 22.6 \\
        5  & 91.6 & 93.1 & 92.4 & 81.3  & 81.3 & 81.3 \\
        10 & 99.6 & 98.9 & 99.3 & 99.9  & 99.9 & 99.9 \\
        15 & 99.7 & 99.5 & 99.6 & 99.1  & 99.8 & 99.4 \\
        20 & 97.8 & 98.2 & 98.0 & 98.7  & 99.7 & 99.2 \\
        25 & 93.9 & 95.6 & 94.7 & 98.4  & 98.5 & 98.4 \\
        30 & --   & --   & --   & 100.0 & 99.9 & 99.9 \\
        \bottomrule
    \end{tabular}
\end{table}

\begin{figure}[!htbp]
    \centering
    \includegraphics[width=0.72\textwidth]
    {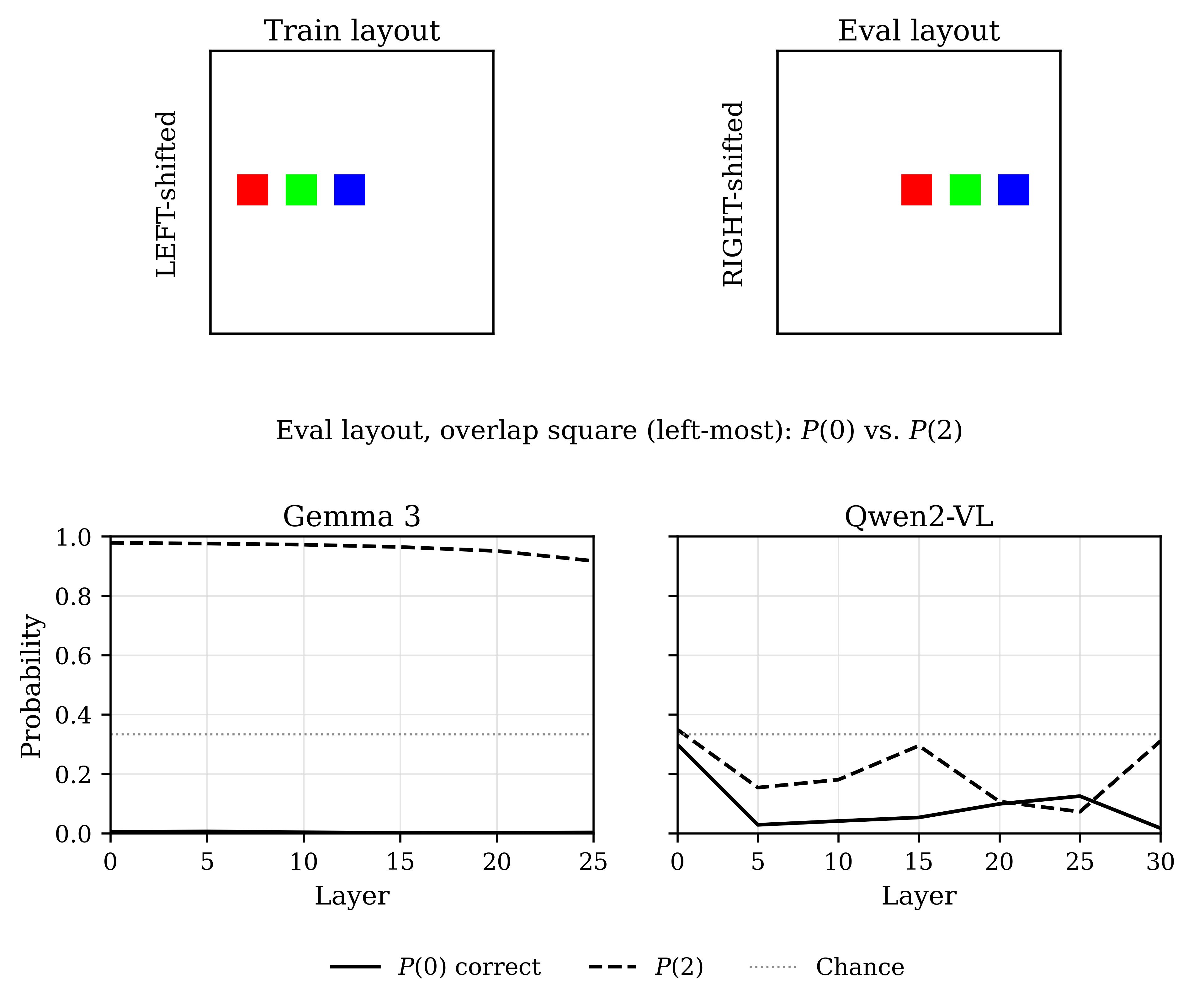}
    \caption{\textbf{Training a probe on one layout does not generalize to the other, suggesting the presence of absolute position information.}
    The overlapping square appears at the same absolute image location in both layouts, while its ordinal label changes from $2$ in the left-shifted layout to $0$ in the right-shifted layout. We plot the probabilities assigned to the two candidate labels, $P(0)$ (solid) and $P(2)$ (dashed). For this square only, chance is shown as a dotted line. A probe trained on one layout fails to correctly classify the overlapping square in the other layout, indicating that it relies on absolute image coordinates rather than purely ordinal position.}
    \label{fig:relative_absolute_probe2}
\end{figure}

\paragraph{Failure to Generalize Across Layouts Reveals Absolute Position Information.} We repeat the same experiment; however, this time we train the probe on one layout (e.g., the left-shifted) and test it on the other layout (e.g., right-shifted). We had a total of 90 training examples and 30 eval examples. Results are presented in Fig.~\ref{fig:relative_absolute_probe2}. If the representation encoded only ordinal position, a probe trained on one layout would generalize perfectly to the other. The observed failure therefore indicates that the probe relies on absolute image coordinates that do not transfer across layouts. Table~\ref{tab:cross_layout_train_test_accuracy} reports the corresponding train and test accuracies.

\begin{table}[!htbp]
    \centering
    \small
    \setlength{\tabcolsep}{5pt}
    \caption{\textbf{Cross-layout probe accuracy across layers.}
    The probe is trained on square tokens from the left-shifted layout and evaluated on square tokens from the right-shifted layout.}
    \label{tab:cross_layout_train_test_accuracy}
    \vspace{2mm}
    \begin{tabular}{c cc cc}
        \toprule
        & \multicolumn{2}{c}{\textbf{Gemma 3}}
        & \multicolumn{2}{c}{\textbf{Qwen2-VL}} \\
        \cmidrule(lr){2-3}\cmidrule(lr){4-5}
        Layer & Train & Test & Train & Test \\
        \midrule
        0  & 100.0 & 0.0  & 51.0  & 17.0 \\
        5  & 100.0 & 1.0  & 100.0 & 48.0 \\
        10 & 100.0 & 2.0  & 100.0 & 49.0 \\
        15 & 100.0 & 10.0 & 100.0 & 36.0 \\
        20 & 100.0 & 8.0  & 100.0 & 59.0 \\
        25 & 100.0 & 9.0  & 100.0 & 60.0 \\
        30 & --    & --   & 100.0 & 33.0 \\
        \bottomrule
    \end{tabular}
\end{table}

\begin{figure}[h]
    \centering
    \includegraphics[width=0.49\linewidth]{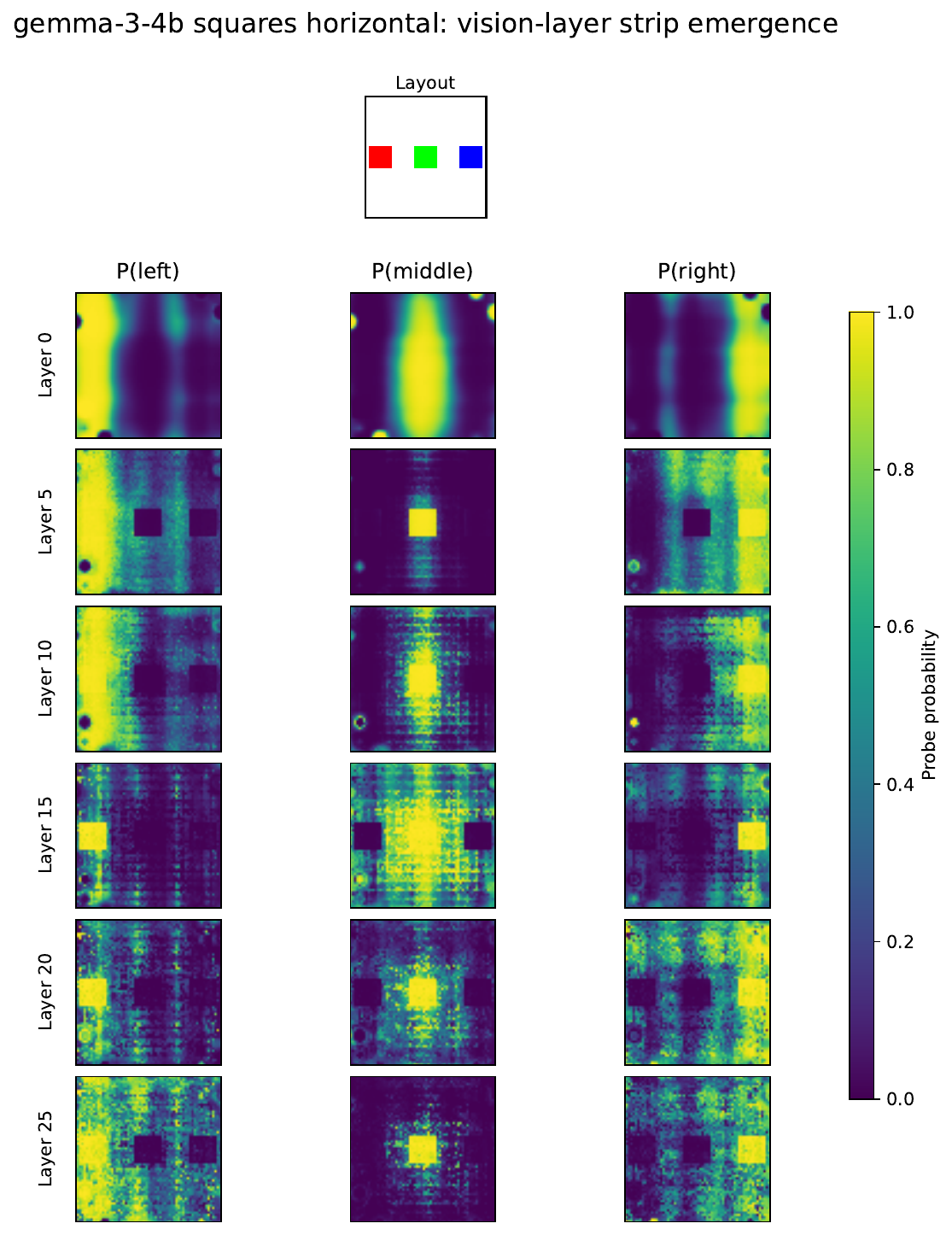}
    \includegraphics[width=0.49\linewidth]{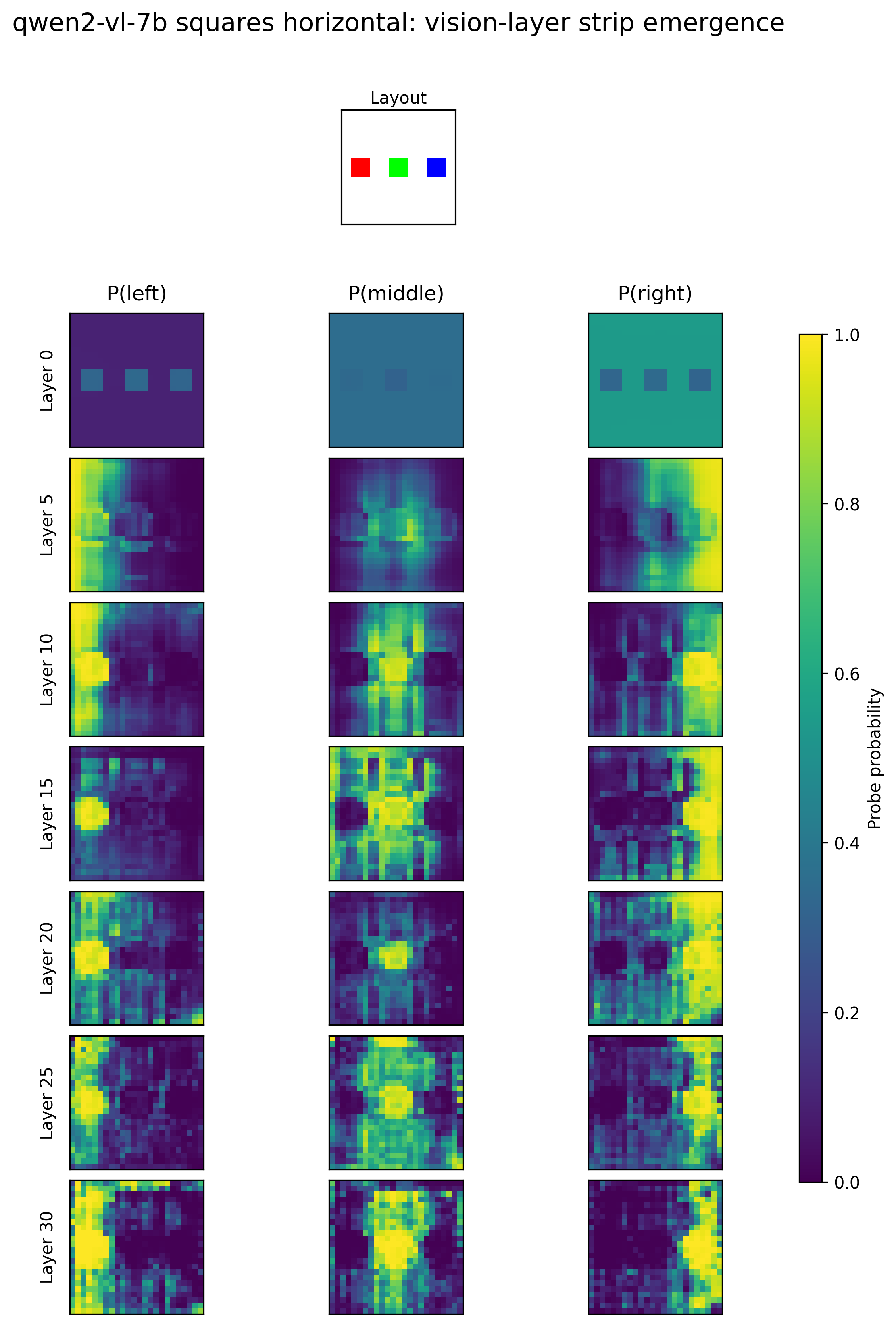}
    \caption{\textbf{Strip emergence across vision-encoder layers} (Squares, horizontal). Each row applies the probe trained at that vision-encoder layer to all tokens; columns are the three position labels. Left: \texttt{Gemma-3-4b-it}, where the strips are present at layer 0. Right: \texttt{Qwen2-VL-7B-Instruct}, where layer 0 carries no position information and the strips form over the first five to ten layers.}
    \label{fig:strip_emergence}
\end{figure}

\begin{figure}[t]
    \centering
    \begin{subfigure}{1\linewidth}
        \centering
        \includegraphics[width=\linewidth]{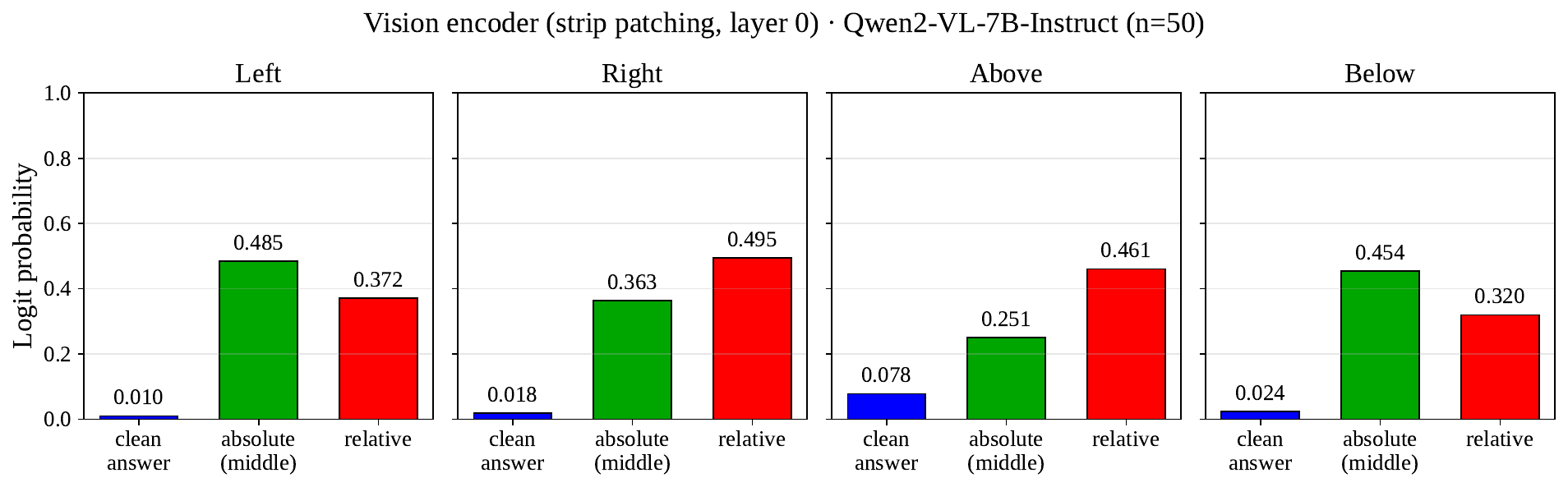}
    \end{subfigure}
    \vspace{0.5em}
    \begin{subfigure}{1\linewidth}
        \centering
        \includegraphics[width=\linewidth]{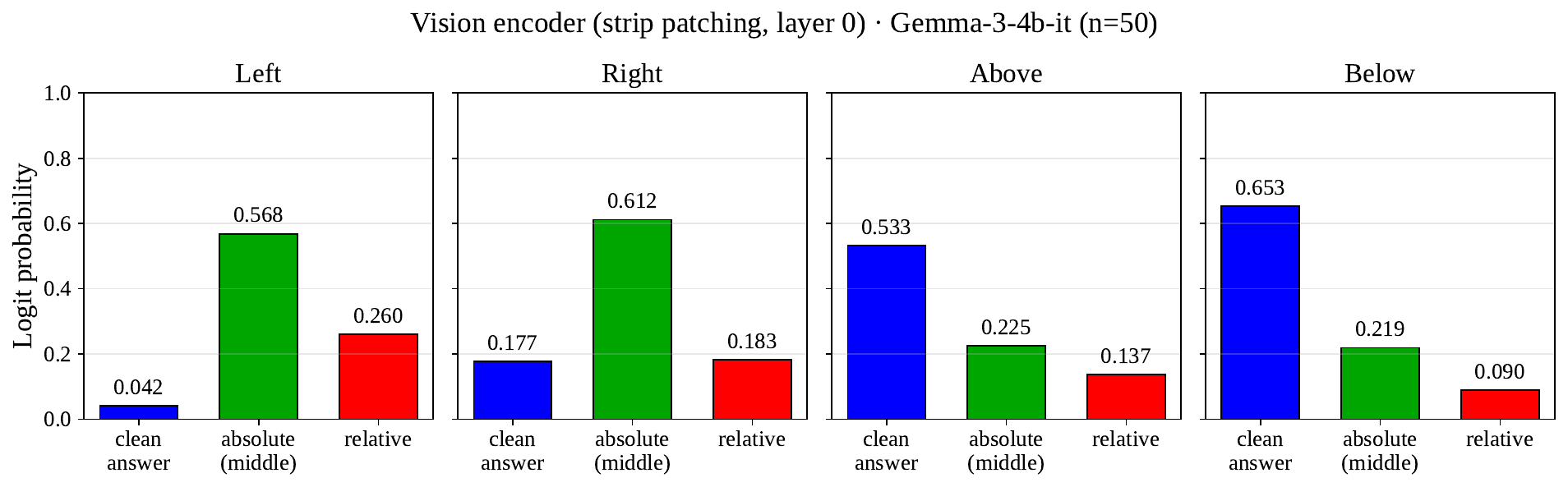}
    \end{subfigure}
    \vspace{0.5em}
    \caption{\textbf{Label probabilities after the intervention}: Intervening on the visual token embeddings generated by the vision encoder using the mirrored counterfactual produces substantial label probabilities for both the absolute and relative expected outputs.}
    \label{fig:rel_vs_abs_ve}
\end{figure}

\clearpage
\paragraph{When do the strips emerge?}
\label{app:strip_emergence}
The strip-like probe patterns of Fig.~\ref{fig:probs} are measured at the output of the vision encoder and projector. To locate where they arise, we train a separate three-way position probe on the object tokens at each vision-encoder layer (Squares setting; 90 training and 30 held-out color permutations) and apply it to all tokens (Fig.~\ref{fig:strip_emergence}, Table~\ref{tab:strip_emergence}). The two models differ sharply in timing. In \texttt{Gemma-3-4b-it}, whose SigLIP encoder adds \emph{learned absolute position embeddings} to the patch embeddings at the input, the layer-0 probe already reaches $100\%$ held-out accuracy and assigns probability $0.90$ to the correct label on the background tokens of the object's column, versus $0.20$ elsewhere: the strips are present before any attention layer. In \texttt{Qwen2-VL-7B-Instruct}, whose encoder has no additive position embedding and injects position only through \emph{2D rotary embeddings} inside attention, the layer-0 probe is at chance ($29\%$) and shows no spatial structure; ordering information and the strips emerge over the first five layers and saturate by layer 10. The overlap-probe accuracies in Table~\ref{tab:overlap_probe_accuracy} (Gemma $80.8\%$ vs.\ Qwen $22.6\%$ at layer 0) show the same timing. This is consistent with the strips being a by-product of how position is made available to attention: with absolute position embeddings, every token carries its own coordinates from the input, and tokens that share a row or column coordinate share a component of their embedding; with 2D-RoPE, attention between tokens depends on their relative row and column offsets, so tokens aligned with an object along one axis are the ones that attend to it most similarly, and position information is propagated along rows and columns as attention layers accumulate. In both cases the resulting representation is causally used by the LM (Sec.~\ref{subsec:vision_encoder}) and combines absolute and relative components (Fig.~\ref{fig:rel_vs_abs_ve}).

\begin{table}[h]
\centering
\small
\caption{Layer-wise vision-encoder probing on the Squares setting (horizontal arrangement; the vertical arrangement behaves the same). A three-way probe is trained on square tokens at each vision-encoder layer. ``Square'', ``Strip'', and ``Outside'' give the mean probability the probe assigns to a label on that label's square tokens, on the background tokens of the same column excluding the square, and on all other tokens, averaged over the three labels and the held-out images.}
\label{tab:strip_emergence}
\vspace{1mm}
\resizebox{\linewidth}{!}{%
\begin{tabular}{c cccc cccc}
\toprule
& \multicolumn{4}{c}{\textbf{Gemma-3-4b-it} (SigLIP, absolute pos.\ emb.)} & \multicolumn{4}{c}{\textbf{Qwen2-VL-7B} (2D-RoPE)} \\
\cmidrule(lr){2-5}\cmidrule(lr){6-9}
Layer & Held-out acc. & Square & Strip & Outside & Held-out acc. & Square & Strip & Outside \\
\midrule
0  & 100.0 & 0.97 & 0.90 & 0.20 & 29.4  & 0.32 & 0.33 & 0.33 \\
5  & 100.0 & 0.98 & 0.67 & 0.24 & 91.3  & 0.68 & 0.65 & 0.27 \\
10 & 100.0 & 0.98 & 0.76 & 0.23 & 100.0 & 0.96 & 0.74 & 0.25 \\
\bottomrule
\end{tabular}
}
\end{table}

\clearpage
\paragraph{Patching Vision Embeddings}

\begin{figure}[h]
    \centering
    \setlength{\fboxsep}{0pt}
    \setlength{\fboxrule}{1.5pt}
    \begin{minipage}{0.59\linewidth}
    \resizebox{\linewidth}{!}{%
    \begin{tabular}{cc}
        Clean & Counterfactual \\

        \tikzmarknode{box2}{\fcolorbox{gray}{white}{\includegraphics[width=0.3\linewidth]{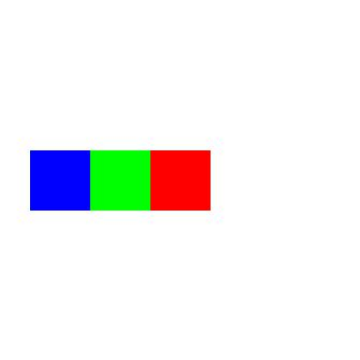}}} &
        \fcolorbox{gray}{white}{\includegraphics[width=0.3\linewidth]{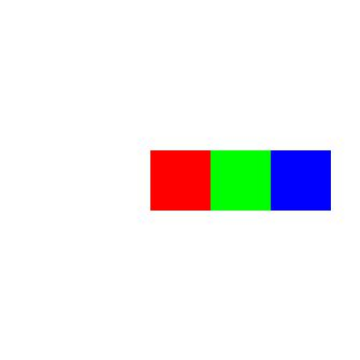}} \\
        \small\parbox{4cm}{\centering\textit{\vspace{1mm} The color of the \textbf{leftmost} square is}}
        &
        \small\parbox{4cm}{\centering\textit{\vspace{1mm} The color of the \textbf{rightmost} square is}} \\[3mm]

        \tikzmarknode{box}{\fcolorbox{gray}{white}{\includegraphics[width=0.3\linewidth]{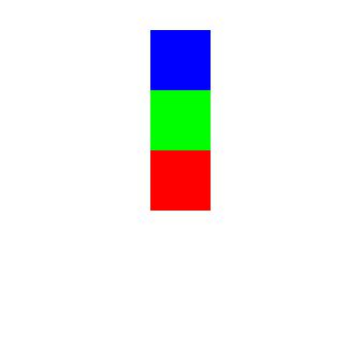}}} &
        \fcolorbox{gray}{white}{\includegraphics[width=0.3\linewidth]{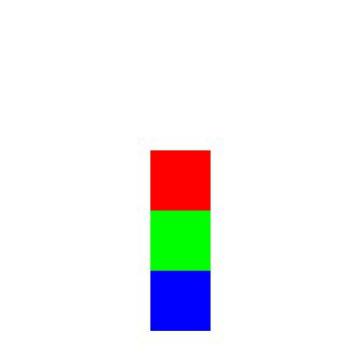}} \\
        \small\parbox{4cm}{\centering\textit{\vspace{1mm} The color of the \textbf{topmost} square is}}
        &
        \small\parbox{4cm}{\centering\textit{\vspace{1mm} The color of the \textbf{bottommost} square is}} \\
    \end{tabular}
    }
    \begin{tikzpicture}[overlay, remember picture]
        \node[anchor=east, rotate=90, font=\scriptsize] at ([xshift=-0.4cm, yshift=1.15cm]box2.west)
        {\textbf{{(a) Horizontal}}};
        \node[anchor=east, rotate=90, font=\scriptsize] at ([xshift=-0.4cm, yshift=1.1cm]box.west)
        {\textbf{{(b) Vertical}}};


        \draw[line width=3pt, color=yellow!80]
        ([xshift=0.06cm,yshift=-0.36cm]box2.north west) rectangle ([xshift=0.55cm,yshift=-2.49cm]box2.north west);
        \draw[line width=3pt, color=yellow!80]
        ([xshift=0.93cm,yshift=-0.36cm]box2.north west) rectangle ([xshift=2.21cm,yshift=-2.49cm]box2.north west);
        \draw[line width=3pt, color=yellow!80]
        ([xshift=4.14cm,yshift=-0.36cm]box2.north west) rectangle ([xshift=5.40cm,yshift=-2.49cm]box2.north west);
        \draw[line width=3pt, color=yellow!80]
        ([xshift=5.78cm,yshift=-0.36cm]box2.north west) rectangle ([xshift=6.29cm,yshift=-2.49cm]box2.north west);

        \draw[->, ultra thick, color=black, line cap=round, arrows={-Stealth[round]}]
        ([xshift=5.0cm,yshift=-1.8cm]box2.north west)
        .. controls ([xshift=4.0cm,yshift=-1.8cm]box2.north west) and ([xshift=2.3cm,yshift=-1.8cm]box2.north west)
        .. ([xshift=1.35cm,yshift=-1.8cm]box2.north west);
        \draw[->, ultra thick, color=black, line cap=round, arrows={-Stealth[round]}]
        ([xshift=5.78cm,yshift=-0.7cm]box2.north west)
        .. controls ([xshift=7.3cm,yshift=-.7cm]box2.north west) and ([xshift=0.6cm,yshift=-.7cm]box2.north west)
        .. ([xshift=0.3cm,yshift=-.7cm]box2.north west);


        \draw[line width=3pt, color=yellow!80]
        ([xshift=0.06cm,yshift=-0.12cm]box.north west) rectangle ([xshift=2.21cm,yshift=-0.65cm]box.north west);
        \draw[line width=3pt, color=yellow!80]
        ([xshift=0.06cm,yshift=-1.03cm]box.north west) rectangle ([xshift=2.21cm,yshift=-2.28cm]box.north west);
        \draw[line width=3pt, color=yellow!80]
        ([xshift=4.14cm,yshift=-0.12cm]box.north west) rectangle ([xshift=6.29cm,yshift=-1.40cm]box.north west);
        \draw[line width=3pt, color=yellow!80]
        ([xshift=4.14cm,yshift=-1.77cm]box.north west) rectangle ([xshift=6.29cm,yshift=-2.28cm]box.north west);

        \draw[->, ultra thick, color=black, line cap=round, arrows={-Stealth[round]}]
        ([xshift=4.14cm,yshift=-2.00cm]box.north west)
        .. controls ([xshift=3.2cm,yshift=-1.7cm]box.north west) and ([xshift=3.1cm,yshift=-0.6cm]box.north west)
        .. ([xshift=2.15cm,yshift=-0.30cm]box.north west);
        \draw[->, ultra thick, color=black, line cap=round, arrows={-Stealth[round]}]
        ([xshift=4.14cm,yshift=-0.60cm]box.north west)
        .. controls ([xshift=3.2cm,yshift=-0.9cm]box.north west) and ([xshift=3.1cm,yshift=-1.4cm]box.north west)
        .. ([xshift=2.15cm,yshift=-1.60cm]box.north west);

    \end{tikzpicture}
    \end{minipage}

    \caption{}
    \vspace{-3mm}
    \label{fig:rel_vs_abs_criss-cross}
\end{figure}

In addition to the probing experiments, we conducted interchange intervention experiments to further characterize the ordinal information encoded by the vision encoder. We perform these experiments as described in Fig.~\ref{fig:rel_vs_abs_criss-cross}, where both the clean and counterfactual images contain squares shifted in a particular direction, and the question asks for the color of the leftmost, rightmost, topmost, or bottommost square. Specifically, if the clean image contains squares shifted to the left, the counterfactual image shifts the squares to the right while reversing their order. We ensure that the rightmost square in the left-shifted configuration is positioned at the center of the image, and similarly for the other shifted configurations.

The intervention is performed over the entire stripe containing the square of the target color. We perform an interchange intervention by replacing the vision-encoder embeddings of the stripe containing the queried square in the clean image with the corresponding stripe from the mirrored counterfactual. This intervention yields different predictions depending on the representation encoded by the vision encoder. 
If the vision embeddings encode ordinal information, then after patching, the model output should change to the color of the square at the opposite end of the ordering. For example, after applying the patching shown in Fig.~\ref{fig:rel_vs_abs_criss-cross}, the output should change from \textcolor{blue}{blue} to \textcolor{red}{red} if the vision encoder encodes ordinal information. In contrast, if the vision encoder encodes absolute positional information, then the output should change from \textcolor{blue}{blue} to \textcolor{green}{green}, because the absolute coordinate values of the intervened \textcolor{blue}{blue} and \textcolor{red}{red} squares would be lower than those of the unintervened \textcolor{green}{green} square.

Fig.~\ref{fig:rel_vs_abs_ve} shows the intervention experiment results. Surprisingly, the expected output for both absolute and ordinal information is substantial, indicating that the vision encoder is forming both types of positional information to encode spatial information. The exception is the vertical configuration with the Gemma model, where patching the output of the vision encoder does not change the final output, indicating a lack of ordering information, consistent with results in Fig.~\ref{fig:criss_cross_strip_gemma}.

\begin{figure}[h]
    \centering
    \begin{subfigure}{\linewidth}
        \centering
        \includegraphics[width=\linewidth]{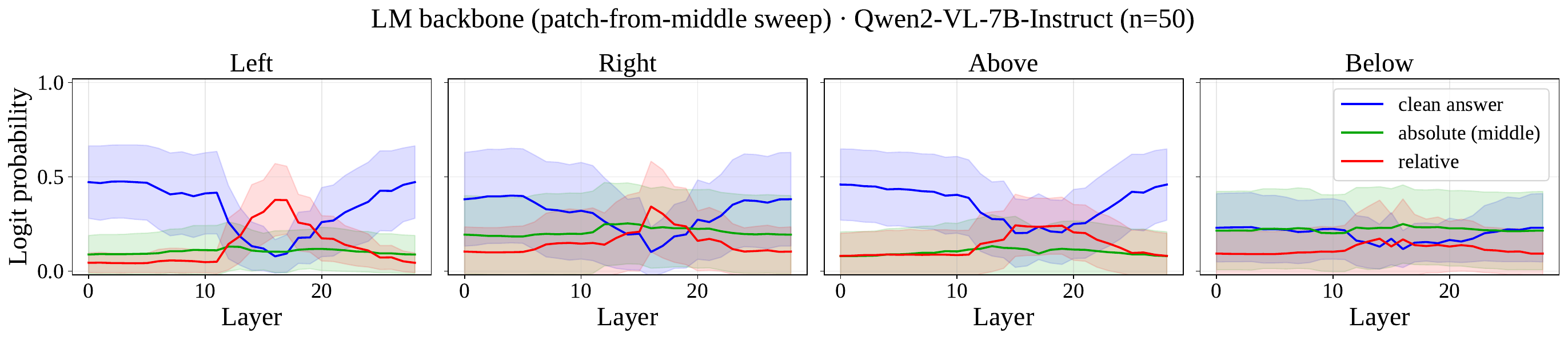}
    \end{subfigure}
    \vspace{0.5em}
    \begin{subfigure}{\linewidth}
        \centering
        \includegraphics[width=\linewidth]{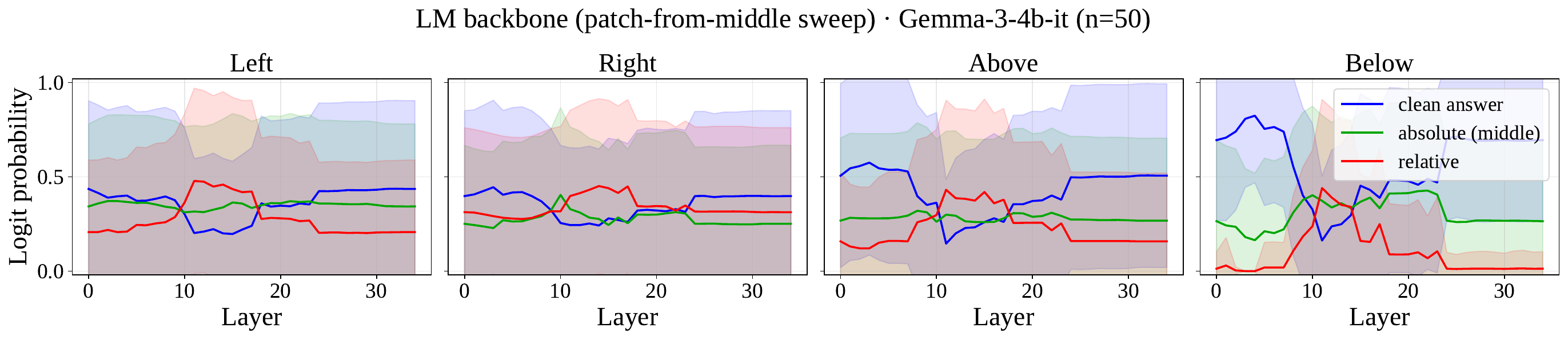}
    \end{subfigure}
    \vspace{0.5em}
    \caption{\textbf{Label probabilities after intervening on the LM backbone's ordering information}: Repeating the mirrored-counterfactual intervention under the vision-ordering ablation of Sec.~\ref{sec:lm_backbone_oi}, where only the LM backbone's own ordering representation can be transferred, produces the relative-order answer almost exclusively, with little probability on the absolute-position answer.}
    \label{fig:rel_vs_abs_lm}
\end{figure}

However, when the same experiment is performed on the LM-backbone ordering information using the setup described in Section~\ref{sec:lm_backbone_oi}, as shown in Fig.~\ref{fig:rel_vs_abs_lm}, we primarily observe the expected output corresponding to the relative ordering information. This suggests that the ordering information encoded by the LM backbone is predominantly relative in nature, consistent with findings from prior studies on LMs~\citep{prakash2024fine,prakash2025languagemodelsuselookbacks,assouel2025visual}.

\begin{figure}[h]
    \centering
    \includegraphics[width=1\linewidth]{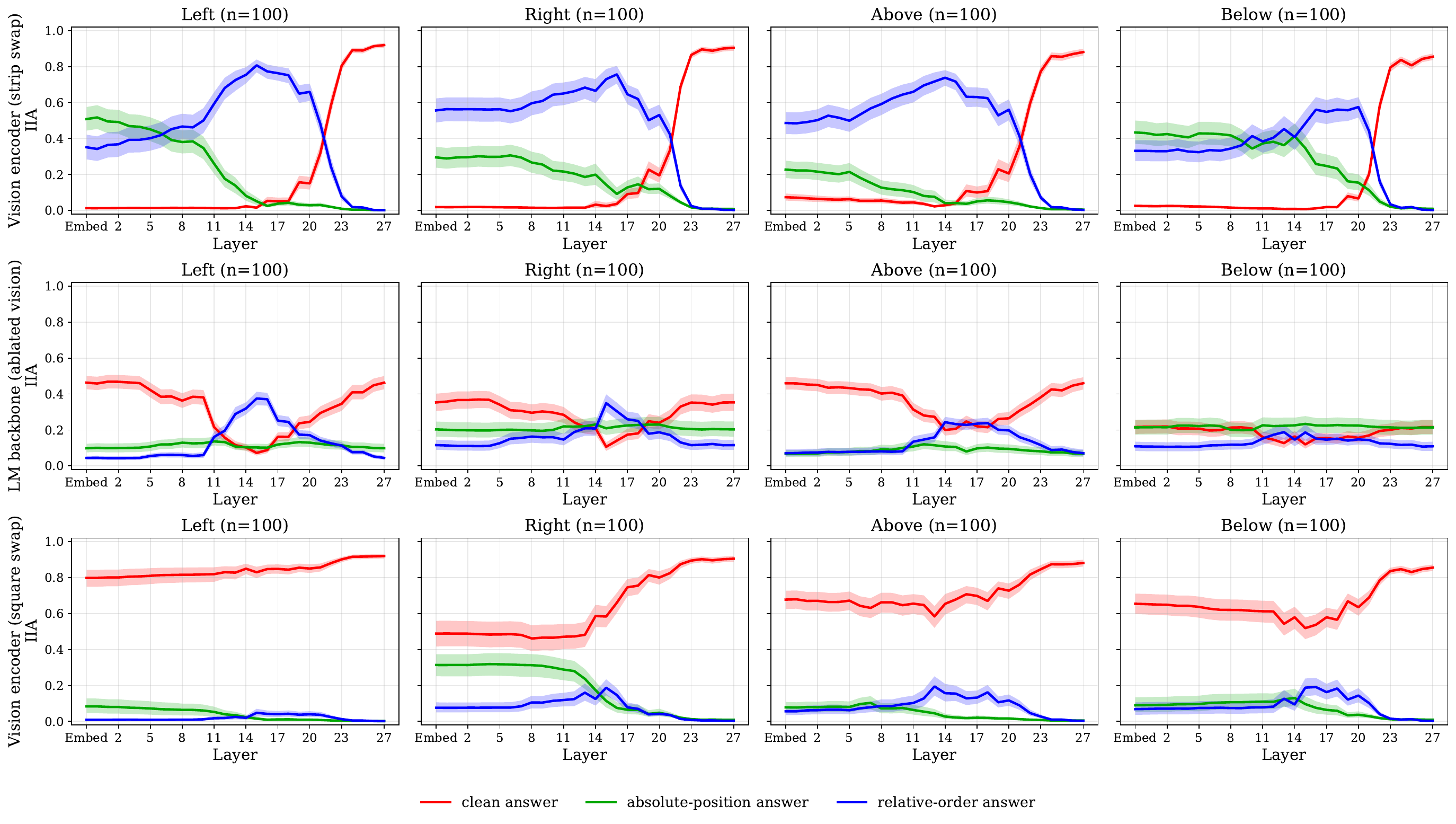}
    \caption{Absolute vs.\ relative position, per-layer curves, \texttt{Qwen2-VL-7B-Instruct}.}
    \label{fig:absrel_layers_qwen}
\end{figure}

\begin{figure}[h]
    \centering
    \includegraphics[width=1\linewidth]{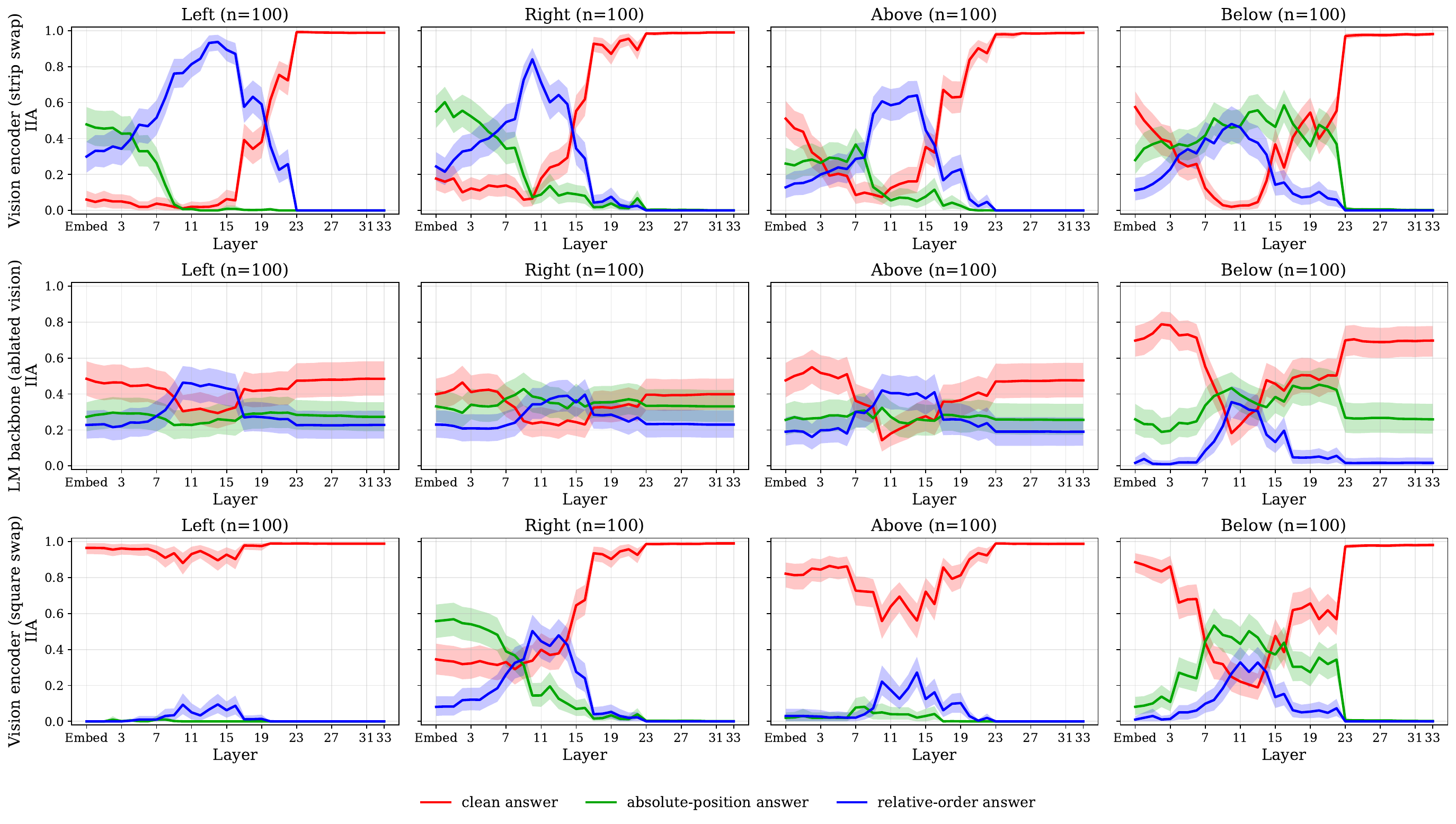}
    \caption{Absolute vs.\ relative position, per-layer curves, \texttt{Gemma-3-4b-it}.}
    \label{fig:absrel_layers_gemma}
\end{figure}

\begin{figure}[h]
    \centering
    \includegraphics[width=1\linewidth]{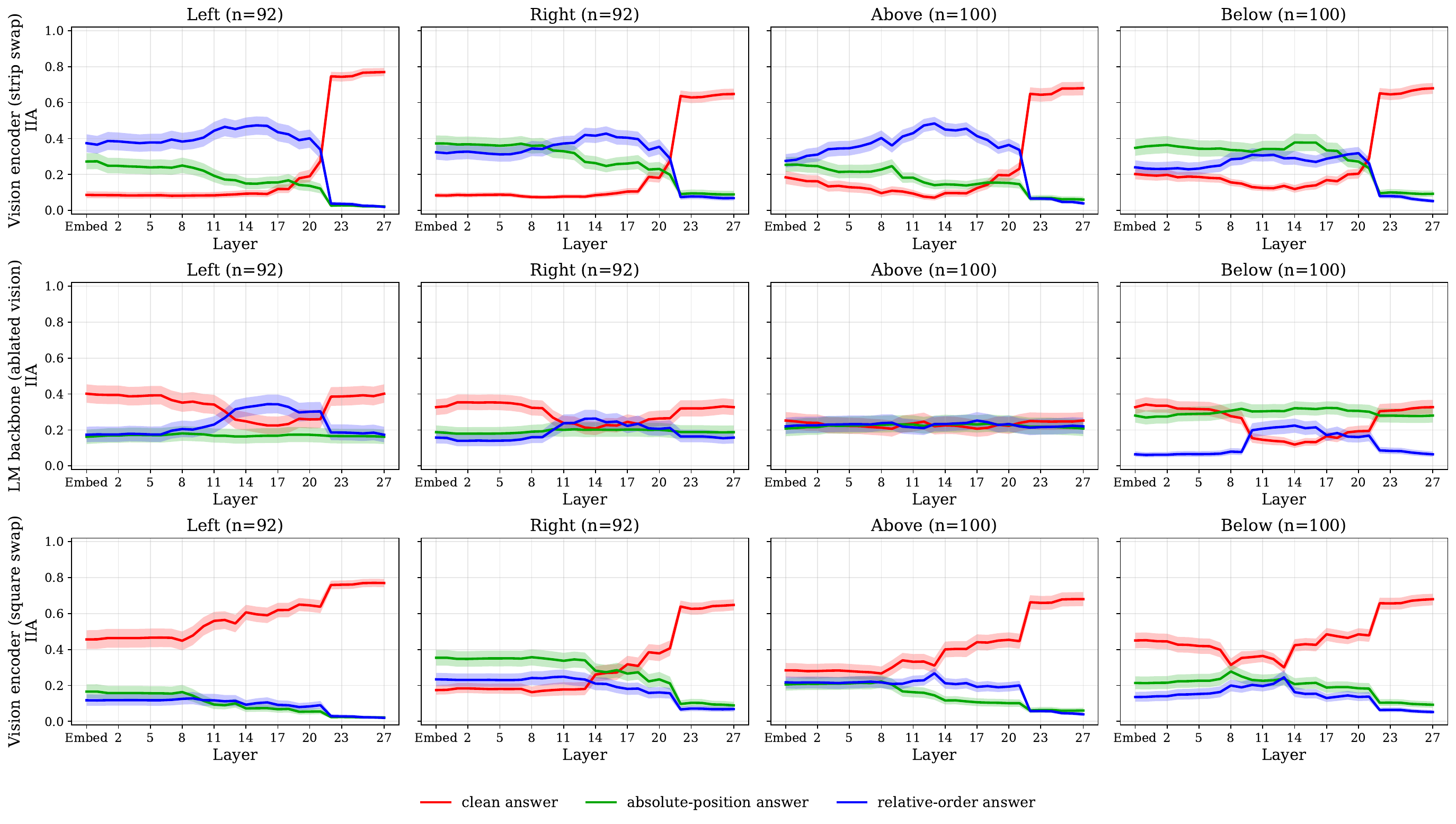}
    \caption{Absolute vs.\ relative position, per-layer curves, \texttt{Qwen2-VL-2B-Instruct}.}
    \label{fig:absrel_layers_qwen2vl2b}
\end{figure}

\clearpage
\paragraph{Per-layer curves with confidence intervals.} Figs.~\ref{fig:absrel_layers_qwen}--\ref{fig:absrel_layers_qwen2vl2b} show the full per-layer curves underlying Figs.~\ref{fig:rel_vs_abs_ve} and~\ref{fig:rel_vs_abs_lm} ($n{=}100$ pairs, 95\% bootstrap CIs), for the strip swap read through the LM (top row), the LM-backbone variant under the vision ablation (middle row), and the square-only swap (bottom row), for \texttt{Qwen2-VL-7B-Instruct}, \texttt{Gemma-3-4b-it}, and \texttt{Qwen2-VL-2B-Instruct}. The vision-encoder swap yields substantial probability for both the absolute and the relative answer, whereas the LM-backbone variant yields the relative answer only.

\clearpage
\subsection{Natural-Image Validation -- Additional results}
\label{app:fixing}

Table~\ref{tab:amplification_full} provides a full breakdown of the amplification results reported in Sec.~\ref{sec:fixing}, separated by spatial direction, together with the results for \texttt{Qwen2-VL-2B-Instruct}, the Qwen model with enough baseline failures for this experiment. Baselines are $232/440$ (Gemma-3-4b, $0.527$), $264/440$ (Qwen2-VL-2B, $0.600$), and $395/439$ (Pixtral-12B, $0.900$; one item is dropped for an out-of-vocabulary prediction). We also ran the protocol on the two larger Qwen models, whose baseline accuracy is already high enough to leave too few failures for meaningful rates, so both are omitted from the tables. \texttt{Qwen2-VL-7B-Instruct} answers $268/279 = 96.1\%$ of its COCO-spatial items correctly, leaving 11 strict failures; probe amplification corrects $3/11$ ($27.3\%$, raising accuracy to $97.1\%$, identically under a single fixed $\alpha$ and per-item $\alpha$), while random directions correct none. \texttt{Qwen3-VL-8B-Instruct} is near ceiling on both datasets ($434/440 = 98.6\%$ on COCO-spatial and $622/625 = 99.5\%$ on What'sUp, leaving 6 and 3 strict failures); probe amplification corrects $4/6$ and $3/3$ of them while random directions correct none. Results are consistent across directions, with probe amplification substantially outperforming random amplification in all cases. The gap between Probe and Probe* indicates room for improvement with better $\alpha$ selection, suggesting the identified ordering subspace is richer than a single fixed coefficient can fully exploit.

\subsubsection{COCO-spatial}
\begin{table*}[h]
\centering
\small
\caption{Full breakdown of amplification results of Sec~\ref{sec:fixing} by spatial direction.
``\% Corr.\ Fail.'' denotes the fraction of baseline failures corrected by the intervention.
In \textit{Random} and \textit{Probe}, the best single fixed $\alpha$ coefficient is chosen across all examples; In \textit{Random*} and \textit{Probe*}, we search for a per-item optimal coefficient, serving as an upper bound on intervention performance. For Pixtral-12B ($n{=}439$; 33 horizontal and 11 vertical failures) the per-axis breakdown of the per-item oracle is not available and is marked ``--''.}
\label{tab:amplification_full}

\begin{tabular}{llcccccc}
\toprule
& & \multicolumn{2}{c}{Gemma3-4B} & \multicolumn{2}{c}{Qwen2-VL-2B} & \multicolumn{2}{c}{Pixtral-12B} \\
\cmidrule(lr){3-4} \cmidrule(lr){5-6} \cmidrule(lr){7-8}
Direction & Intervention & Acc. & \% Corr. Fail. & Acc. & \% Corr. Fail. & Acc. & \% Corr. Fail. \\
\midrule
\multirow{5}{*}{Horizontal}
  & None      & 0.51 & --     & 0.62 & --     & 0.88 & --     \\
  & Random    & 0.54 &  8.0\% & 0.67 & 12.3\% & 0.88 &  0.0\% \\
  & Random*   & 0.56 & 10.1\% & 0.67 & 12.3\% & --   & --     \\
  & Probe     & 0.62 & 23.9\% & 0.80 & 47.2\% & 0.92 & 30.3\% \\
  & Probe*    & 0.70 & 39.1\% & 0.84 & 58.5\% & --   & --     \\
\midrule
\multirow{5}{*}{Vertical}
  & None      & 0.56 & --     & 0.57 & --     & 0.93 & --     \\
  & Random    & 0.59 &  7.0\% & 0.63 & 14.3\% & 0.94 &  9.1\% \\
  & Random*   & 0.60 &  8.5\% & 0.63 & 14.3\% & --   & --     \\
  & Probe     & 0.70 & 32.4\% & 0.72 & 35.7\% & 0.94 & 18.2\% \\
  & Probe*    & 0.75 & 42.3\% & 0.78 & 48.6\% & --   & --     \\
\midrule
\multirow{5}{*}{Overall}
  & None      & 0.53 & --     & 0.60 & --     & 0.90 & --     \\
  & Random    & 0.56 &  7.7\% & 0.65 & 13.1\% & 0.90 &  2.3\% \\
  & Random*   & 0.57 &  9.6\% & 0.65 & 13.1\% & 0.90 &  2.3\% \\
  & Probe     & 0.65 & 26.8\% & 0.77 & 42.6\% & 0.93 & 27.3\% \\
  & Probe*    & 0.72 & 40.2\% & 0.82 & 54.5\% & 0.95 & 54.5\% \\
\bottomrule
\end{tabular}
\end{table*}

\subsubsection{What'sUp}
\label{app:fixing_whatsup}
Table~\ref{tab:amplification_whatsup} reports amplification results on the What'sUp composite dataset, evaluated using the same strict two-ordering protocol as COCO-spatial. Results are consistent with the COCO-spatial findings, with probe amplification consistently outperforming random amplification across all models, confirming that the intervention generalizes across naturalistic settings. 

\begin{table*}[h]
\centering
\caption{Amplification results on the What'sUp dataset using the strict two-ordering evaluation protocol.
``\% Corr.\ Fail.'' denotes the fraction of baseline failures corrected by the intervention.
In \textit{Random} and \textit{Probe}, the best single fixed $\alpha$ coefficient is chosen across all examples; In \textit{Random*} and \textit{Probe*}, we search for a per-item optimal coefficient, serving as an upper bound on intervention performance. Pixtral-12B is evaluated on 484 composite queries (15 strict failures).}
\label{tab:amplification_whatsup}
\begin{tabular}{lcccccc}
\toprule
& \multicolumn{2}{c}{Gemma3-4b} & \multicolumn{2}{c}{Qwen2-VL-2B} & \multicolumn{2}{c}{Pixtral-12B} \\
\cmidrule(lr){2-3} \cmidrule(lr){4-5} \cmidrule(lr){6-7}
Intervention & Acc. & \% Corr. Fail. & Acc. & \% Corr. Fail. & Acc. & \% Corr. Fail. \\
\midrule
None     & 0.79 & --     & 0.82 & --     & 0.97 & --     \\
Random   & 0.83 & 19.1\% & 0.84 &  9.9\% & 0.97 &  6.7\% \\
Random*  & 0.84 & 23.2\% & 0.85 & 13.5\% & 0.97 &  6.7\% \\
Probe    & 0.91 & 55.2\% & 0.90 & 43.2\% & 0.98 & 40.0\% \\
Probe*   & 0.93 & 68.0\% & 0.91 & 47.7\% & 0.99 & 53.3\% \\
\bottomrule
\end{tabular}
\end{table*}

\subsubsection{Other spatial relations: on/under and front/behind}
\label{app:other_relations}
\begin{figure}[h]
    \centering
    \setlength{\tabcolsep}{3pt}
    \begin{tabular}{cccc}
        \includegraphics[width=0.235\linewidth]{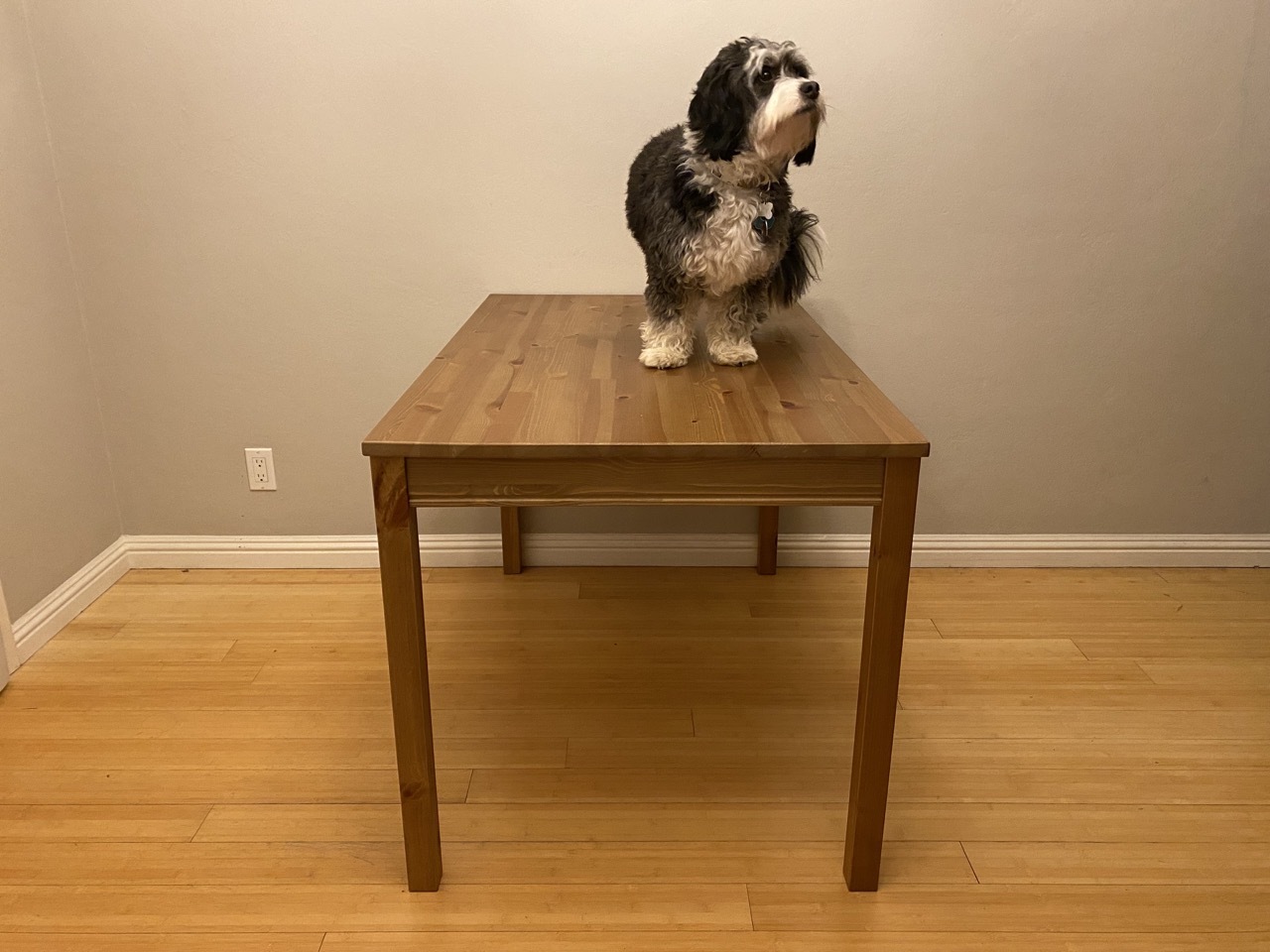} &
        \includegraphics[width=0.235\linewidth]{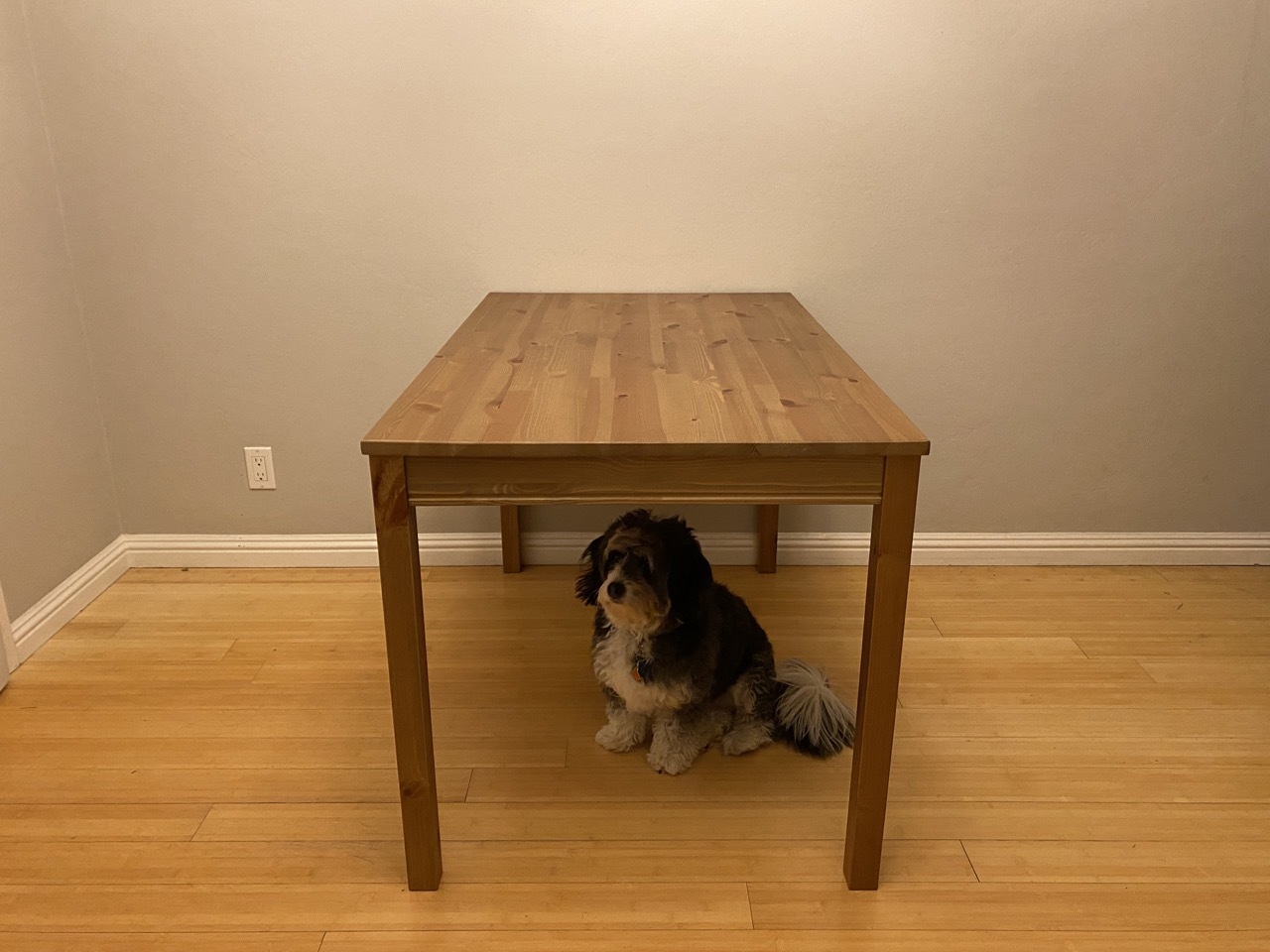} &
        \includegraphics[width=0.235\linewidth]{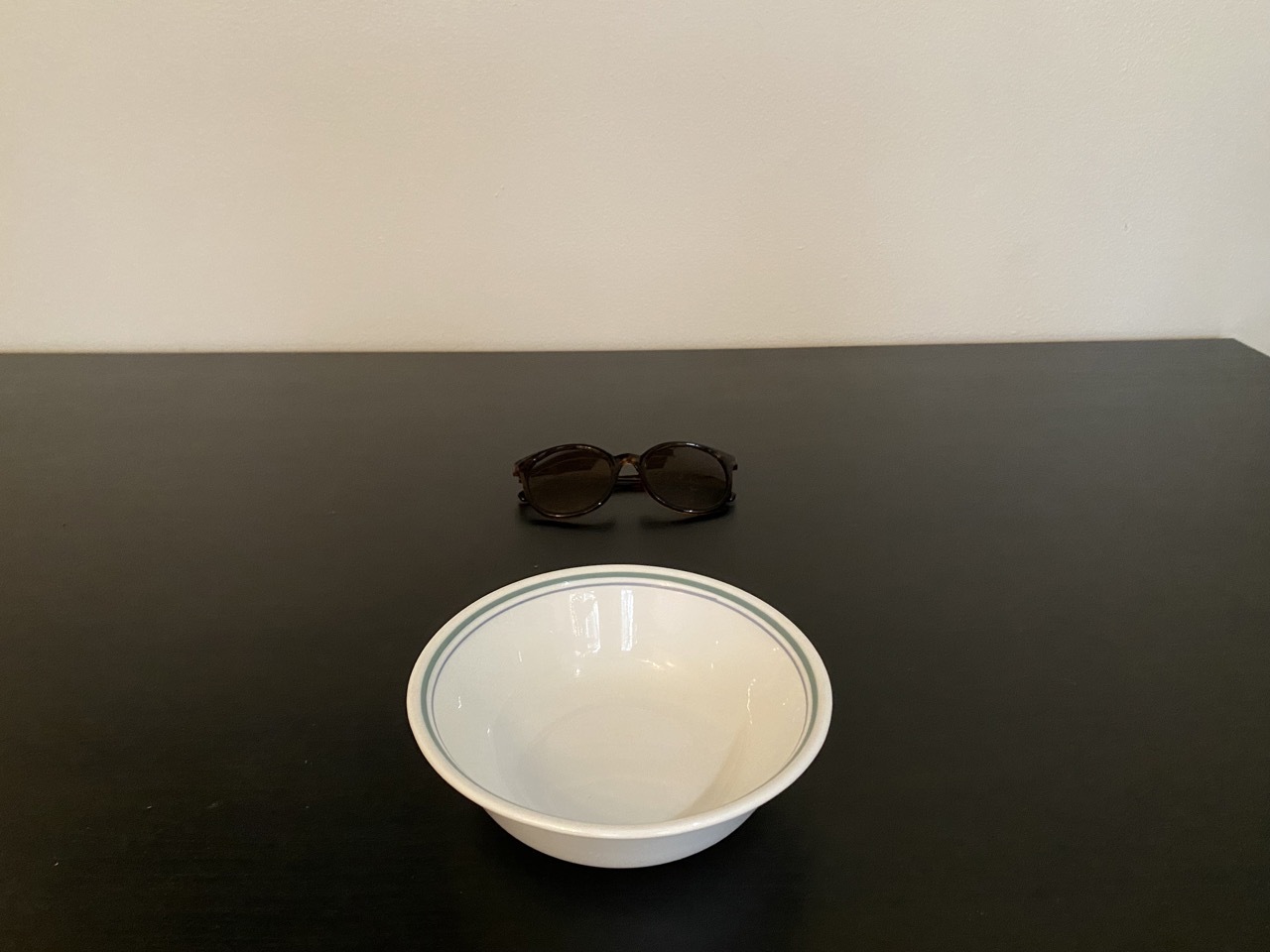} &
        \includegraphics[width=0.235\linewidth]{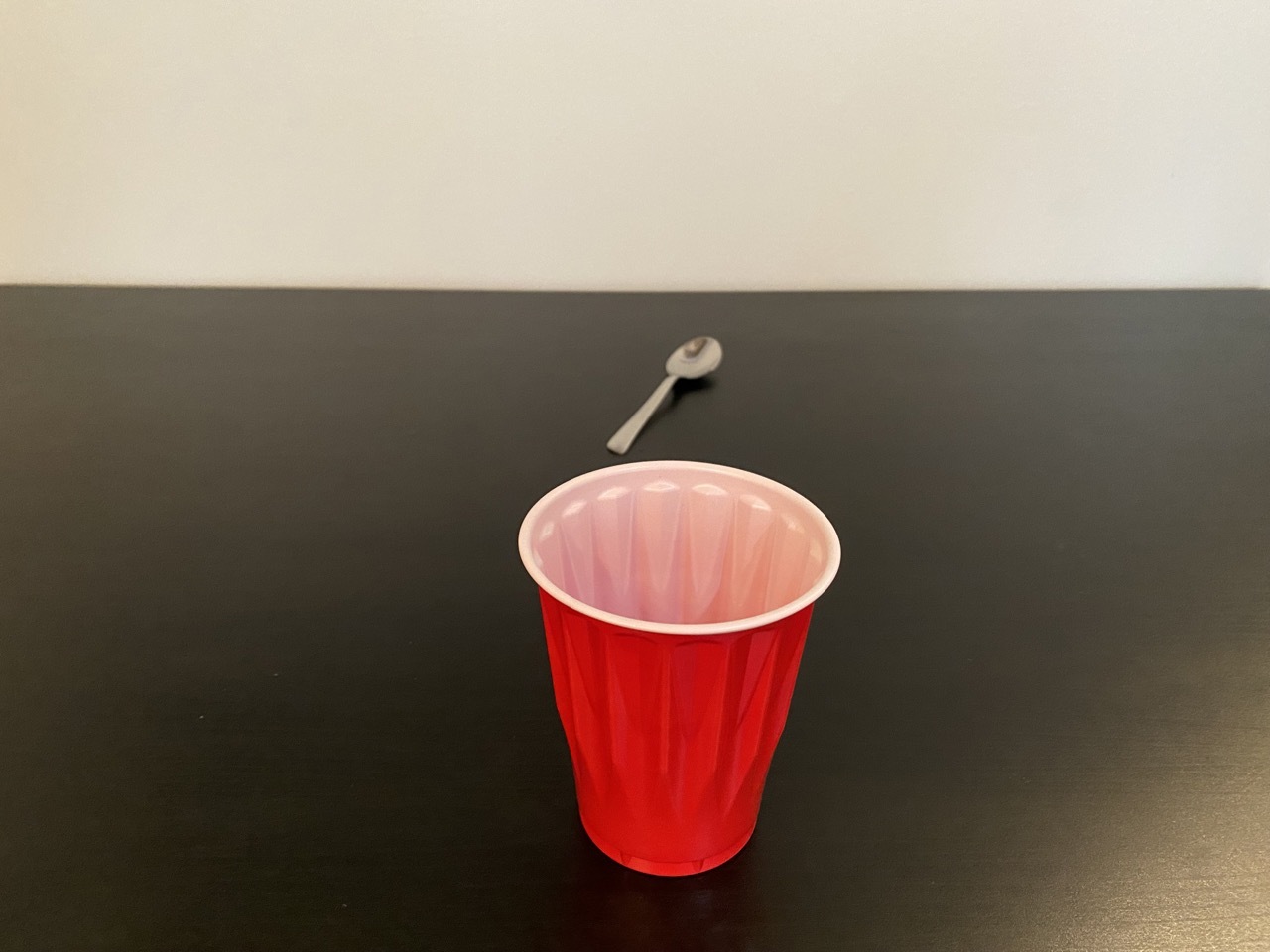} \\
        \small dog \emph{on} table & \small dog \emph{under} table & \small bowl \emph{in front of} sunglasses & \small cup \emph{in front of} spoon
    \end{tabular}
    \caption{\textbf{Examples of the original What'sUp images used for the on/under and front/behind relations}~\citep{kamath2023whatsup}. Each image contains a reference object (table, sunglasses, spoon) and a target object placed on or under it (left two) or in front of it (right two).}
    \label{fig:app:other_relations}
\end{figure}

The main experiments use left/right (What'sUp) and left/right plus above/below (COCO-spatial). We apply the same protocol to the original, unstitched What'sUp images (examples in Fig.~\ref{fig:app:other_relations}) for two further relation pairs: \emph{on/under} (161 images; labels mapped to the spatial answers \texttt{above}/\texttt{below}) and \emph{front/behind} (204 images; answers \texttt{front}/\texttt{behind}, prompt ``Look at the \{reference\}, is the \{target\} in front of it, or behind it?''). Object boxes come from DINO-assisted detection filtered to the query objects; probes are trained on strict-correct images only, and amplification is applied to all image tokens of strict-wrong images. Table~\ref{tab:other_relations} reports fixed-$\alpha$ and per-item oracle fix rates, bootstrap CIs, and the held-out fix rate under the validation/test $\alpha$-selection protocol of App.~\ref{app:alpha_split}. In all three cells the probe direction corrects the majority of failures and far exceeds the random-direction control, including on front/behind, a depth relation not studied mechanistically in this paper. Gemma-3's on/under failures are almost exclusively \texttt{on} items (53 of 55), so its probe direction is effectively tested on one label.

\begin{table}[h]
\centering
\small
\caption{Amplification on additional What'sUp relations. Fix rates are over strict-wrong images (95\% bootstrap CIs over the failures); ``Acc.'' is strict accuracy on all images after amplification. Held-out: $\alpha$ selected on validation failures and evaluated on 20 held-out failures. Front/behind for Gemma-3 was not run.}
\label{tab:other_relations}
\vspace{1mm}
\resizebox{\linewidth}{!}{%
\begin{tabular}{llc l ccc c}
\toprule
Relation & Model & Baseline & Method & $\alpha$ & Fix rate [95\% CI] & Acc. & Held-out fix \\
\midrule
\multirow{4}{*}{on/under}     & \multirow{2}{*}{Qwen2-VL-2B} & \multirow{2}{*}{72.7\%} & Random & 7  & 13.6\% [4.5, 25.0]  & 76.4\% & 15.0\% \\
                              &                              &                         & Probe  & 8  & 68.2\% [52.3, 81.8] & 91.3\% & 65.0\% \\
                              & \multirow{2}{*}{Gemma-3-4b}  & \multirow{2}{*}{65.8\%} & Random & 14 & 16.4\% [7.3, 27.3]  & 71.4\% & 10.0\% \\
                              &                              &                         & Probe  & 12 & 52.7\% [40.0, 65.5] & 83.9\% & 45.0\% \\
\midrule
\multirow{2}{*}{front/behind} & \multirow{2}{*}{Qwen2-VL-2B} & \multirow{2}{*}{73.0\%} & Random & 15 & 27.3\% [16.4, 40.0] & 80.4\% & 30.0\% \\
                              &                              &                         & Probe  & 12 & 76.4\% [63.6, 87.3] & 93.6\% & 70.0\% \\
\bottomrule
\end{tabular}
}
\end{table}

\subsubsection{Probe transfer across datasets and axes}
\label{app:probe_transfer}
If the probe directions capture a general ordering representation rather than dataset-specific features, a probe trained on one dataset or axis should still correct failures on another. We test four transfers on \texttt{Qwen2-VL-2B-Instruct} (Table~\ref{tab:probe_transfer}): (i) the What'sUp \texttt{left}/\texttt{middle}/\texttt{right} probe applied to COCO-spatial horizontal failures (using its \texttt{left}/\texttt{right} columns); (ii) the COCO horizontal probe applied to What'sUp failures; (iii) the COCO horizontal probe applied to COCO \emph{vertical} failures, reading \texttt{left} as \texttt{above} and \texttt{right} as \texttt{below}; and (iv) a single four-class \texttt{left}/\texttt{right}/\texttt{above}/\texttt{below} probe trained on both COCO axes and applied to all COCO failures. In every case the transferred probe corrects far more failures than random directions, and in (i) and (iv) it approaches the in-domain probe. Under the held-out $\alpha$ protocol of App.~\ref{app:alpha_split}, the What'sUp$\rightarrow$COCO probe corrects $46.9\%$ of held-out failures versus $6.2\%$ for random directions. The horizontal$\rightarrow$vertical transfer is the most surprising: the horizontal probe classifies vertical object tokens only at $60\%$, yet amplifying it corrects a third of the vertical failures, suggesting a shared ``first/second'' ordering component across axes.

\begin{table}[h]
\centering
\small
\caption{Probe transfer on \texttt{Qwen2-VL-2B-Instruct}. Probe accuracy is given on the source train / source test / target tokens. Fix rates are over the target's strict failures with the best fixed $\alpha$ (\textsuperscript{*}: per-item oracle). The in-domain column repeats the fixed-$\alpha$ result of a probe trained on the target itself (Tables~\ref{tab:amplification_full} and~\ref{tab:amplification_whatsup}; for (iii) the in-domain vertical probe from the same run).}
\label{tab:probe_transfer}
\vspace{1mm}
\resizebox{\linewidth}{!}{%
\begin{tabular}{llccccc}
\toprule
Transfer & Target failures & Probe acc.\ (train / test / target) & Random\textsuperscript{*} & Transferred & Transferred\textsuperscript{*} & In-domain \\
\midrule
(i)   What'sUp $\rightarrow$ COCO horiz.      & 106 & 97.4 / 95.2 / 85.7 & 16.0\% & 48.1\% & 55.7\% & 47.2\% \\
(ii)  COCO horiz.\ $\rightarrow$ What'sUp     & 111 & 93.7 / 89.1 / 92.4 & 11.7\% & 32.4\% & 34.2\% & 43.2\% \\
(iii) COCO horiz.\ $\rightarrow$ COCO vert.   & --  & 98.9 / 94.1 / 60.0 &  1.9\% & 33.3\% & 33.3\% & 35.2\% \\
(iv)  Shared 4-class $\rightarrow$ all COCO   & 176 & 90.2 / 64.5 / --   & 22.7\% & 42.0\% & 54.0\% & 42.6\% \\
\bottomrule
\end{tabular}
}
\end{table}

\subsubsection{Comparison to chain-of-thought prompting}
\label{app:cot}
A natural question is whether the failures corrected by ordering amplification could be recovered simply by prompting. We evaluate the unmodified models on the same COCO-spatial items under a chain-of-thought (CoT) prompt that asks the model to reason step by step and end with ``\texttt{Answer: <word>}'' (greedy decoding), using the same strict two-ordering protocol. Unlike the amplification sweep, which re-evaluates only the baseline failures and assumes baseline-correct items are kept, CoT changes the prompt for every item, so we re-evaluate the full set; for comparison we also report the amplification at its operating point ($\alpha{=}6$) re-evaluated on the full set. Table~\ref{tab:cot} shows that the effect of CoT is model-dependent: it helps \texttt{Qwen2-VL-2B} (0.600$\rightarrow$0.752), correcting many failures but also breaking three times as many previously correct items as the amplification does, and it is \emph{net harmful} for \texttt{Gemma-3-4b} (0.527$\rightarrow$0.464), where it breaks 44\% of previously correct items and vertical accuracy collapses from 0.559 to 0.385. Inspecting Gemma's generations, the spatial percepts in the reasoning are often correct, but the final answer is bound to the wrong referent (e.g., ``the giraffe is clearly above the bird $\rightarrow$ Answer: above'' when asked about the bird). CoT thus reshuffles which items are answered correctly rather than supplying ordering information, and, unlike the amplification, it provides no causal handle on the mechanism.

\begin{table}[h]
\centering
\small
\caption{Chain-of-thought prompting vs.\ ordering amplification on COCO-spatial (strict two-ordering accuracy on all 440 items; chance $=0.25$). ``Fix'': fraction of baseline failures corrected; ``break'': fraction of baseline-correct items made incorrect. Qwen2-VL-2B is evaluated in the environment of Table~\ref{tab:fixing} (\texttt{transformers} 4.47.1).}
\label{tab:cot}
\vspace{1mm}
\begin{tabular}{lcccccc}
\toprule
& \multicolumn{3}{c}{Qwen2-VL-2B} & \multicolumn{3}{c}{Gemma-3-4b} \\
\cmidrule(lr){2-4}\cmidrule(lr){5-7}
& Acc. & Fix & Break & Acc. & Fix & Break \\
\midrule
Baseline (one-word prompt)                        & 0.600 & --     & --     & 0.527 & --     & -- \\
Ordering amp., $\alpha{=}6$, full re-evaluation   & 0.709 & 36.9\% &  6.4\% & 0.536 & 22.6\% & 18.5\% \\
Ordering amp., per-item $\alpha$ (Table~\ref{tab:fixing}) & 0.82 & 54.5\% & -- & 0.72 & 40.2\% & -- \\
CoT prompting, full re-evaluation                 & 0.752 & 65.3\% & 18.2\% & 0.464 & 35.6\% & 44.0\% \\
\bottomrule
\end{tabular}
\end{table}

\subsubsection{Side effects on non-spatial VQA}
\label{app:side_effects}
\begin{figure}[h]
    \centering
    \includegraphics[width=1\linewidth]{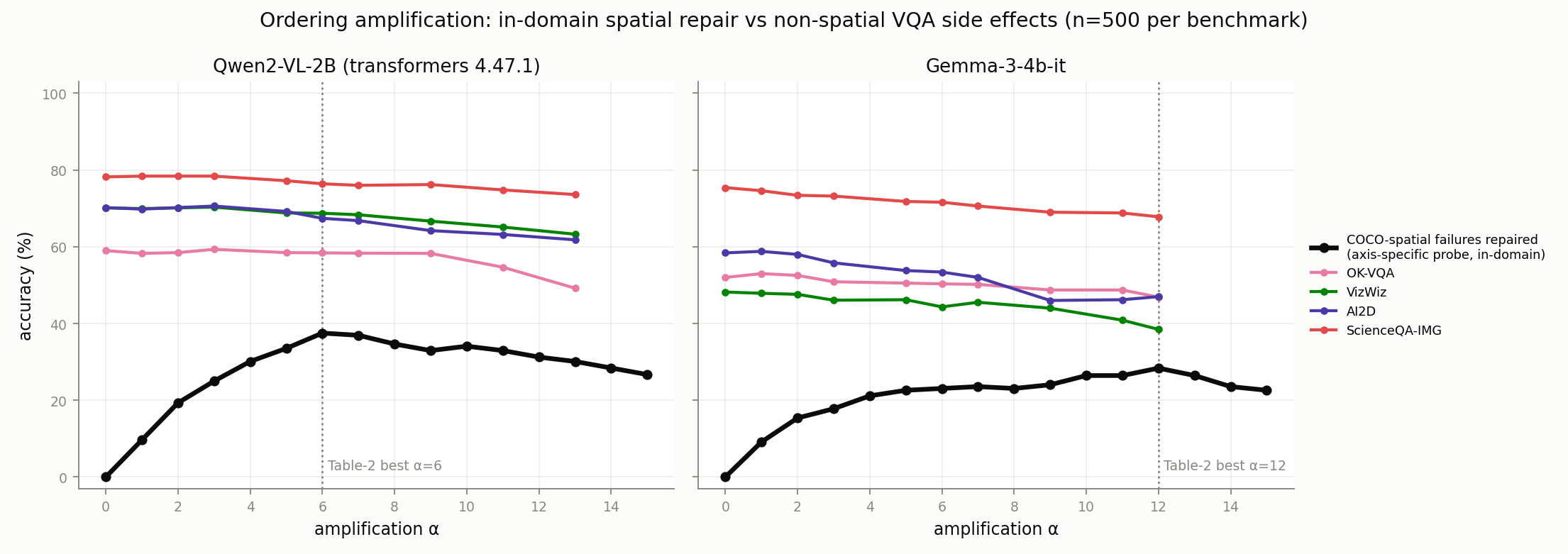}
    \caption{\textbf{Dose--response of ordering amplification.} In-domain repair of COCO-spatial failures (axis-specific probe) and accuracy on non-spatial VQA benchmarks as a function of $\alpha$, for \texttt{Qwen2-VL-2B} (left) and \texttt{Gemma-3-4b} (right). The paper's operating point is $\alpha{=}6$.}
    \label{fig:side_effects}
\end{figure}

Amplifying the ordering directions perturbs every image token, so it could degrade capabilities unrelated to spatial reasoning. We measure this on seven non-spatial VQA benchmarks (POPE, GQA, MMBench, OK-VQA, VizWiz, AI2D, ScienceQA-IMG; 500 items each) by applying the both-axis amplification at the paper's operating point $\alpha{=}6$ and comparing to $\alpha{=}0$ (Table~\ref{tab:side_effects}). For in-domain COCO-spatial we report the fraction of baseline failures repaired with the axis-specific probe, as in Table~\ref{tab:fixing}. At $\alpha{=}6$ the worst degradation is $-5.0$ points (GQA for Qwen2-VL-2B, AI2D for Gemma-3), while the in-domain repair rates are $37.5\%$ and $23.1\%$. Fig.~\ref{fig:side_effects} shows the full $\alpha$ sweep: the in-domain effect follows an inverted-U with a peak around $\alpha \approx 5$--$6$, and the non-spatial degradation grows with $\alpha$, becoming severe only at large $\alpha$ (e.g., $-16$ points on GQA for Qwen2-VL-2B at $\alpha{=}13$). A more conservative operating point, $\alpha{=}3$, keeps every non-spatial drop within $2$ points for both models while retaining roughly three quarters of the repair effect.

\begin{table}[h]
\centering
\small
\caption{Accuracy (\%) on non-spatial VQA benchmarks before ($\alpha{=}0$) and after ($\alpha{=}6$) both-axis ordering amplification; $n{=}500$ items per benchmark. The first row gives the in-domain repair rate of COCO-spatial baseline failures.}
\label{tab:side_effects}
\vspace{1mm}
\resizebox{\linewidth}{!}{%
\begin{tabular}{lcc}
\toprule
Benchmark & Qwen2-VL-2B: $\alpha{=}0 \rightarrow \alpha{=}6$ ($\Delta$) & Gemma-3-4b: $\alpha{=}0 \rightarrow \alpha{=}6$ ($\Delta$) \\
\midrule
COCO-spatial failures repaired (in-domain) & 0 $\rightarrow$ 37.5 ($+37.5$) & 0 $\rightarrow$ 23.1 ($+23.1$) \\
\midrule
POPE          & 89.0 $\rightarrow$ 85.6 ($-3.4$) & 86.0 $\rightarrow$ 83.2 ($-2.8$) \\
GQA           & 61.6 $\rightarrow$ 56.6 ($-5.0$) & 37.0 $\rightarrow$ 38.6 ($+1.6$) \\
MMBench       & 78.2 $\rightarrow$ 77.4 ($-0.8$) & 65.2 $\rightarrow$ 64.4 ($-0.8$) \\
OK-VQA        & 59.0 $\rightarrow$ 58.4 ($-0.6$) & 52.0 $\rightarrow$ 50.3 ($-1.7$) \\
VizWiz        & 70.1 $\rightarrow$ 68.7 ($-1.4$) & 48.2 $\rightarrow$ 44.3 ($-3.9$) \\
AI2D          & 70.2 $\rightarrow$ 67.4 ($-2.8$) & 58.4 $\rightarrow$ 53.4 ($-5.0$) \\
ScienceQA-IMG & 78.2 $\rightarrow$ 76.4 ($-1.8$) & 75.4 $\rightarrow$ 71.6 ($-3.8$) \\
\bottomrule
\end{tabular}
}
\end{table}

\clearpage
\subsection{Probing}
\label{app:probing}
\begin{figure}[h]
    \centering
    \begin{tabular}{cc}
        \includegraphics[width=0.20\linewidth]{figures_main/squares.pdf} &
        \includegraphics[width=0.75\linewidth]{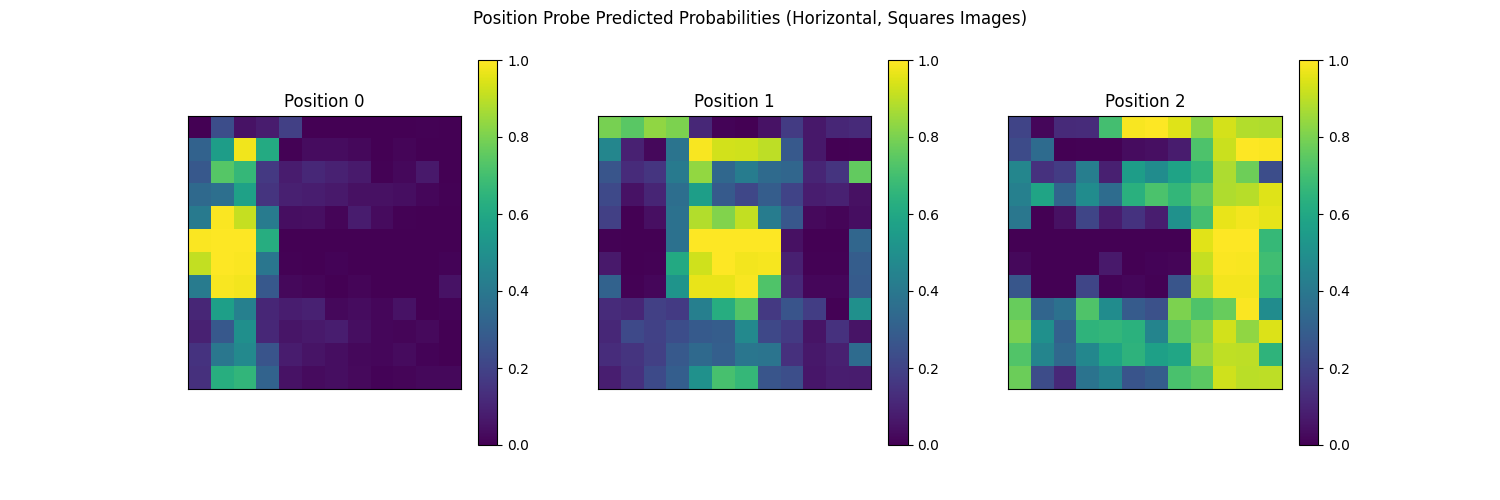} \\
    \end{tabular}
    \caption{Horizontal position probe for square images, Qwen}
    \label{fig:qwen_square_horizontal}
\end{figure}

\begin{figure}[h]
    \centering
    \begin{tabular}{cc}
    \includegraphics[width=0.20\linewidth,angle=90]{figures_main/squares.pdf} &
        \includegraphics[width=0.75\linewidth]{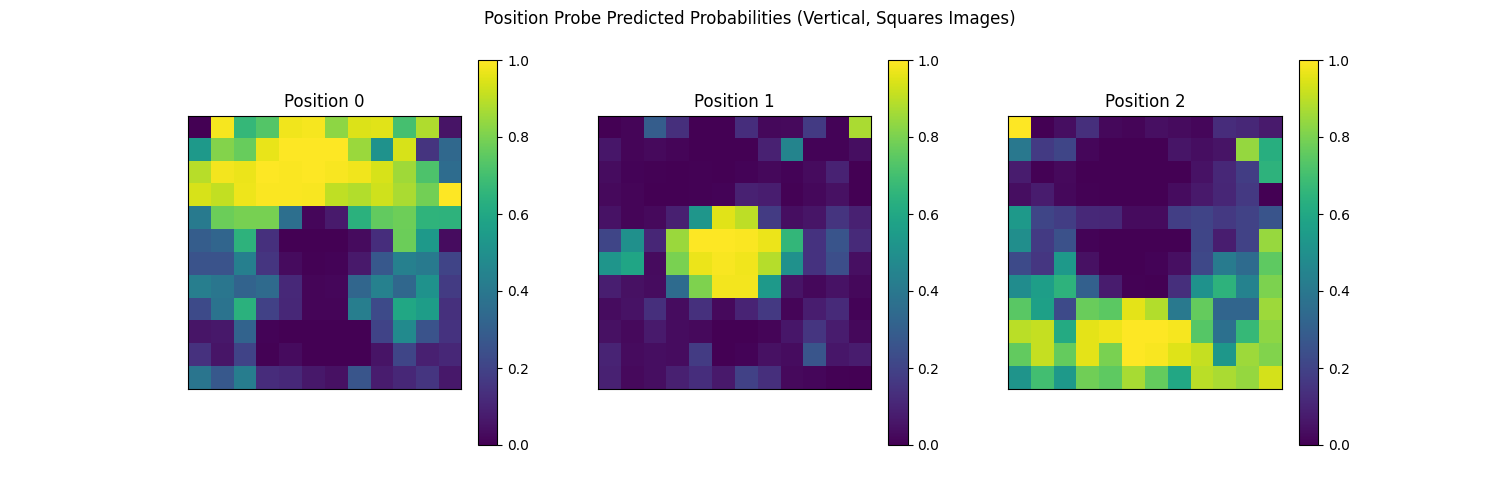} \\
    \end{tabular}
    \caption{Vertical position probe for square images, Qwen}
    \label{fig:qwen_square_vertical}
\end{figure}

\begin{figure}[h]
    \centering
    \begin{tabular}{cc}
    \includegraphics[width=0.20\linewidth]{figures_main/shapes.pdf} &
        \includegraphics[width=0.75\linewidth]{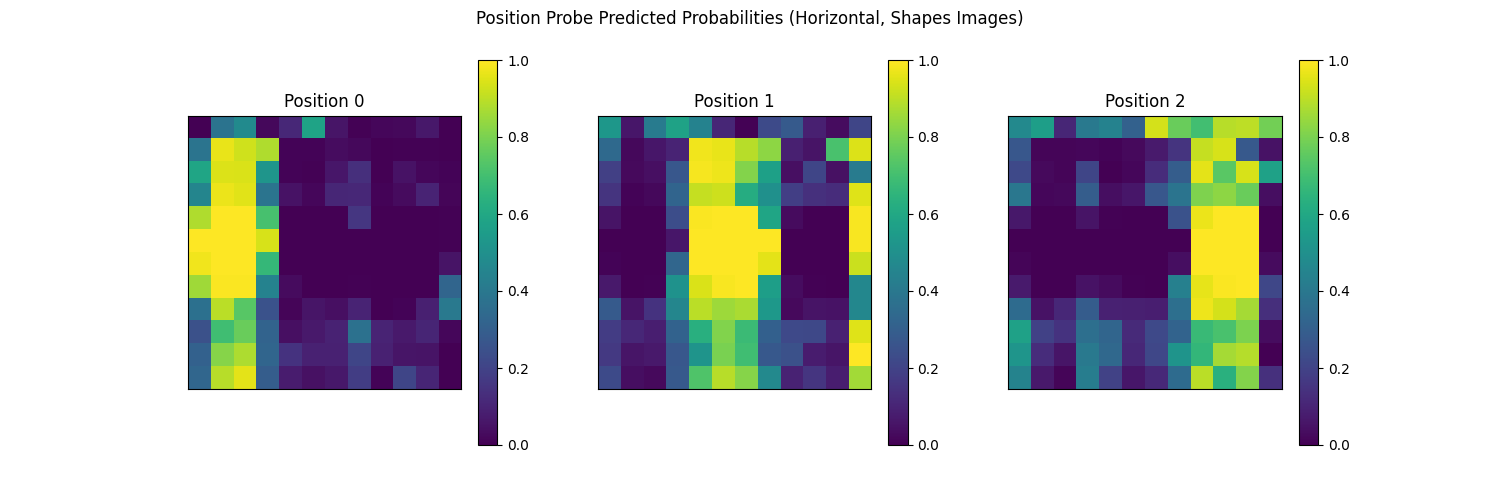} \\
    \end{tabular}
    \caption{Horizontal position probe for shape images, Qwen}
    \label{fig:qwen_shapes_horizontal}
\end{figure}

\begin{figure}[h]
    \centering
    \begin{tabular}{cc}
    \includegraphics[width=0.20\linewidth]{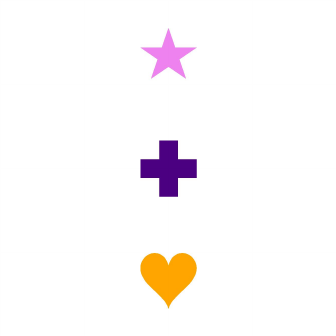} &
        \includegraphics[width=0.75\linewidth]{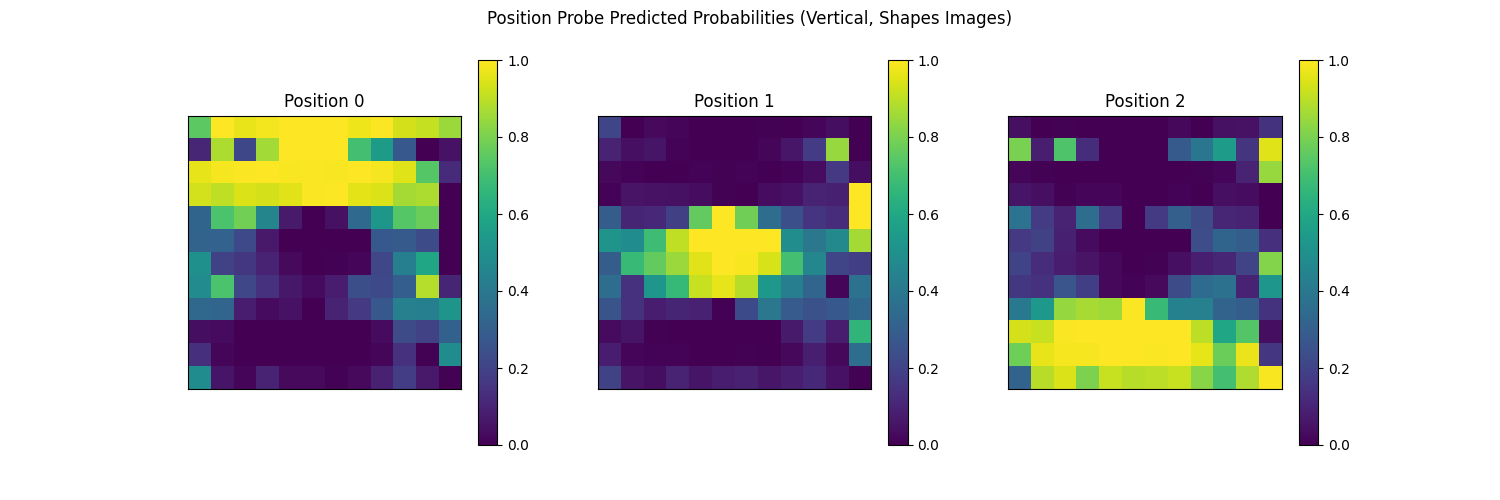} \\
    \end{tabular}
    \caption{Vertical position probe for shape images, Qwen}
    \label{fig:qwen_shapes_vertical}
\end{figure}

\begin{figure}[h]
    \centering
    \begin{tabular}{cc}
    \includegraphics[width=0.20\linewidth]{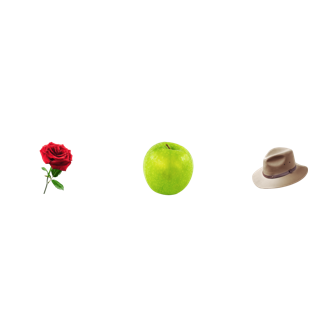} &
        \includegraphics[width=0.75\linewidth]{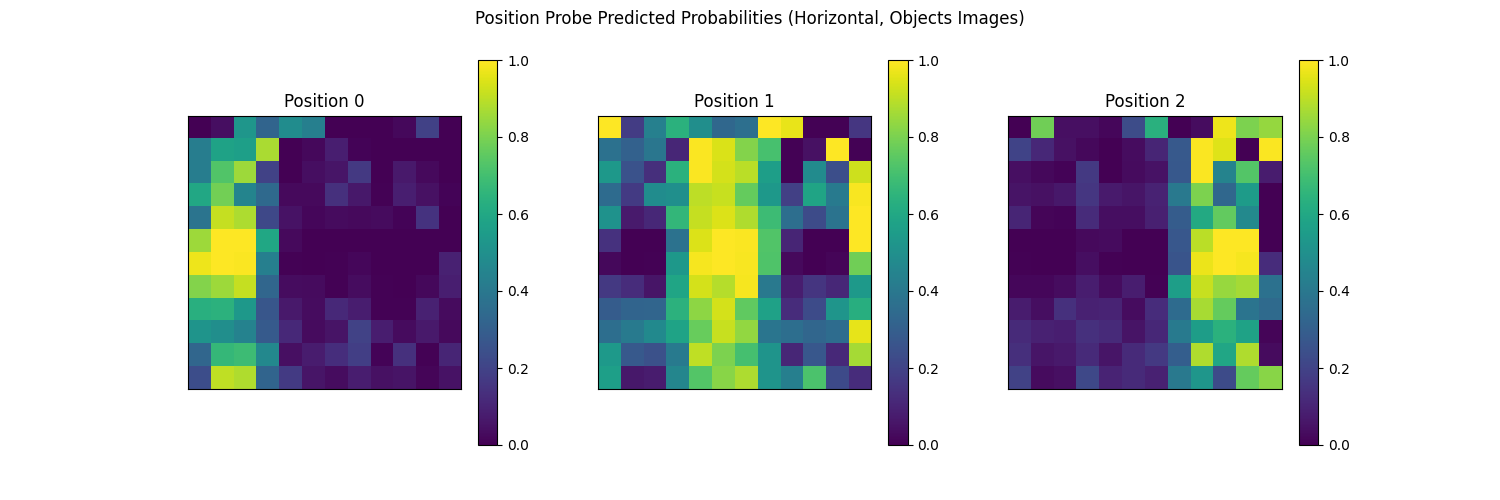} \\
    \end{tabular}
    \caption{Horizontal position probe for object images, Qwen}
    \label{fig:qwen_object_horizontal}
\end{figure}

\begin{figure}[h]
    \centering
    \begin{tabular}{cc}
    \includegraphics[width=0.20\linewidth]{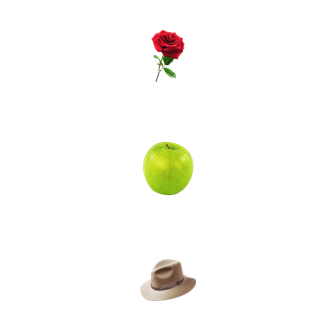} &
        \includegraphics[width=0.75\linewidth]{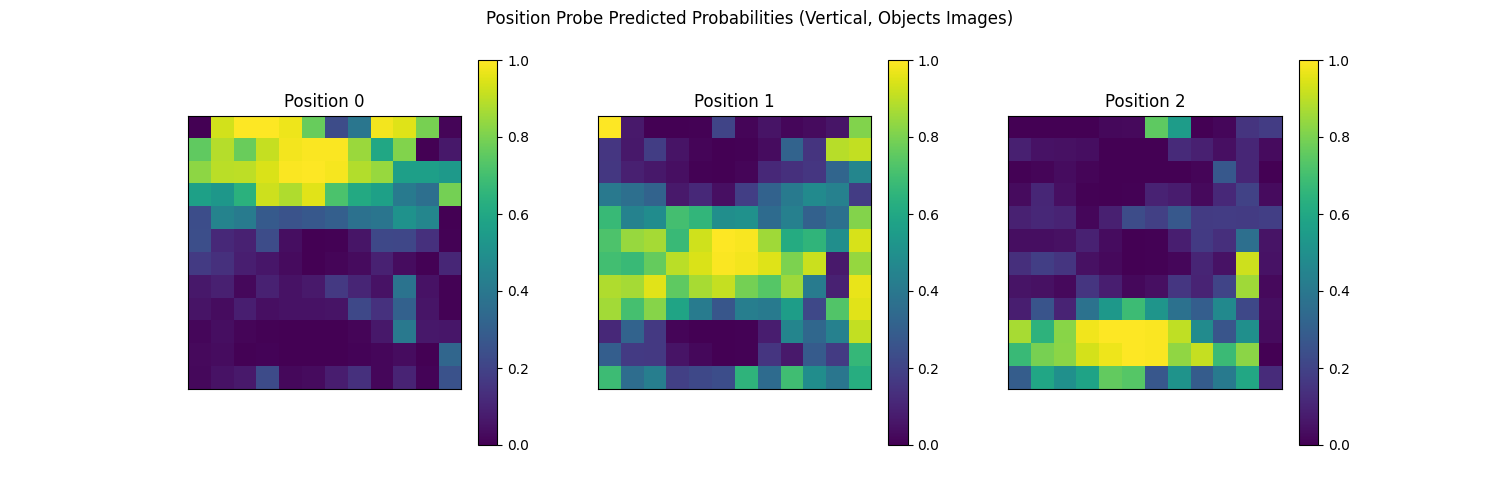} \\
    \end{tabular}
    \caption{Vertical position probe for object images, Qwen}
    \label{fig:qwen_object_vertical}
\end{figure}

\begin{figure}[h]
    \centering
    \begin{tabular}{cc}
        \includegraphics[width=0.20\linewidth]{figures_main/squares.pdf} &
        \includegraphics[width=0.75\linewidth]{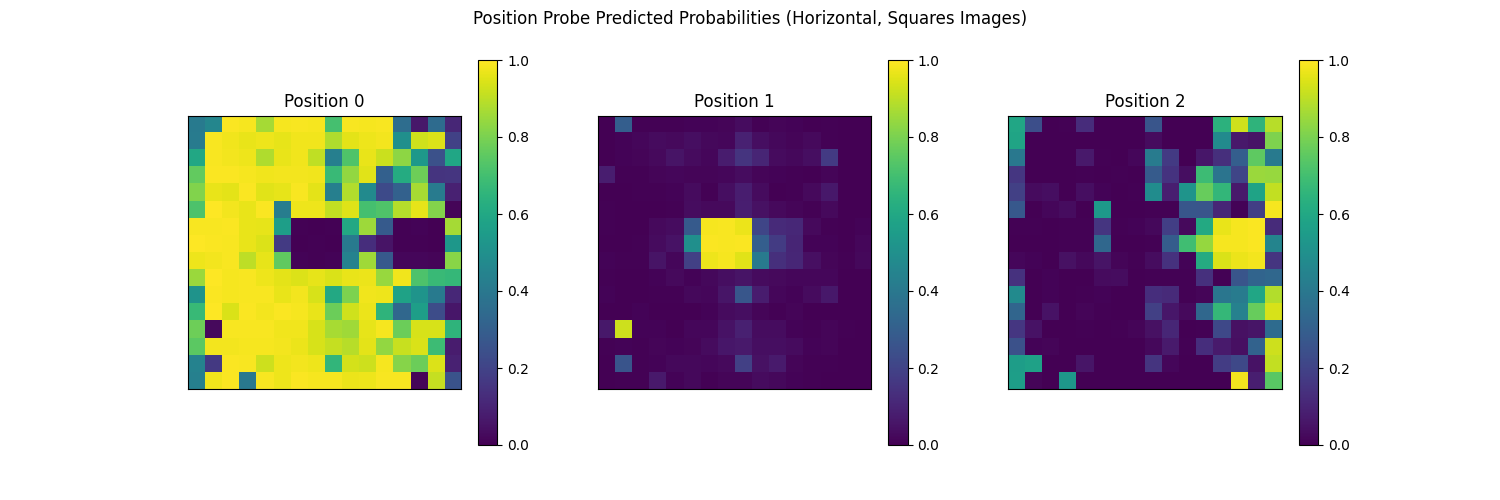} \\
    \end{tabular}
    \caption{Horizontal position probe for square images, Gemma}
    \label{fig:gemma_square_horizontal}
\end{figure}

\begin{figure}[h]
    \centering
    \begin{tabular}{cc}
    \includegraphics[width=0.20\linewidth,angle=90]{figures_main/squares.pdf} &
        \includegraphics[width=0.75\linewidth]{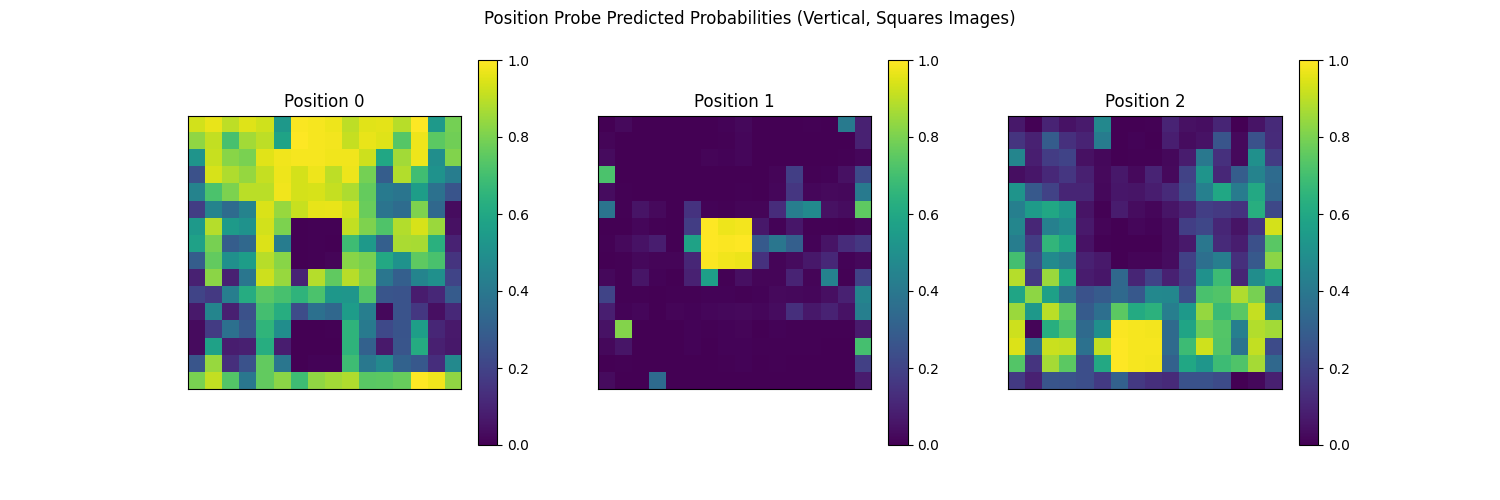} \\
    \end{tabular}
    \caption{Vertical position probe for square images, Gemma}
    \label{fig:gemma_square_vertical}
\end{figure}

\begin{figure}[h]
    \centering
    \begin{tabular}{cc}
    \includegraphics[width=0.20\linewidth]{figures_main/shapes.pdf} &
        \includegraphics[width=0.75\linewidth]{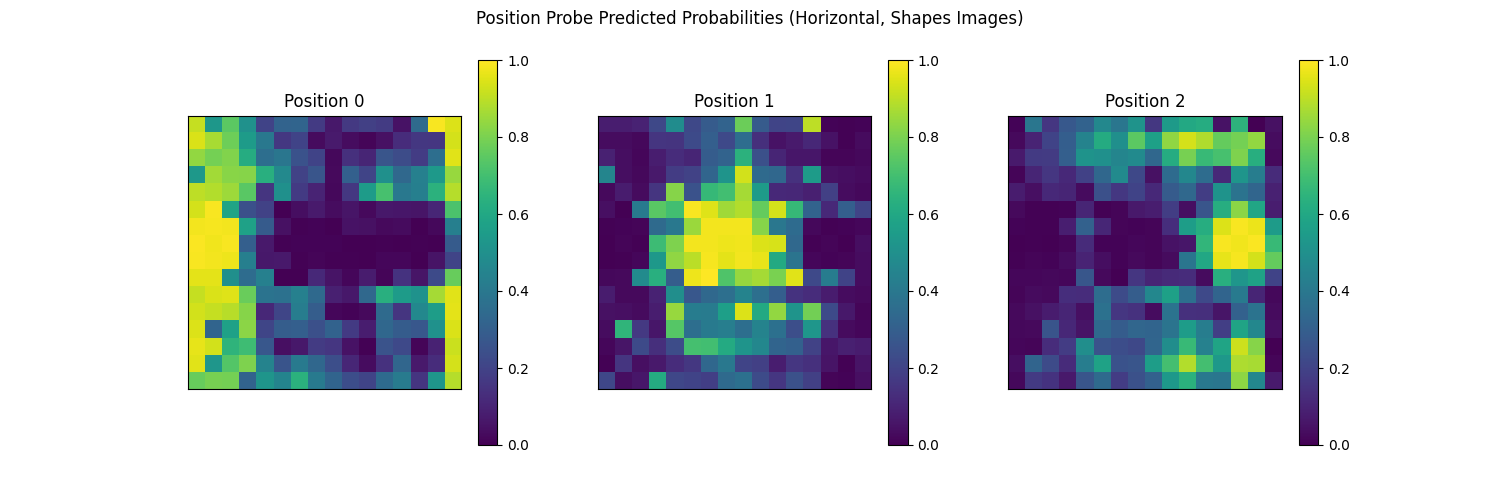} \\
    \end{tabular}
    \caption{Horizontal position probe for shape images, Gemma}
    \label{fig:gemma_shapes_horizontal}
\end{figure}

\begin{figure}[h]
    \centering
    \begin{tabular}{cc}
    \includegraphics[width=0.20\linewidth]{figures_main/shapes2.pdf} &
        \includegraphics[width=0.75\linewidth]{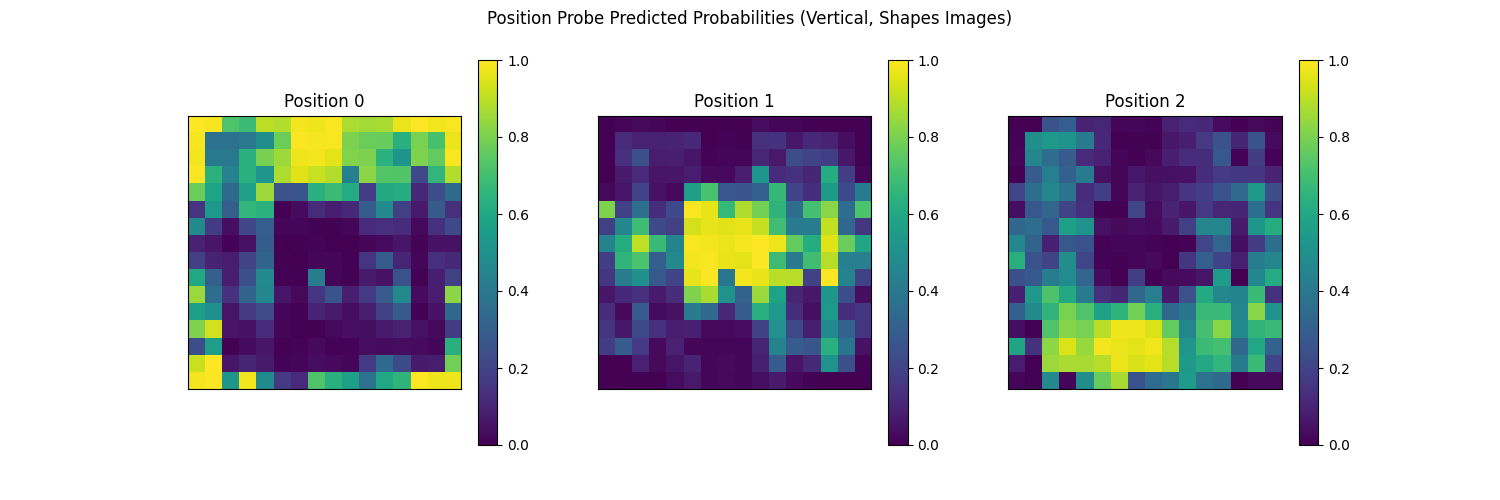} \\
    \end{tabular}
    \caption{Vertical position probe for shape images, Gemma}
    \label{fig:gemma_shapes_vertical}
\end{figure}

\begin{figure}[h]
    \centering
    \begin{tabular}{cc}
    \includegraphics[width=0.20\linewidth]{figures_main/objects.png} &
        \includegraphics[width=0.75\linewidth]{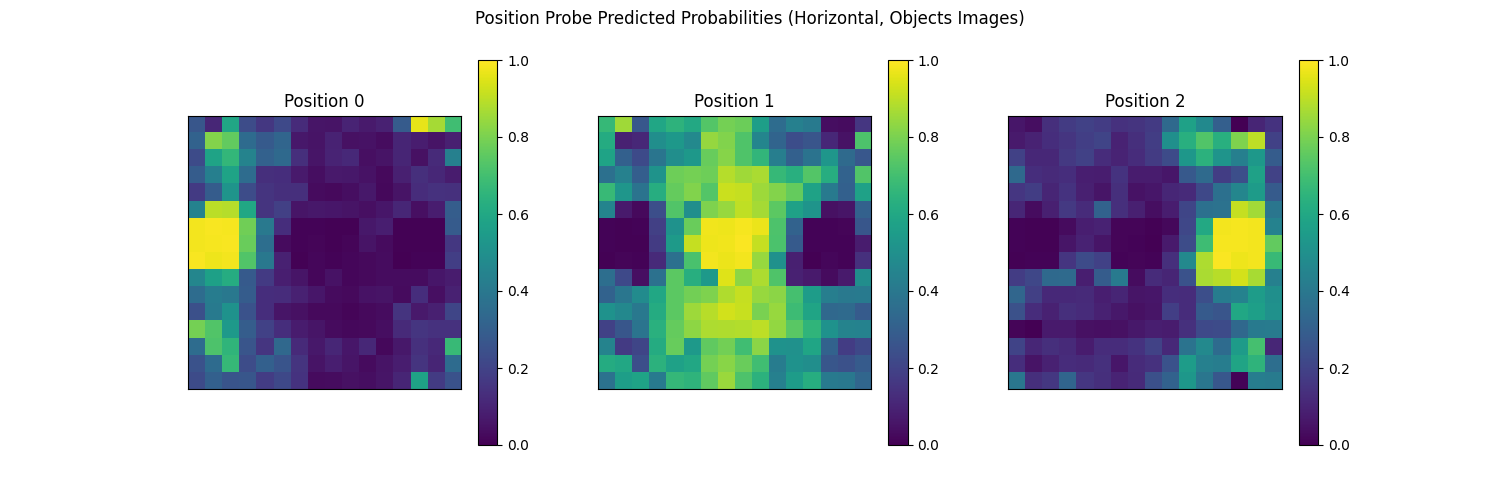} \\
    \end{tabular}
    \caption{Horizontal position probe for object images, Gemma}
    \label{fig:gemma_object_horizontal}
\end{figure}

\begin{figure}[h]
    \centering
    \begin{tabular}{cc}
    \includegraphics[width=0.20\linewidth]{figures_main/objects2.png} &
        \includegraphics[width=0.75\linewidth]{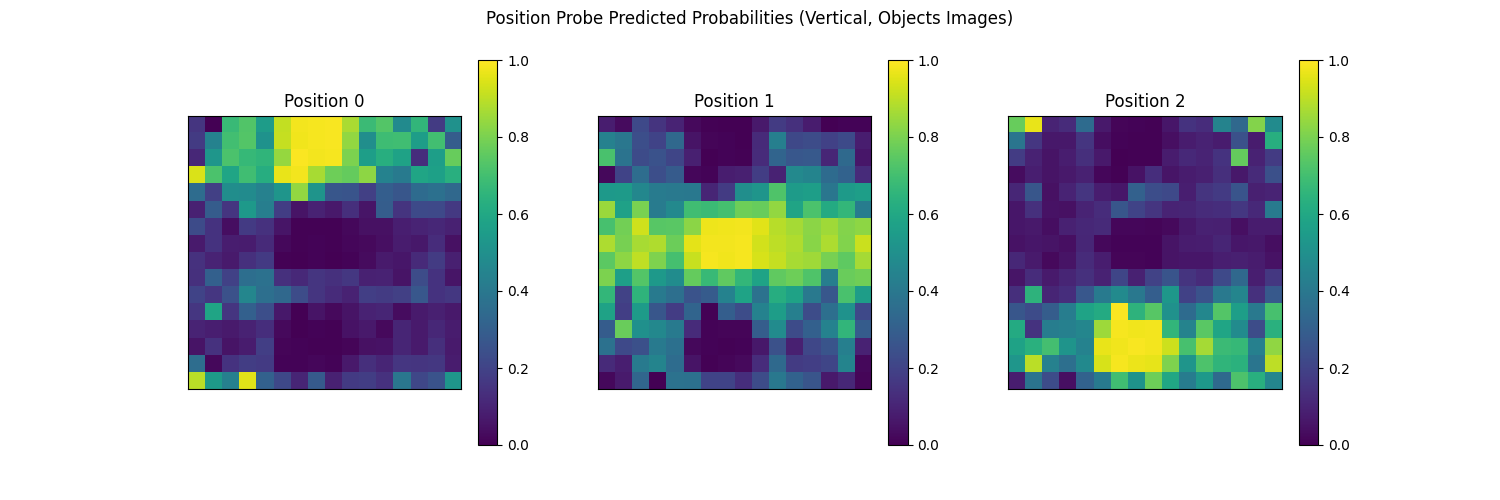} \\
    \end{tabular}
    \caption{Vertical position probe for object images, Gemma}
    \label{fig:gemma_object_vertical}
\end{figure}

\begin{figure}[h]
    \centering
    \includegraphics[width=0.85\linewidth]{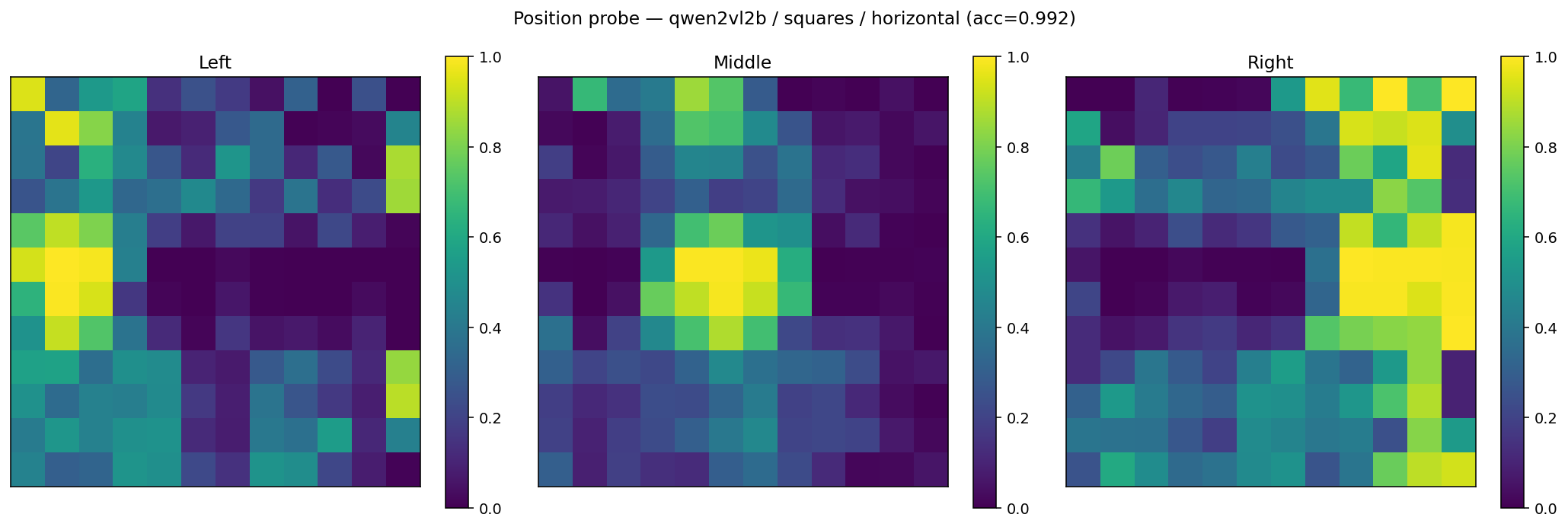}\\
    \includegraphics[width=0.85\linewidth]{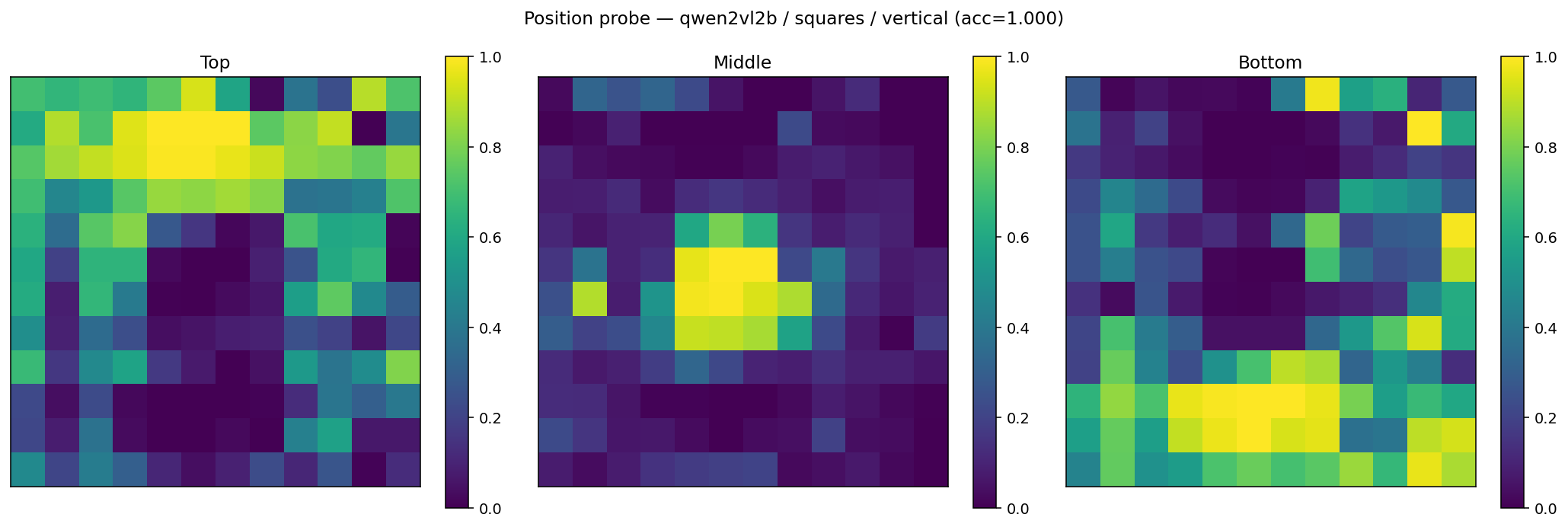}\\
    \includegraphics[width=0.85\linewidth]{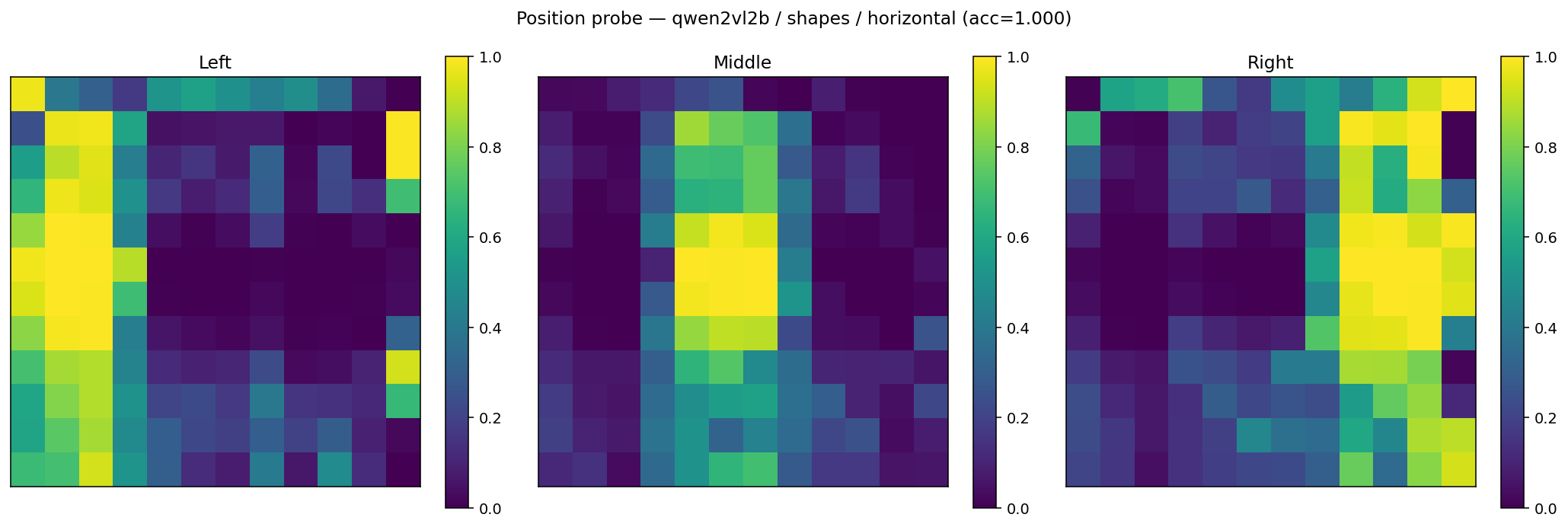}\\
    \includegraphics[width=0.85\linewidth]{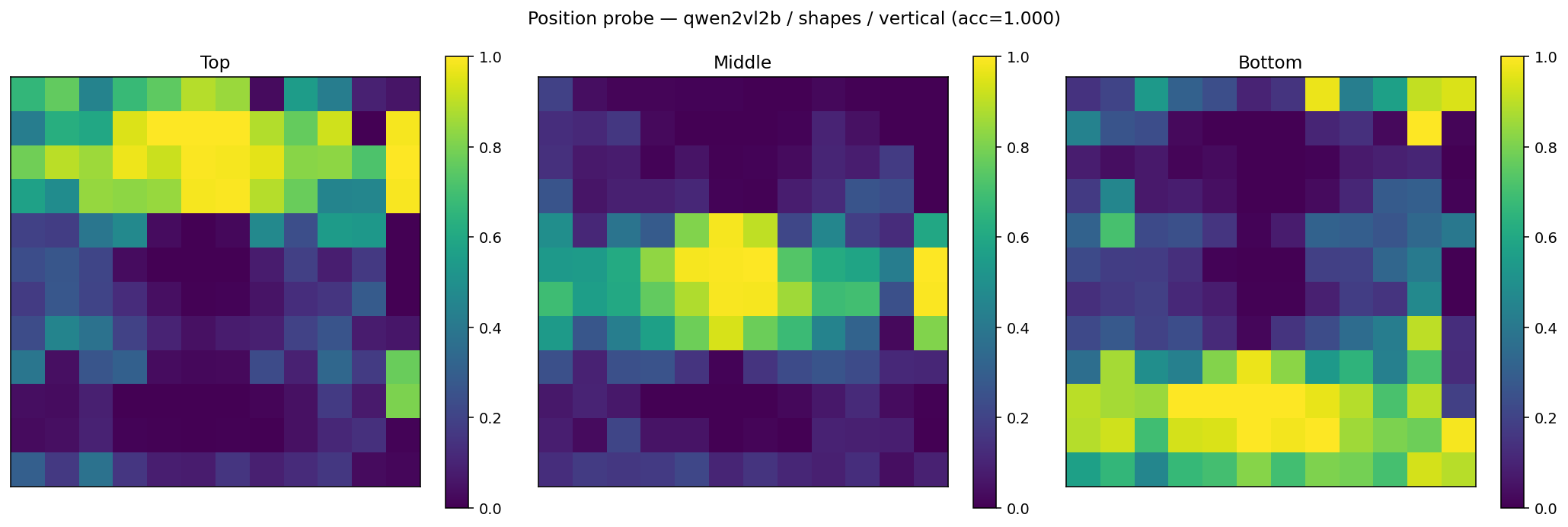}
    \caption{Position-probe predictions for \texttt{Qwen2-VL-2B-Instruct} on the Squares (horizontal, vertical) and Shapes (horizontal, vertical) settings.}
    \label{fig:probe_qwen2vl2b_a}
\end{figure}

\begin{figure}[h]
    \centering
    \includegraphics[width=0.85\linewidth]{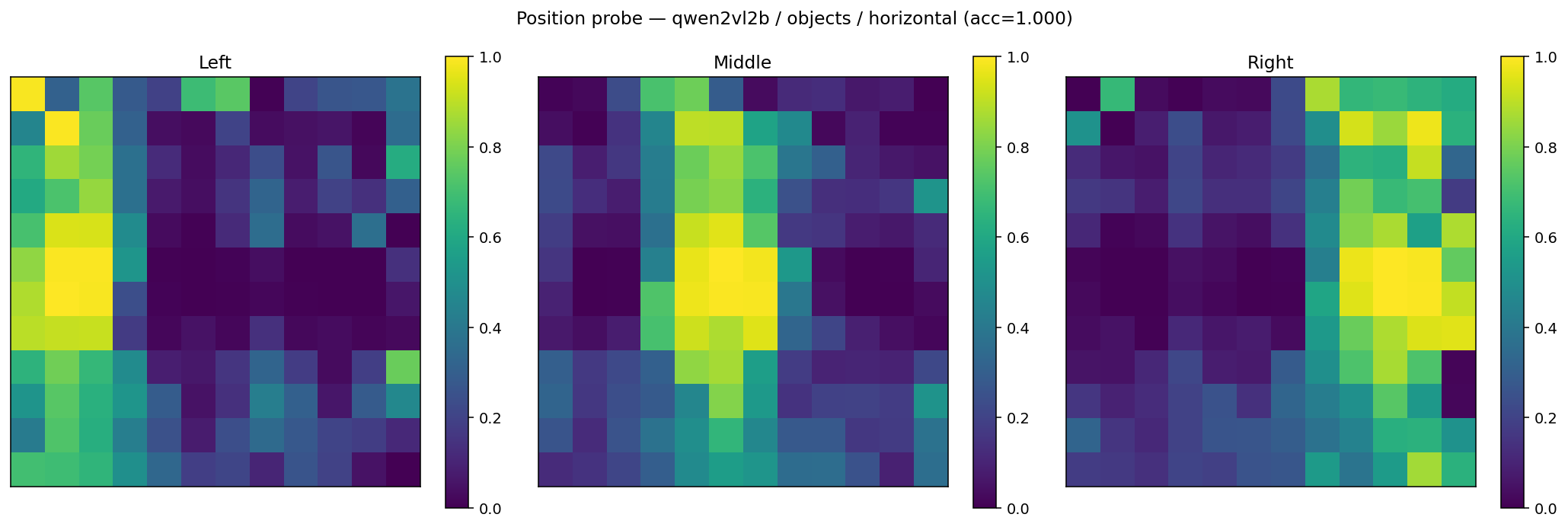}\\
    \includegraphics[width=0.85\linewidth]{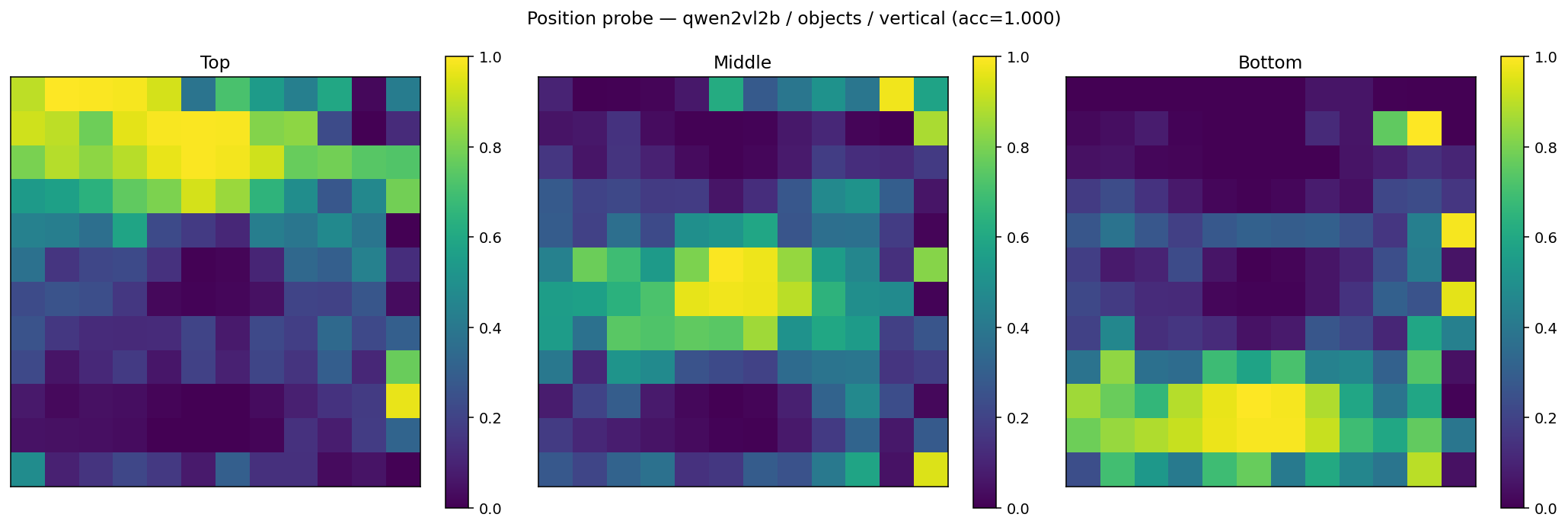}\\
    \includegraphics[width=0.85\linewidth]{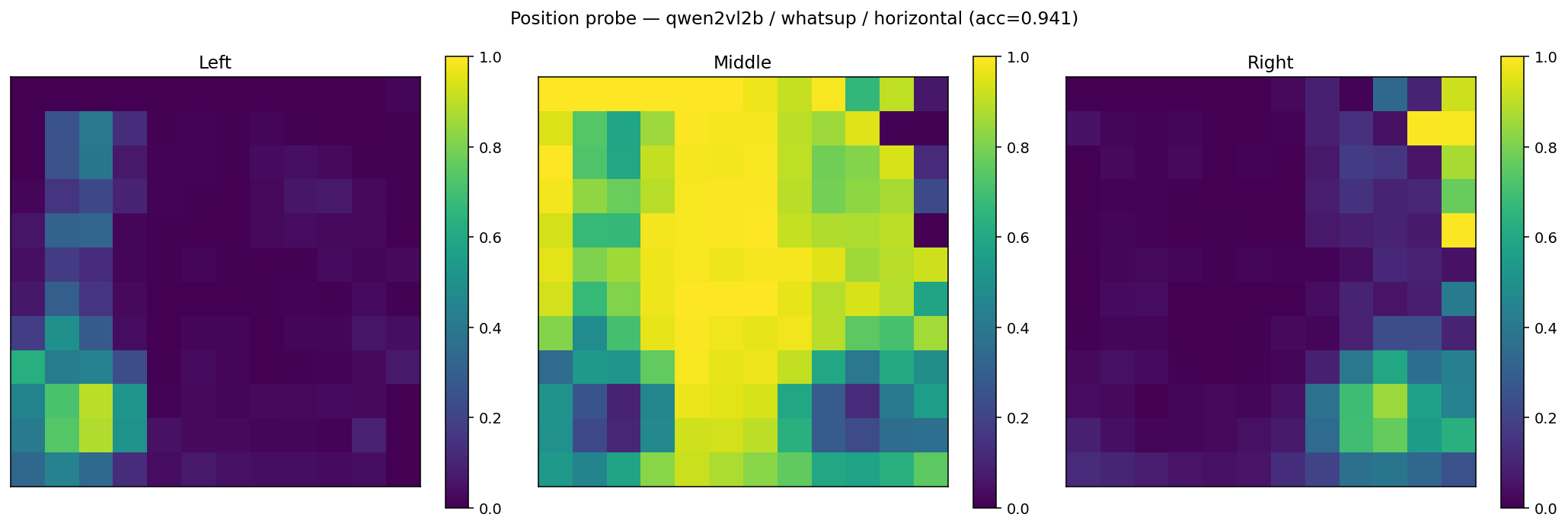}
    \caption{Position-probe predictions for \texttt{Qwen2-VL-2B-Instruct} on the Objects (horizontal, vertical) and What'sUp (horizontal) settings.}
    \label{fig:probe_qwen2vl2b_b}
\end{figure}

\begin{figure}[h]
    \centering
    \includegraphics[width=0.85\linewidth]{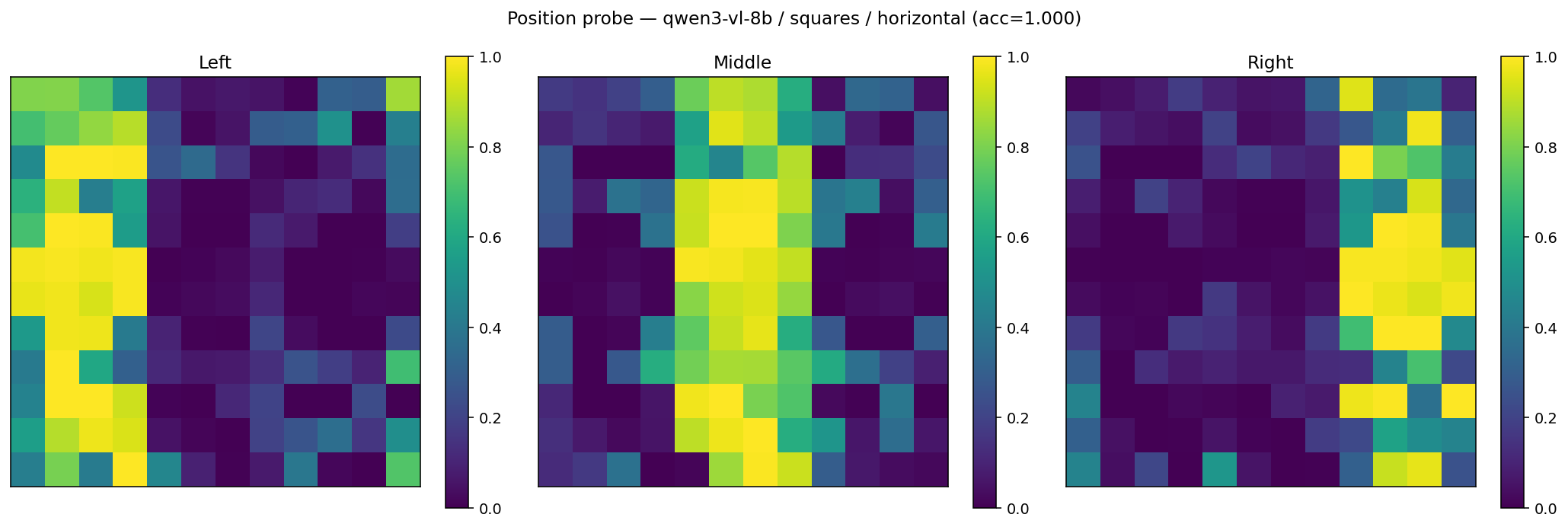}\\
    \includegraphics[width=0.85\linewidth]{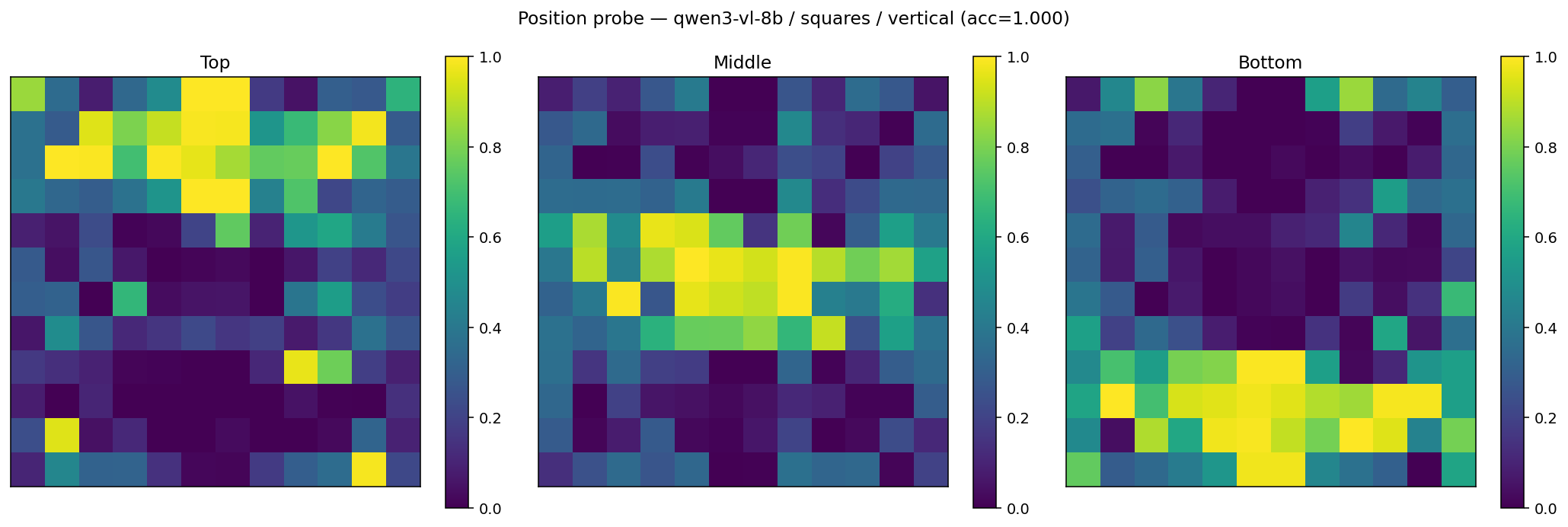}\\
    \includegraphics[width=0.85\linewidth]{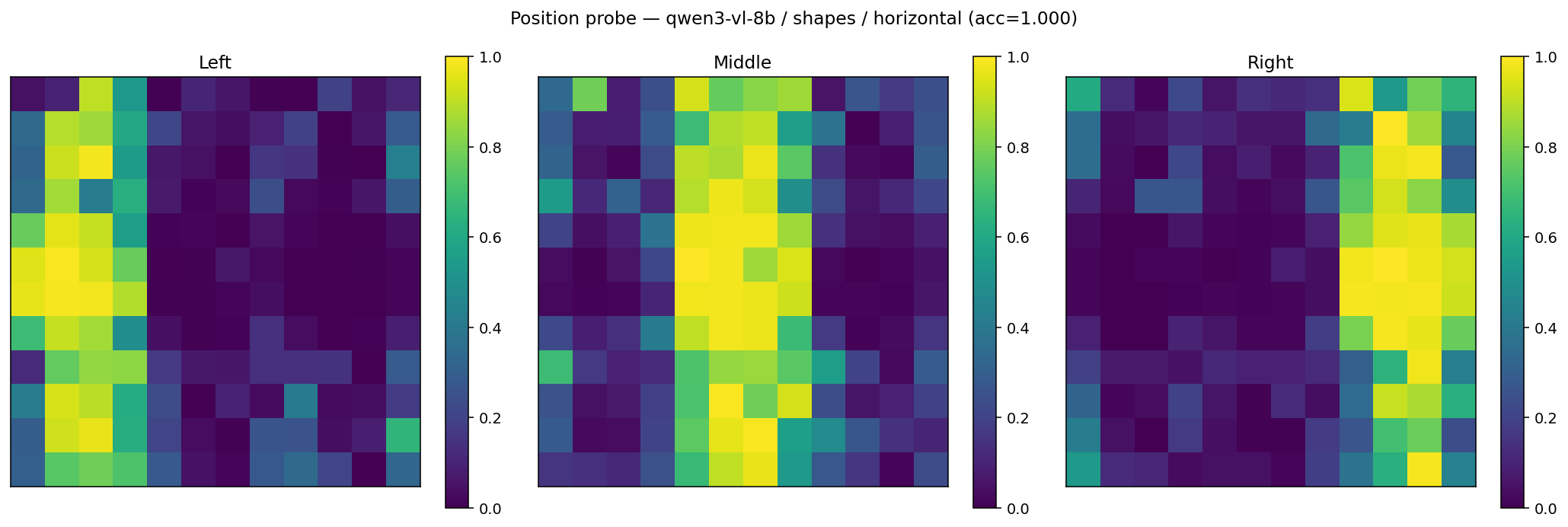}\\
    \includegraphics[width=0.85\linewidth]{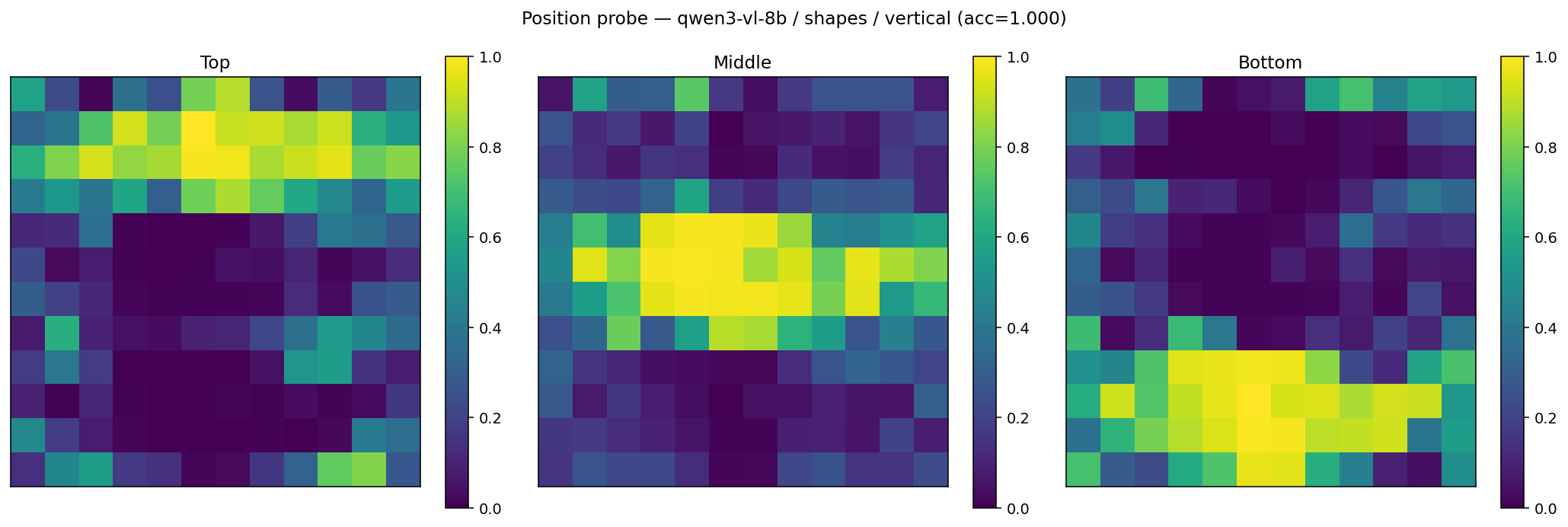}
    \caption{Position-probe predictions for \texttt{Qwen3-VL-8B-Instruct} on the Squares (horizontal, vertical) and Shapes (horizontal, vertical) settings.}
    \label{fig:probe_qwen3vl_a}
\end{figure}

\begin{figure}[h]
    \centering
    \includegraphics[width=0.85\linewidth]{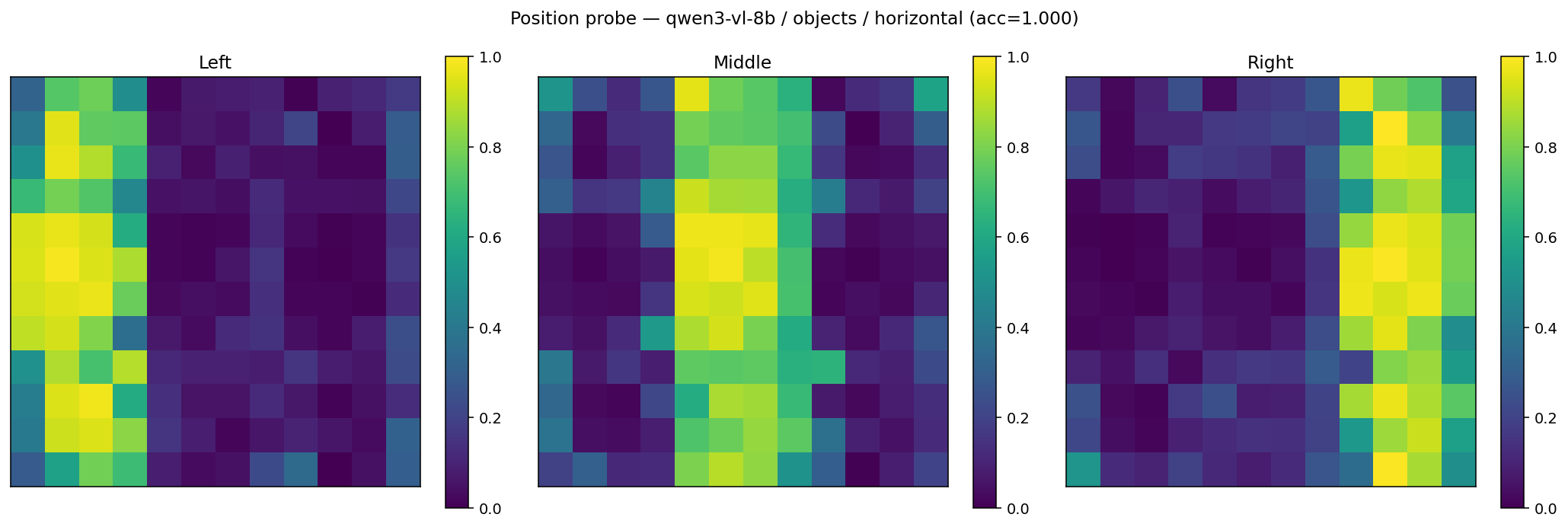}\\
    \includegraphics[width=0.85\linewidth]{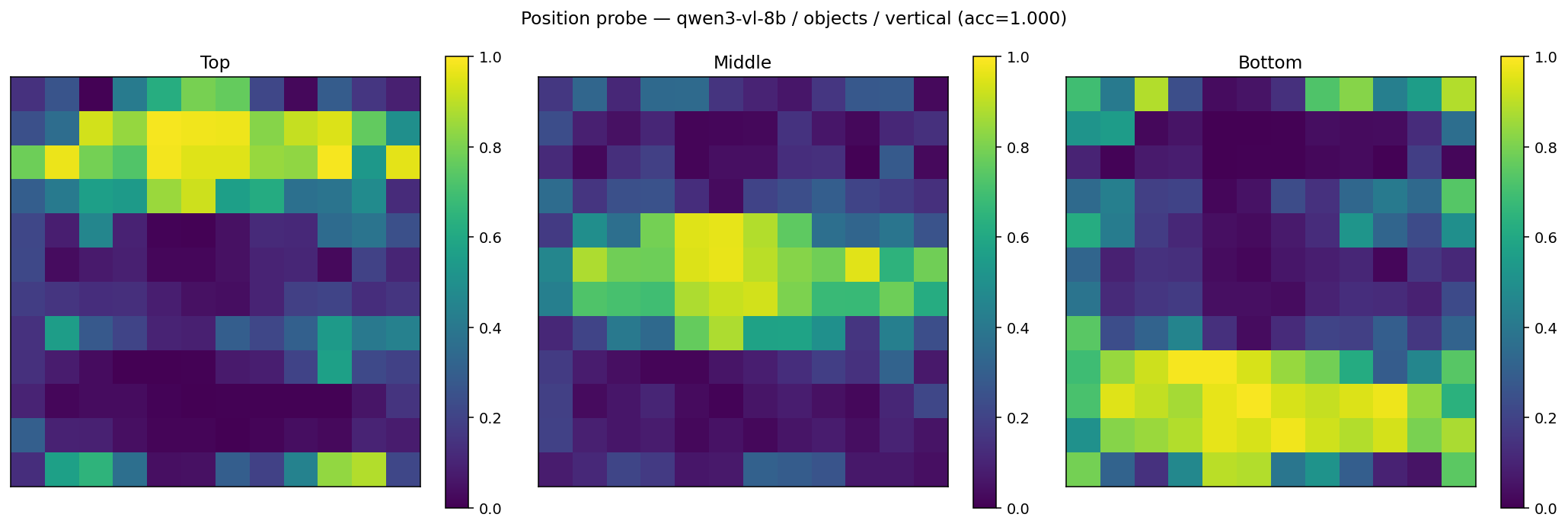}\\
    \includegraphics[width=0.85\linewidth]{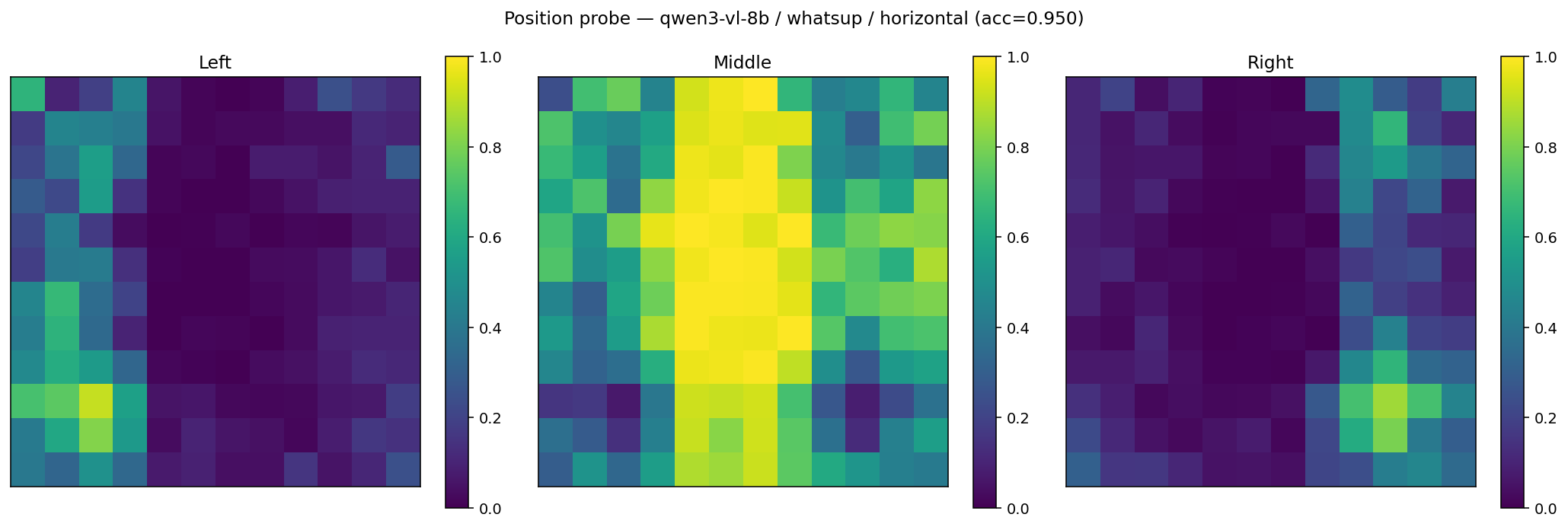}
    \caption{Position-probe predictions for \texttt{Qwen3-VL-8B-Instruct} on the Objects (horizontal, vertical) and What'sUp (horizontal) settings.}
    \label{fig:probe_qwen3vl_b}
\end{figure}

One of the main findings of this work is that ordering information from the vision encoder is not concentrated in a small set of square tokens; instead, it is distributed across neighboring background tokens as well. In this subsection, we present probing results across multiple settings and models, demonstrating the ubiquity of this phenomenon of diffused ordering information. Results are demonstrated in Fig.~\ref{fig:qwen_square_horizontal}-Fig.~\ref{fig:qwen_object_vertical}. We also show similar results with \texttt{Gemma-3-4b-it} in Fig.~\ref{fig:gemma_square_horizontal}-Fig.~\ref{fig:gemma_object_vertical}, and with \texttt{Qwen2-VL-2B-Instruct} and \texttt{Qwen3-VL-8B-Instruct} in Figs.~\ref{fig:probe_qwen2vl2b_a}--\ref{fig:probe_qwen3vl_b}

\clearpage

\subsection{Probing on grids}
\label{app:grid}

\begin{figure}[h]
    \centering
    \centering
    \begin{tabular}{ccc}
        \includegraphics[width=0.3\linewidth]{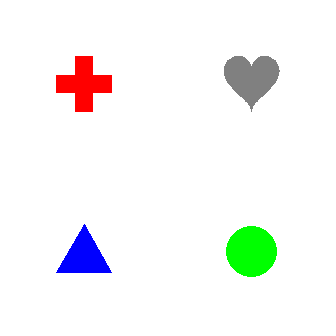}
        &
        \includegraphics[width=0.3\linewidth]{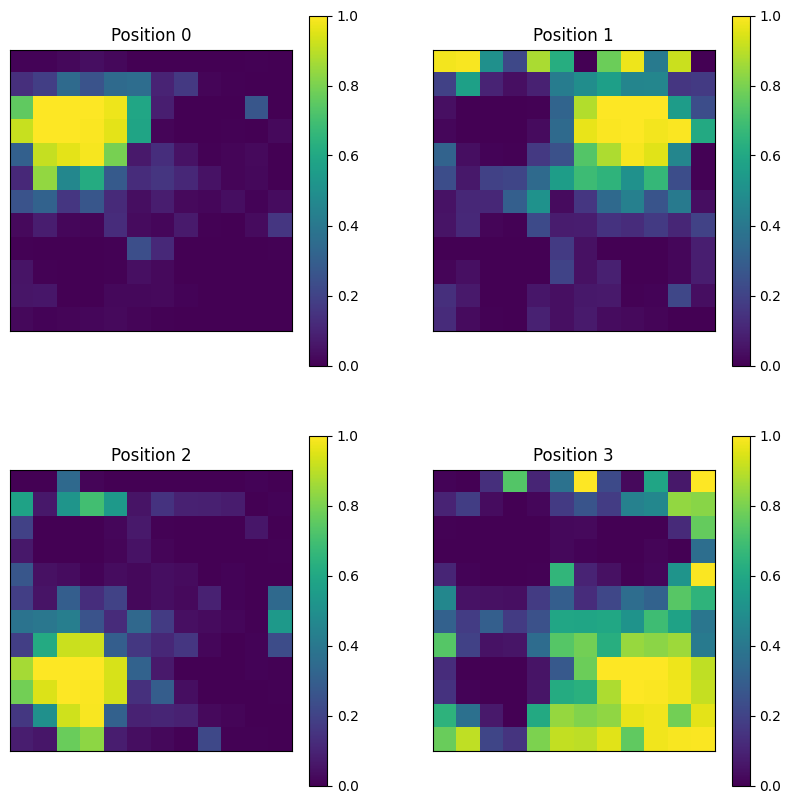}
        &
        \includegraphics[width=0.3\linewidth]{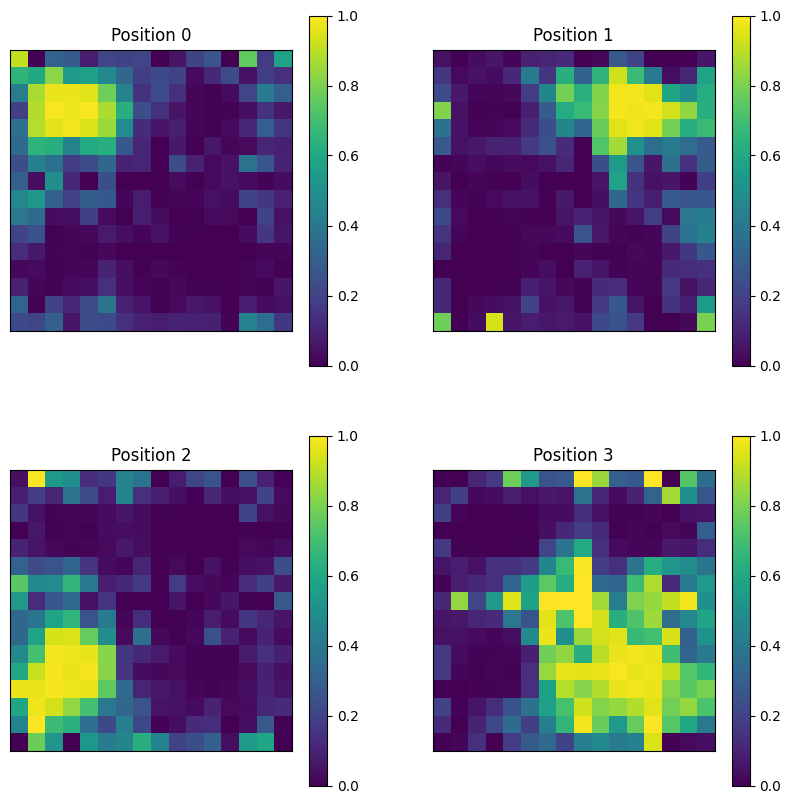} \\
    \end{tabular}
    \caption{Qwen and Gemma probing results on 2x2 shapes grid.}
    \label{fig:2x2_grid}
\end{figure}

\begin{figure}[h]
    \centering
    \centering
    \begin{tabular}{ccc}
        \includegraphics[width=0.3\linewidth]{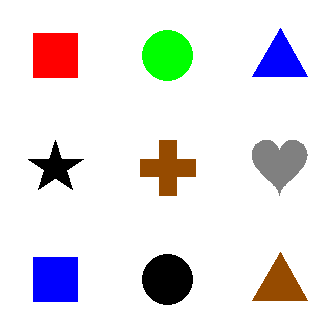}
        &
        \includegraphics[width=0.3\linewidth]{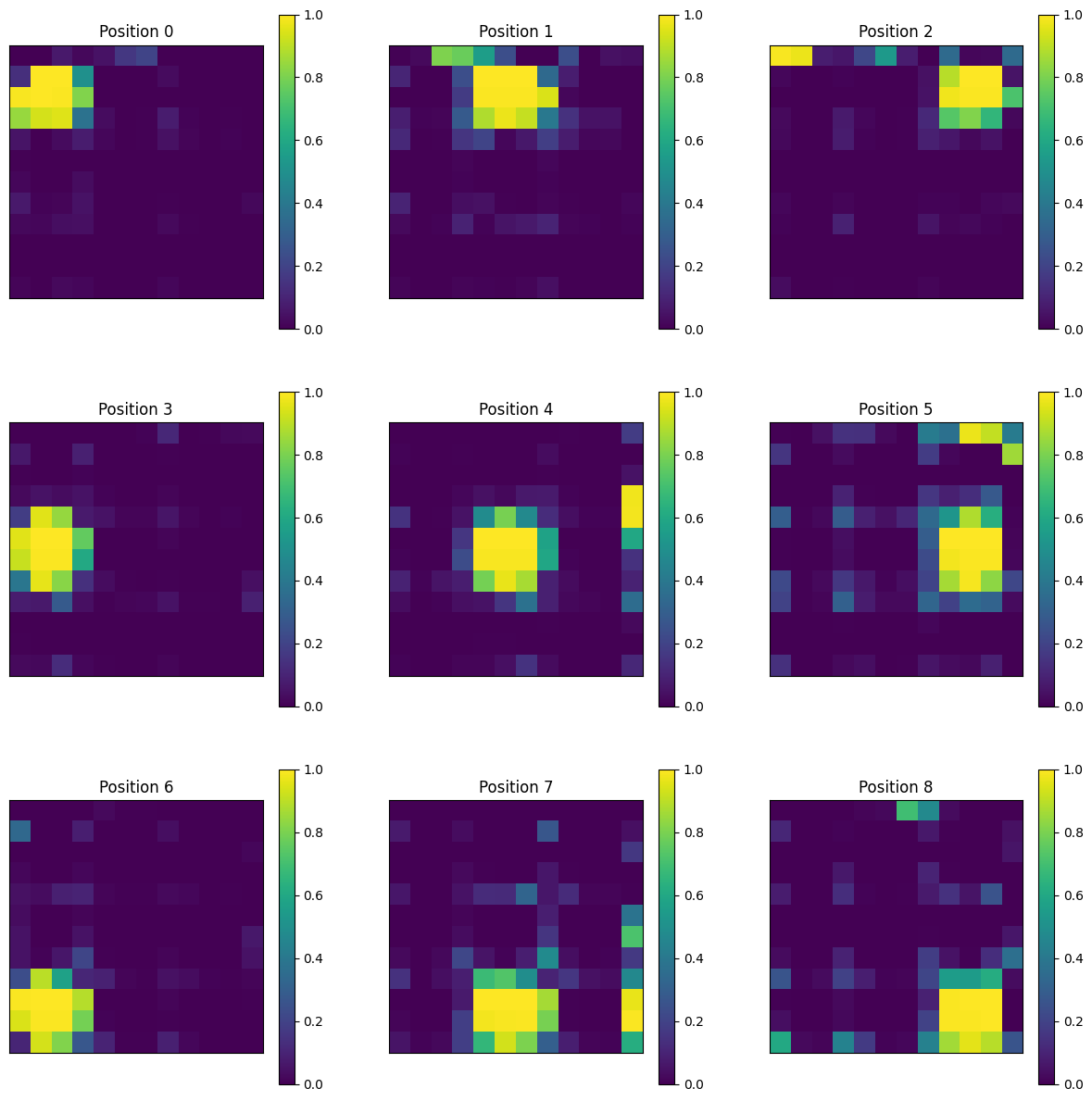}
        &
        \includegraphics[width=0.3\linewidth]{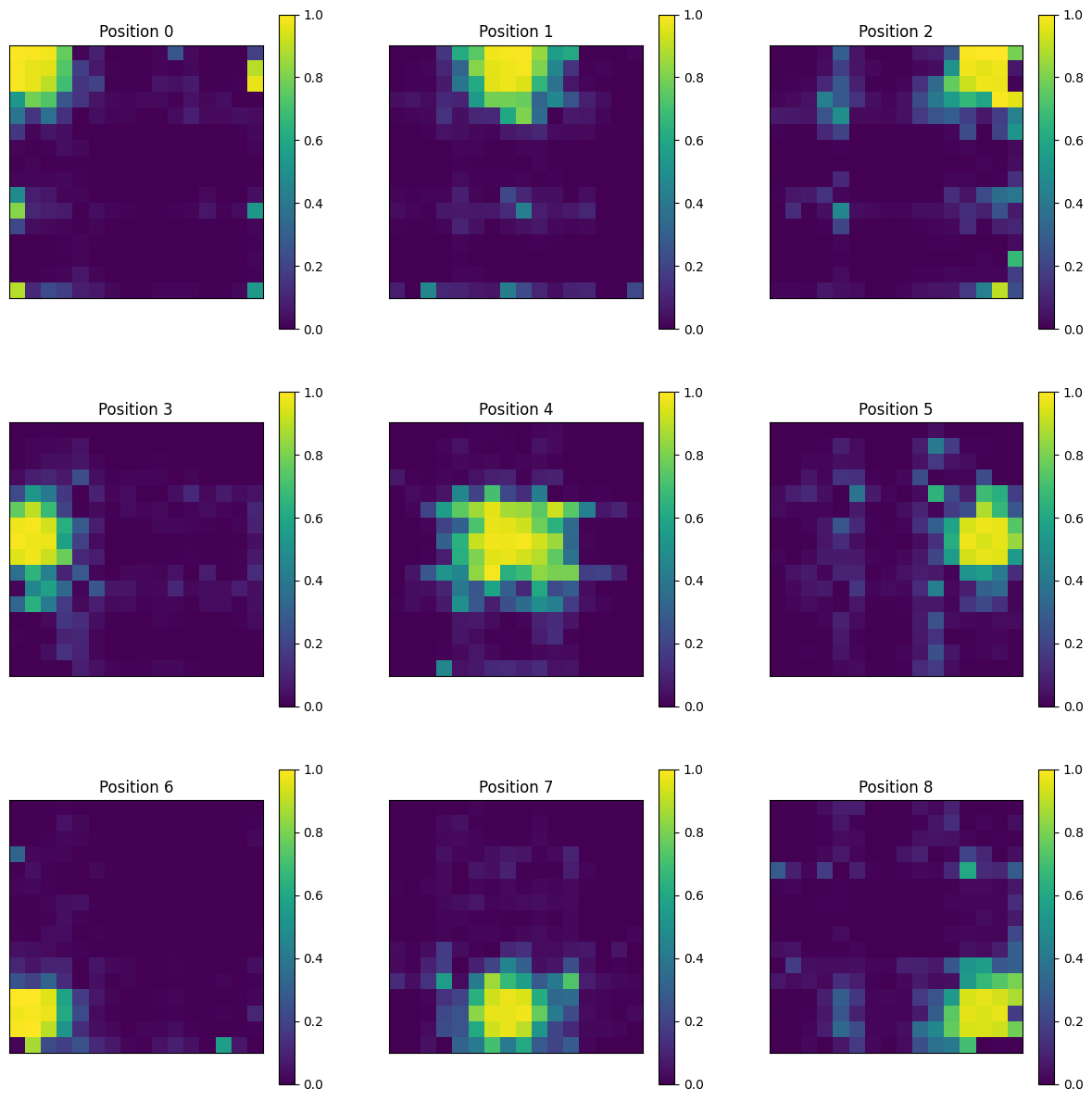} \\
    \end{tabular}
    \caption{Qwen and Gemma probing results on 3x3 shapes grid.}
    \label{fig:3x3_grid}
\end{figure}

We extend the probing results from subsection~\ref{app:probing} to grid configurations consisting of both horizontal and vertical relationships between objects, with findings presented in Fig.~\ref{fig:2x2_grid} and Fig.~\ref{fig:3x3_grid}. Probes are trained on $2 \times 2$ token objects for \texttt{Qwen2-VL-7B-Instruct} and $3 \times 3$ token objects for \texttt{Gemma3-4b-it}, with results exhibiting diffusion of positional information to surrounding tokens. 

\clearpage
\subsection{Probing for What'sUp}
\label{app:prob_whatsup}
\begin{figure}[h]
    \centering
    \centering
    \begin{tabular}{ccc}
        \includegraphics[width=0.32\linewidth]{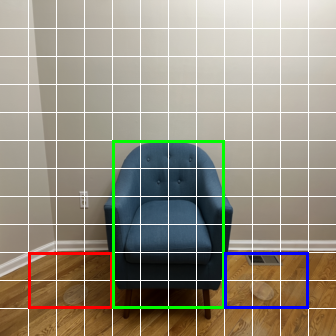}
        &
        \includegraphics[width=0.32\linewidth]{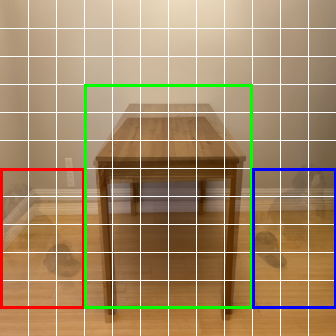}
        &
        \includegraphics[width=0.32\linewidth]{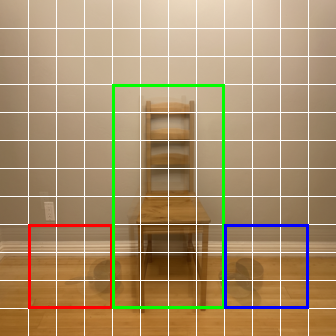} \\
        
        \includegraphics[width=0.32\linewidth]{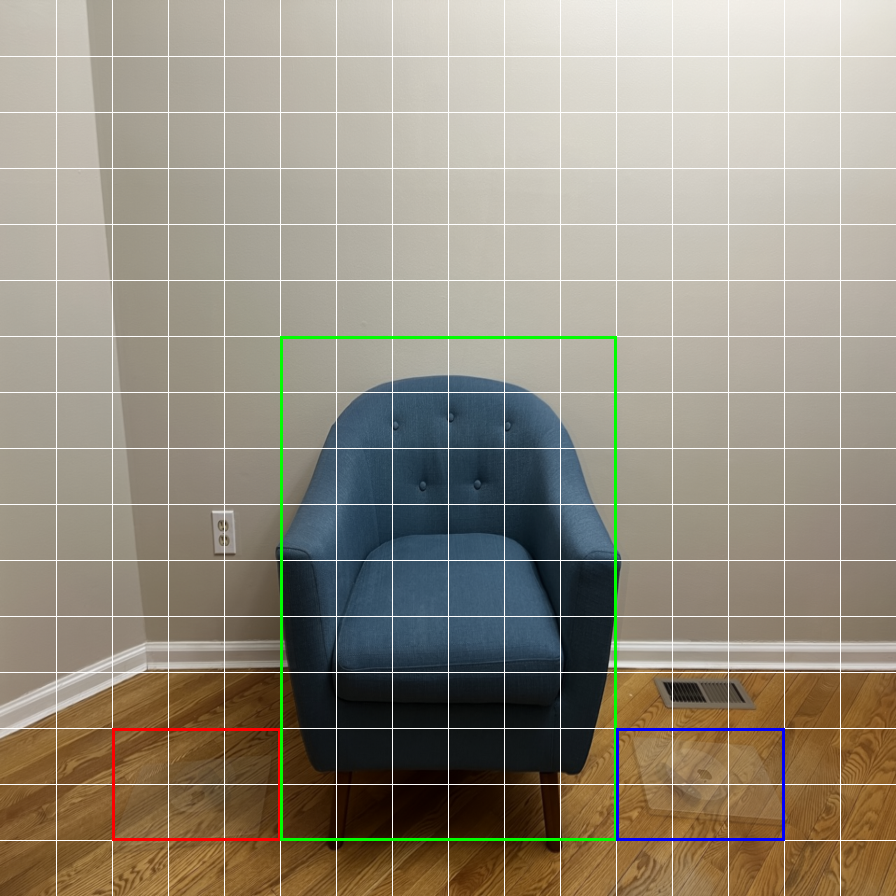}
        &
        \includegraphics[width=0.32\linewidth]{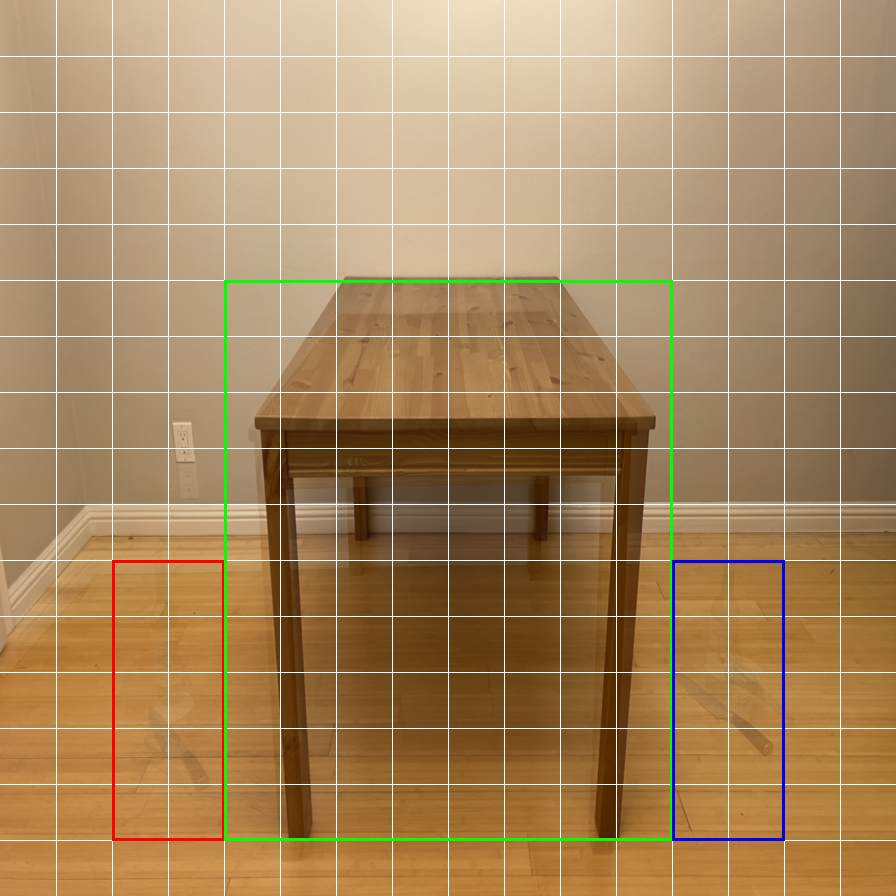}
        &
        \includegraphics[width=0.32\linewidth]{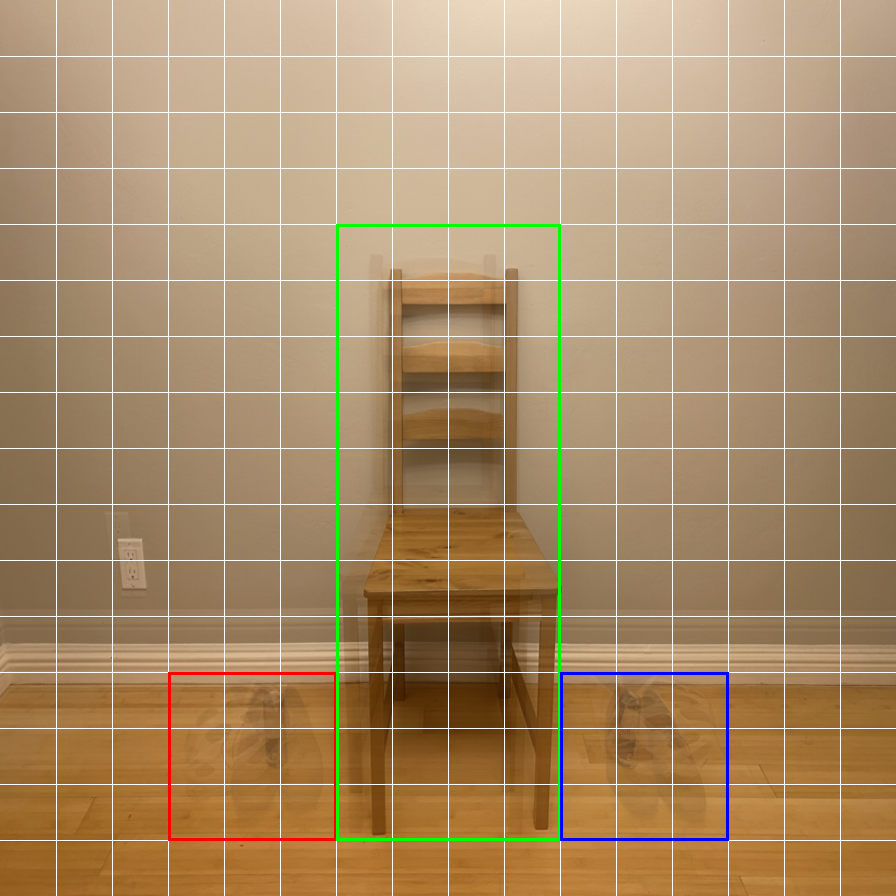}
        
    \end{tabular}
    \caption{Preprocessing for What’sUp images: constructing bounding boxes around each object.}
    \label{app:whatsup_preprocessing}
\end{figure}

\begin{figure}[h]
    \centering
    \includegraphics[width=0.9\linewidth]{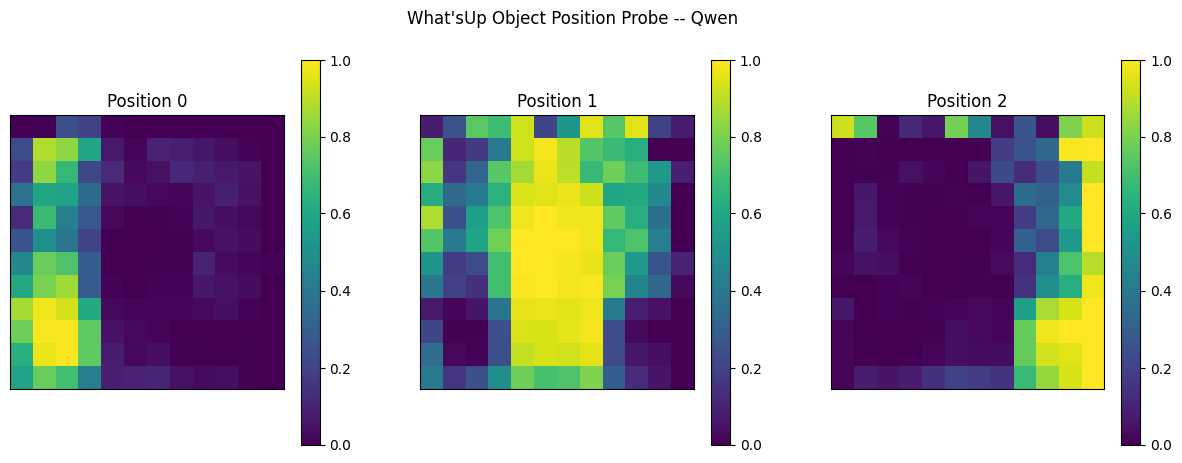}
    \caption{Qwen What'sUp probe}
    \includegraphics[width=0.9\linewidth]{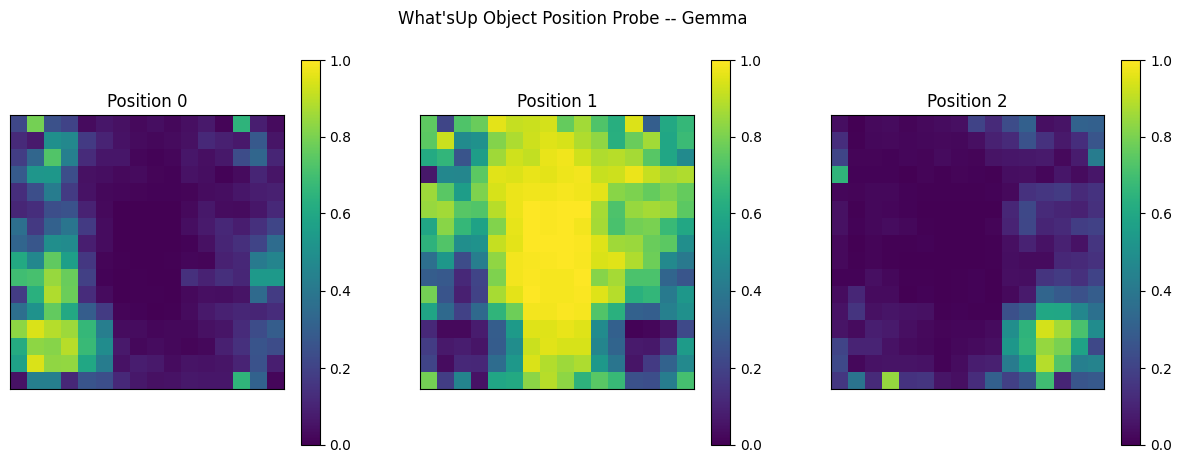}
    \caption{Gemma What'sUp probe}
    
    \label{fig:both_model_whatsup}
\end{figure}

Finally, we also show that the ordering information is diffused across background tokens in real-world images from What'sUp. Training data is extracted from the embeddings within the bounding boxes outlined in Fig.~\ref{app:whatsup_preprocessing}. Results are consistent with previous findings, as illustrated in Fig.~\ref{fig:both_model_whatsup}.

\clearpage
\subsection{Experimental results with realistic background}
\label{app:probe_real_back}
We also conduct probing experiments (Sec.~\ref{subsec:probing}) and intervention experiments (Sec.~\ref{subsec:vision_encoder}) using real background images to demonstrate that ordering information remains distributed across background tokens in real-world settings. We tested $15$ different realistic backgrounds, and found a consistent behavior across all.

\begin{figure}[h]
    \centering
    \begin{tabular}{cc}
    \includegraphics[width=0.20\linewidth]{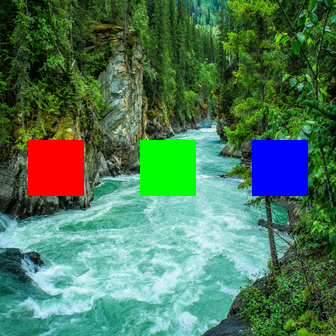} &
        \includegraphics[width=0.75\linewidth]{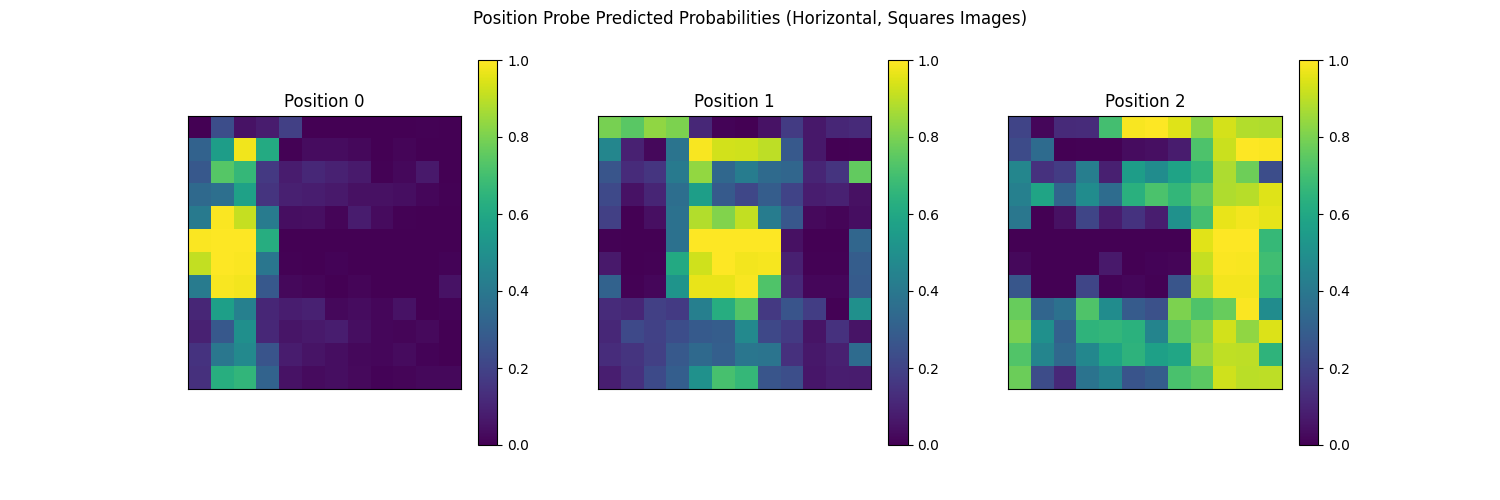} \\
    \end{tabular}
    \caption{Horizontal position probe for square images with real background, Qwen}
    \label{fig:probe_qwen_square_horizonatal_real_background}
\end{figure}

\begin{figure}[h]
    \centering
    \begin{tabular}{cc}
    \includegraphics[width=0.20\linewidth]{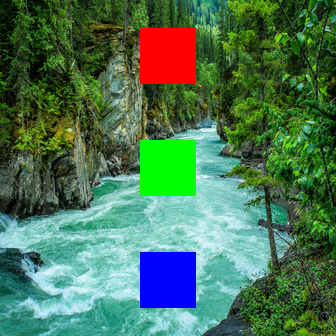} &
        \includegraphics[width=0.75\linewidth]{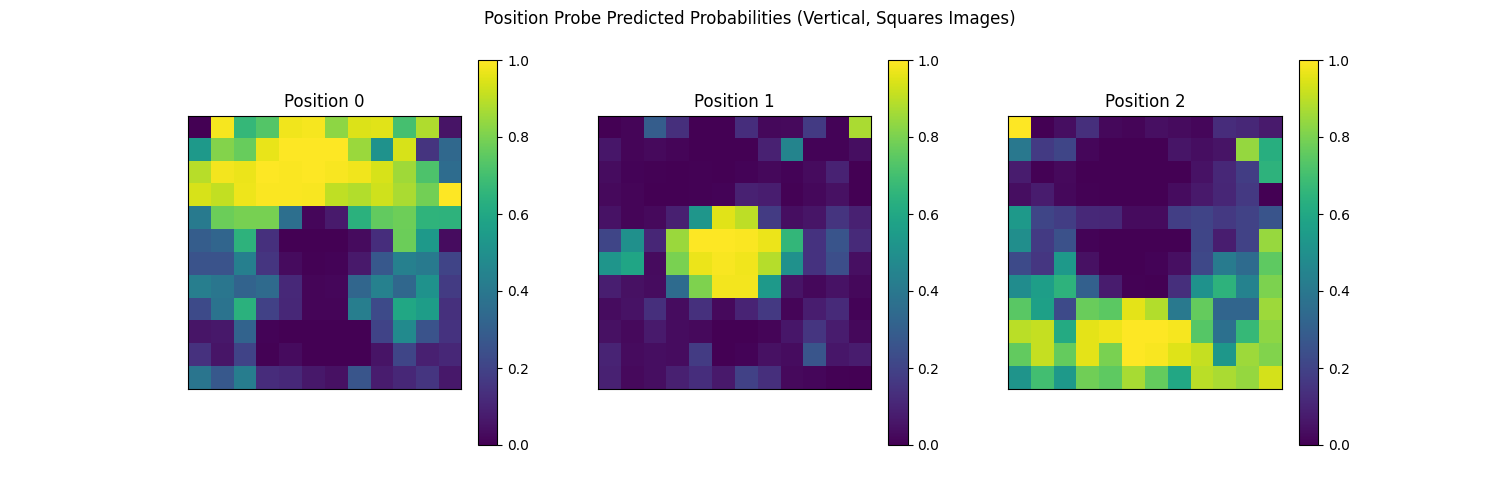} \\
    \end{tabular}
    \caption{Vertical position probe for square images with real background, Qwen}
    \label{fig:probe_qwen_square_vertical_real_background}
\end{figure}

\begin{figure}[h]
    \centering
    \begin{tabular}{cc}
    \includegraphics[width=0.20\linewidth]{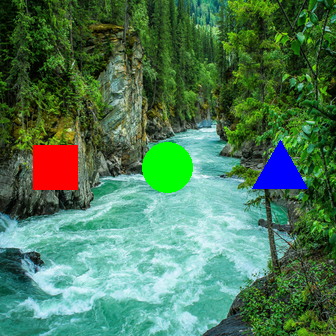} &
        \includegraphics[width=0.75\linewidth]{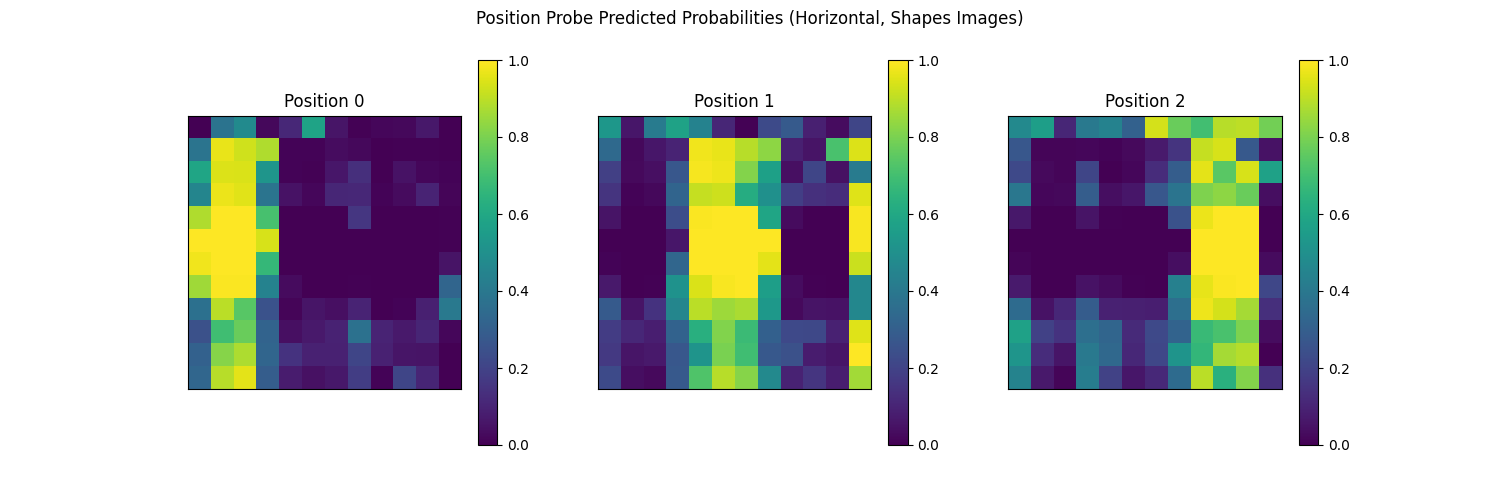} \\
    \end{tabular}
    \caption{Horizontal position probe for shapes images with real background, Qwen}
    \label{fig:probe_qwen_shape_horizonatal_real_background}
\end{figure}

\begin{figure}[h]
    \centering
    \begin{tabular}{cc}
    \includegraphics[width=0.20\linewidth]{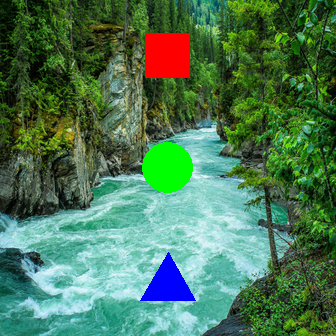} &
        \includegraphics[width=0.75\linewidth]{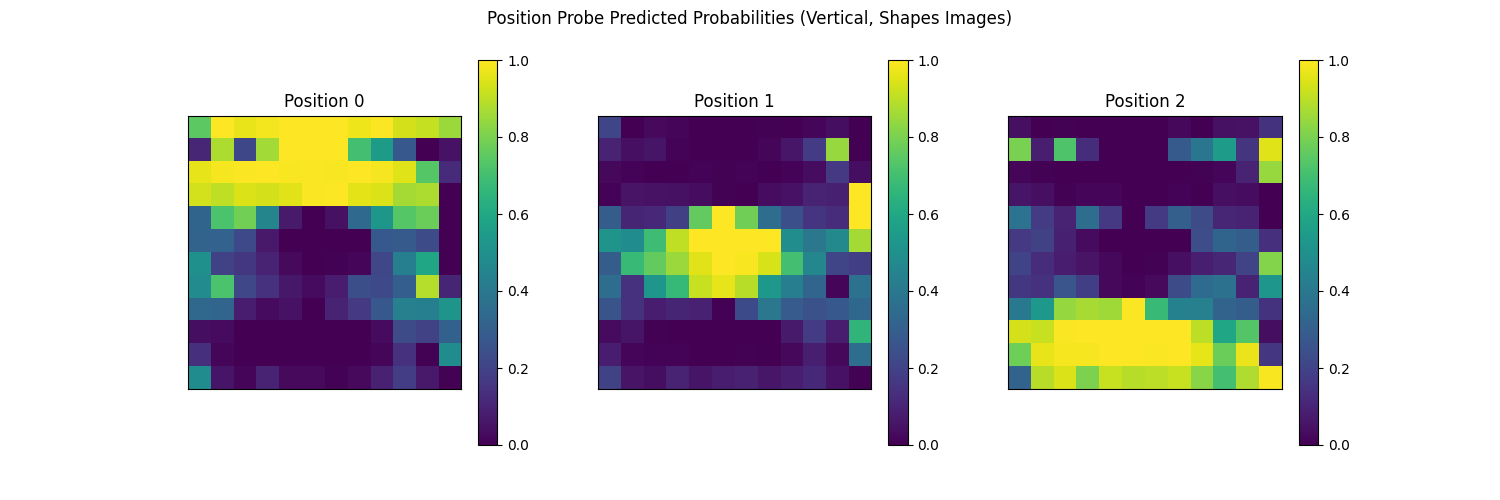} \\
    \end{tabular}
    \caption{Vertical position probe for shapes images with real background, Qwen}
    \label{fig:probe_qwen_shape_vertical_real_background}
\end{figure}

\begin{figure}[h]
    \centering
    \begin{tabular}{cc}
    \includegraphics[width=0.20\linewidth]{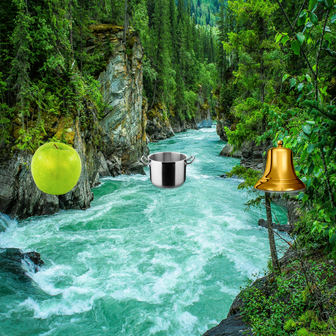} &
        \includegraphics[width=0.75\linewidth]{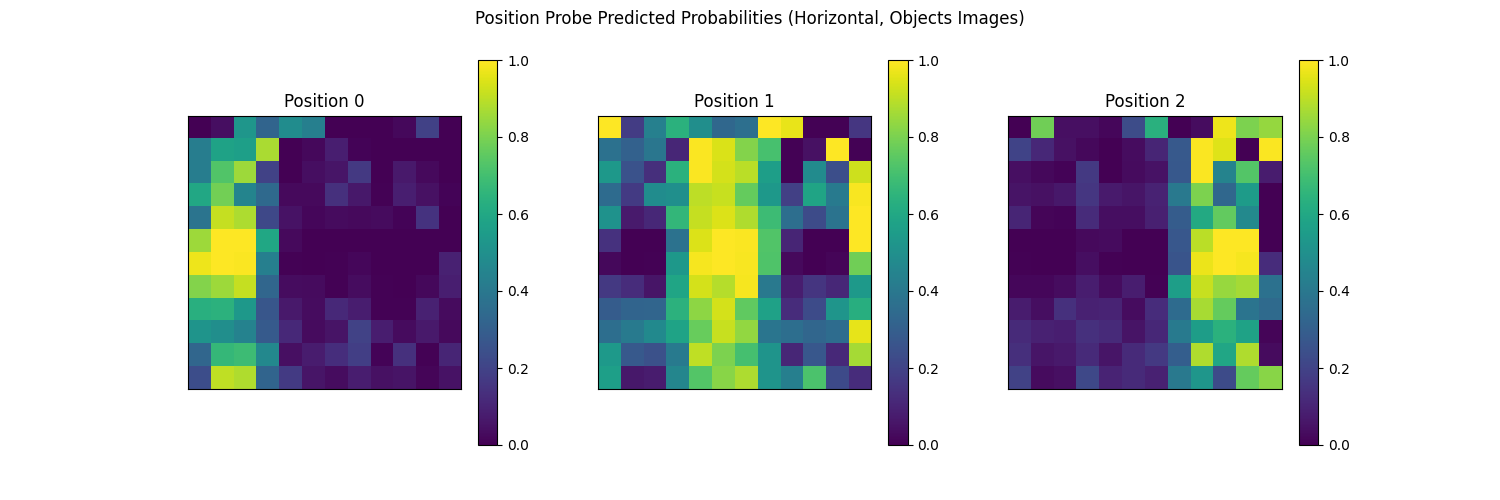} \\
    \end{tabular}
    \caption{Horizontal position probe for objects images with real background, Qwen}
    \label{fig:probe_qwen_objects_horizonatal_real_background}
\end{figure}

\begin{figure}[h]
    \centering
    \begin{tabular}{cc}
    \includegraphics[width=0.20\linewidth]{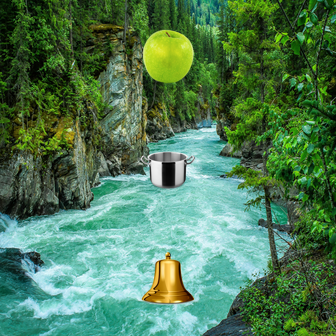} &
        \includegraphics[width=0.75\linewidth]{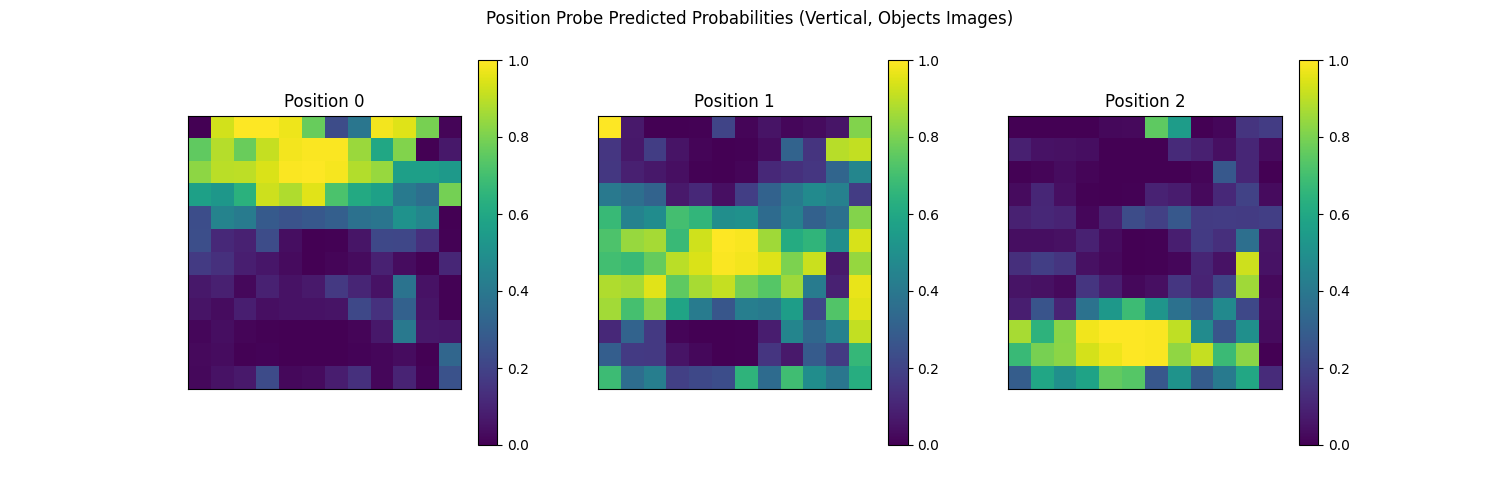} \\
    \end{tabular}
    \caption{Vertical position probe for objects images with real background, Qwen}
    \label{fig:probe_qwen_objects_vertical_real_background}
\end{figure}

\begin{figure}[h]
    \centering
    \begin{tabular}{cc}
    \includegraphics[width=0.20\linewidth]{figures_main/squares_horizontal_real_background.png} &
        \includegraphics[width=0.75\linewidth]{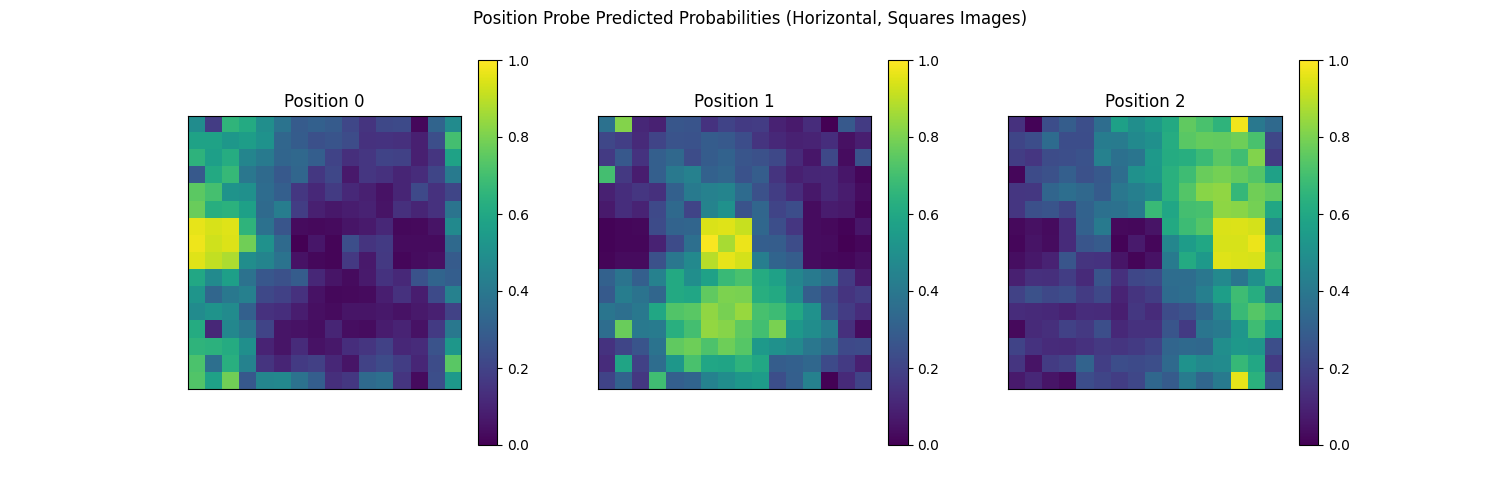} \\
    \end{tabular}
    \caption{Horizontal position probe for square images with real background, Gemma}
    \label{fig:probe_gemma_square_horizonatal_real_background}
\end{figure}

\begin{figure}[h]
    \centering
    \begin{tabular}{cc}
    \includegraphics[width=0.20\linewidth]{figures_main/squares_vertical_real_background.png} &
        \includegraphics[width=0.75\linewidth]{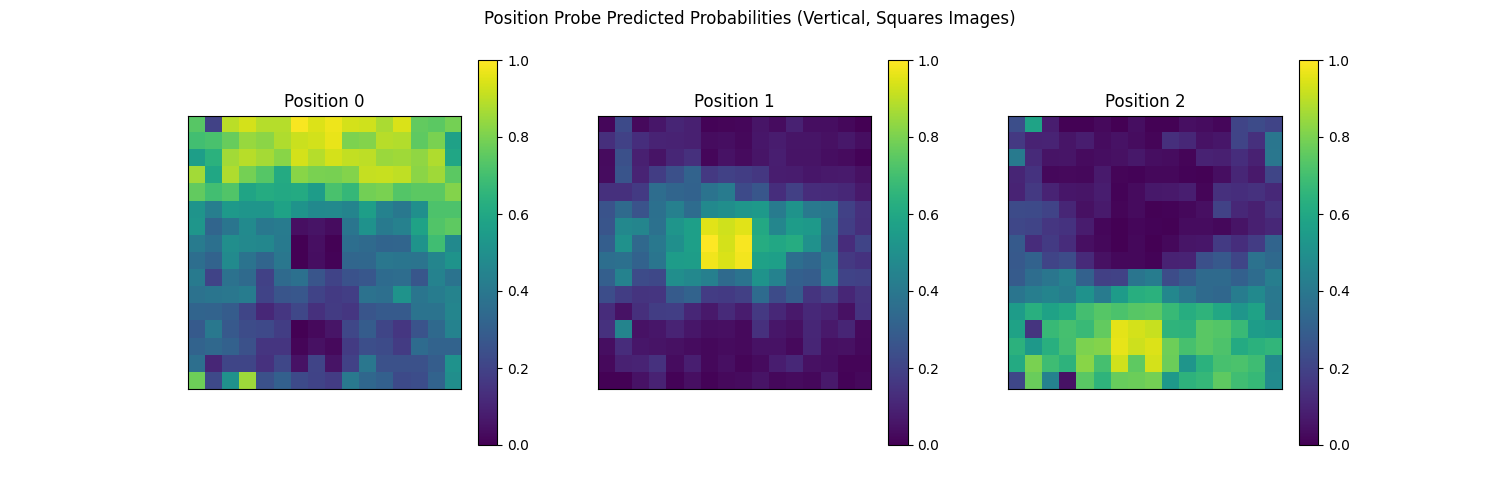} \\
    \end{tabular}
    \caption{Vertical position probe for square images with real background, Gemma}
    \label{fig:probe_gemma_square_vertical_real_background}
\end{figure}

\begin{figure}[h]
    \centering
    \begin{tabular}{cc}
    \includegraphics[width=0.20\linewidth]{figures_main/shapes_horizontal_real_background.png} &
        \includegraphics[width=0.75\linewidth]{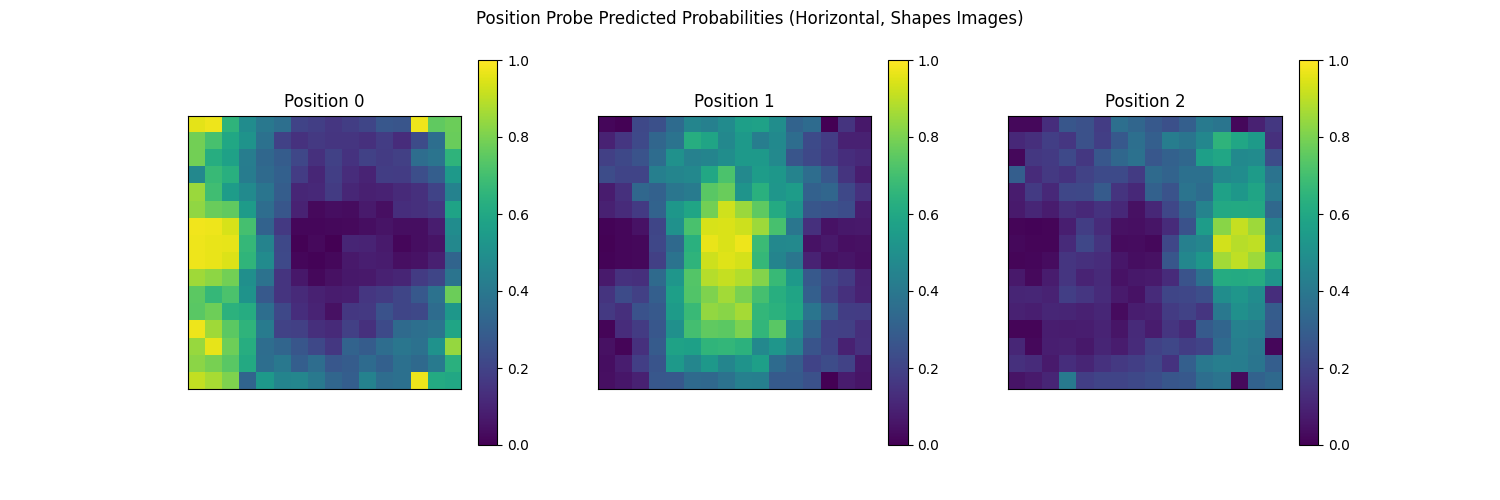} \\
    \end{tabular}
    \caption{Horizontal position probe for shapes images with real background, Gemma}
    \label{fig:probe_gemma_shape_horizonatal_real_background}
\end{figure}

\begin{figure}[h]
    \centering
    \begin{tabular}{cc}
    \includegraphics[width=0.20\linewidth]{figures_main/shapes_vertical_real_background.png} &
        \includegraphics[width=0.75\linewidth]{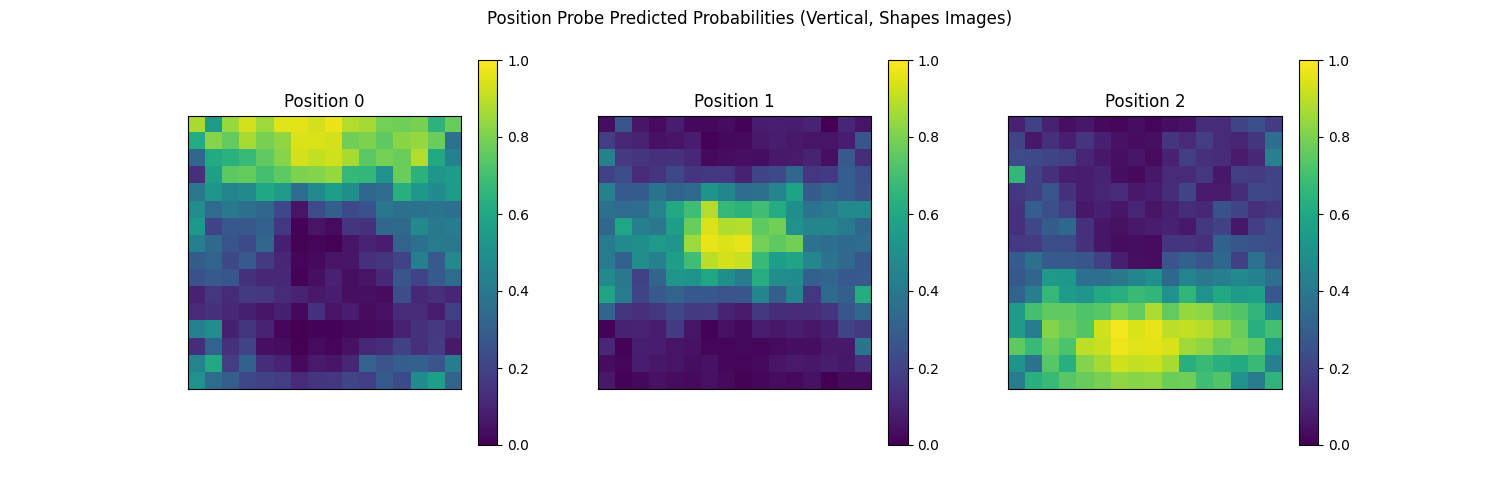} \\
    \end{tabular}
    \caption{Vertical position probe for shapes images with real background, Gemma}
    \label{fig:probe_gemma_shape_vertical_real_background}
\end{figure}

\begin{figure}[h]
    \centering
    \begin{tabular}{cc}
    \includegraphics[width=0.20\linewidth]{figures_main/objects_horizontal_real_background.png} &
        \includegraphics[width=0.75\linewidth]{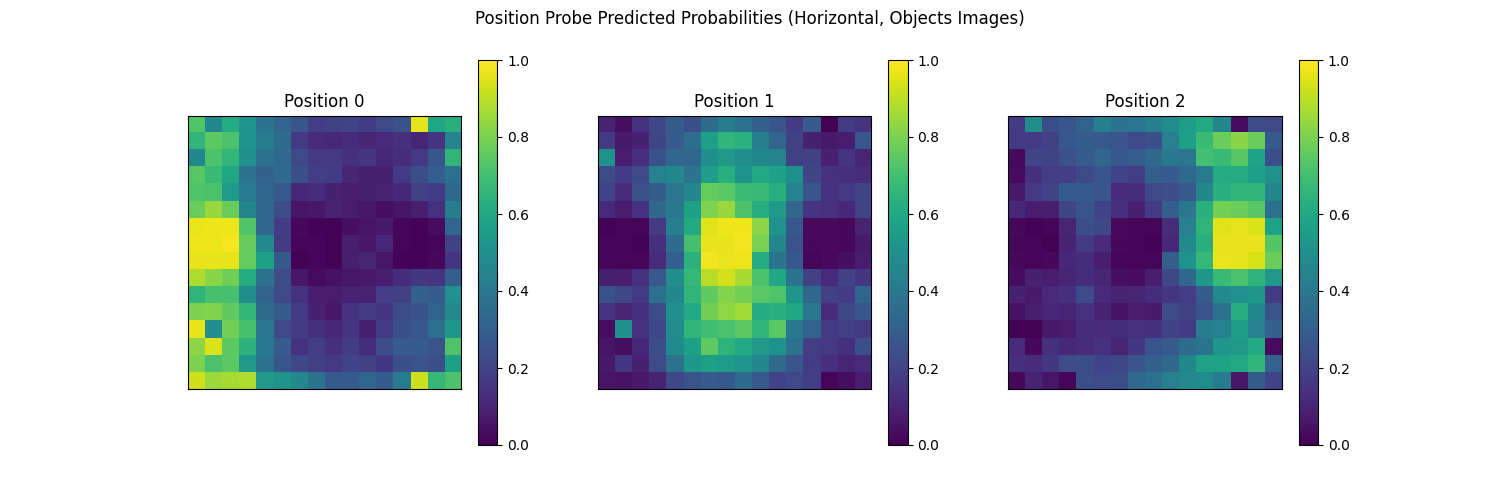} \\
    \end{tabular}
    \caption{Horizontal position probe for objects images with real background, Gemma}
    \label{fig:probe_gemma_objects_horizonatal_real_background}
\end{figure}

\begin{figure}[h]
    \centering
    \begin{tabular}{cc}
    \includegraphics[width=0.20\linewidth]{figures_main/objects_vertical_real_background.png} &
        \includegraphics[width=0.75\linewidth]{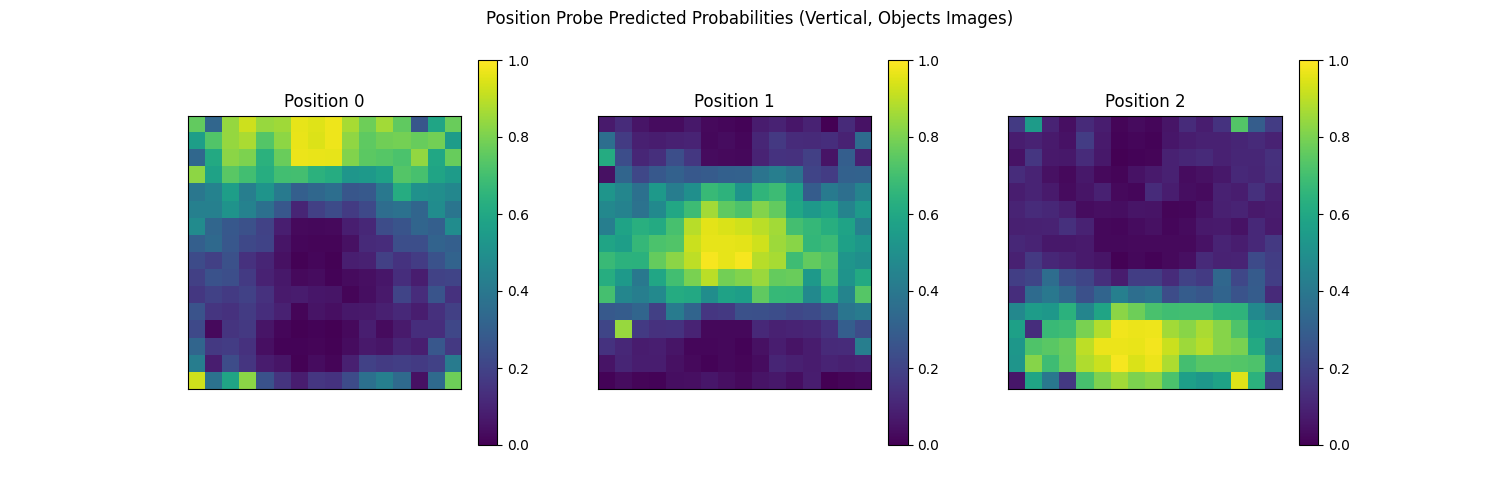} \\
    \end{tabular}
    \caption{Vertical position probe for objects images with real background, Gemma}
    \label{fig:probe_gemma_objects_vertical_real_background}
\end{figure}

\begin{figure}[h]
    \centering
    \includegraphics[width=1\linewidth]{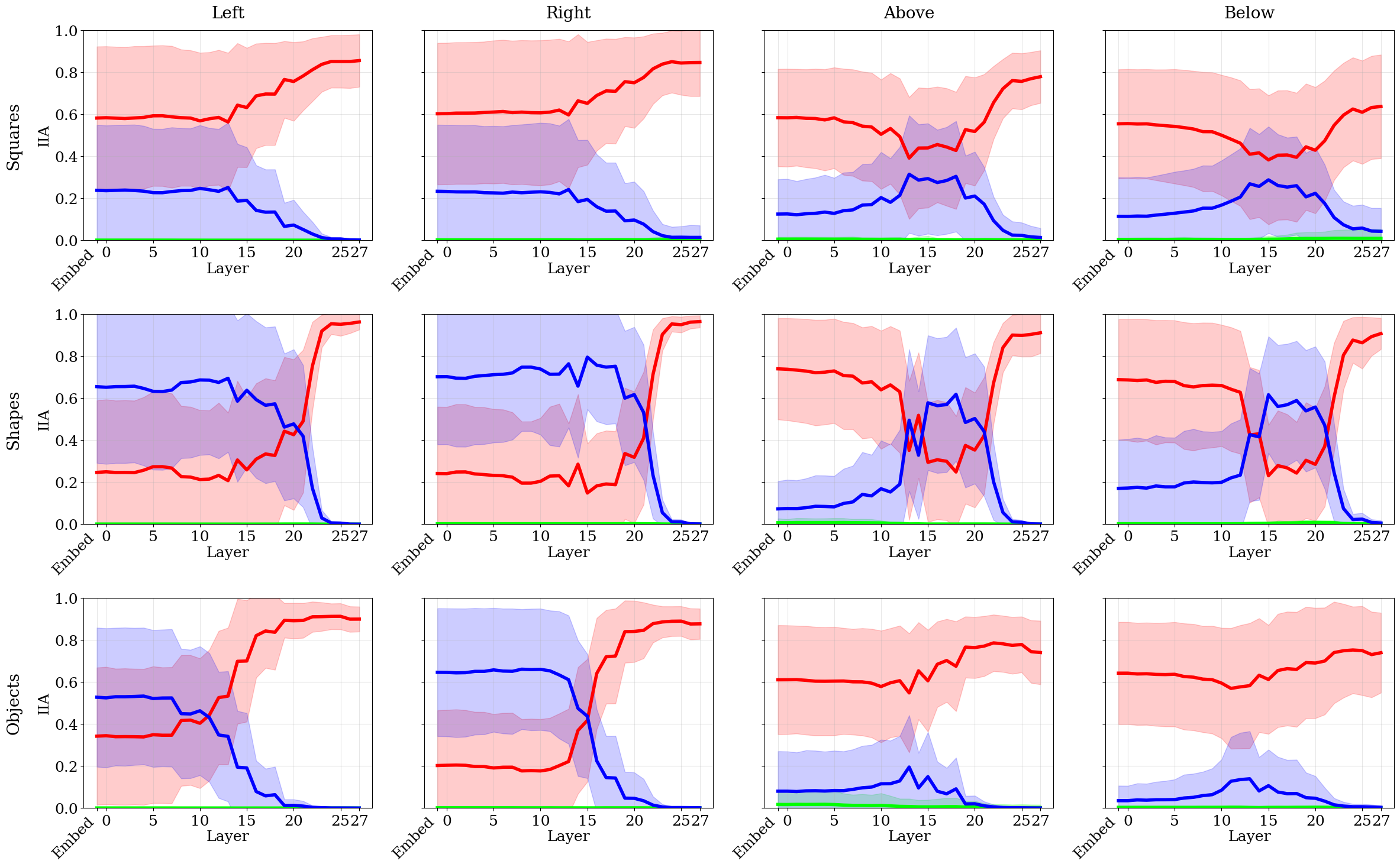}
    \caption{Interchange interventions applied only to visual tokens corresponding to
the squares in \texttt{Qwen2-VL-7B-Instruct} model, with real background images.}
    \label{fig:qwen_criss_cross_squares_real_backgrounds}
\end{figure}

\begin{figure}[h]
    \centering
    \includegraphics[width=1\linewidth]{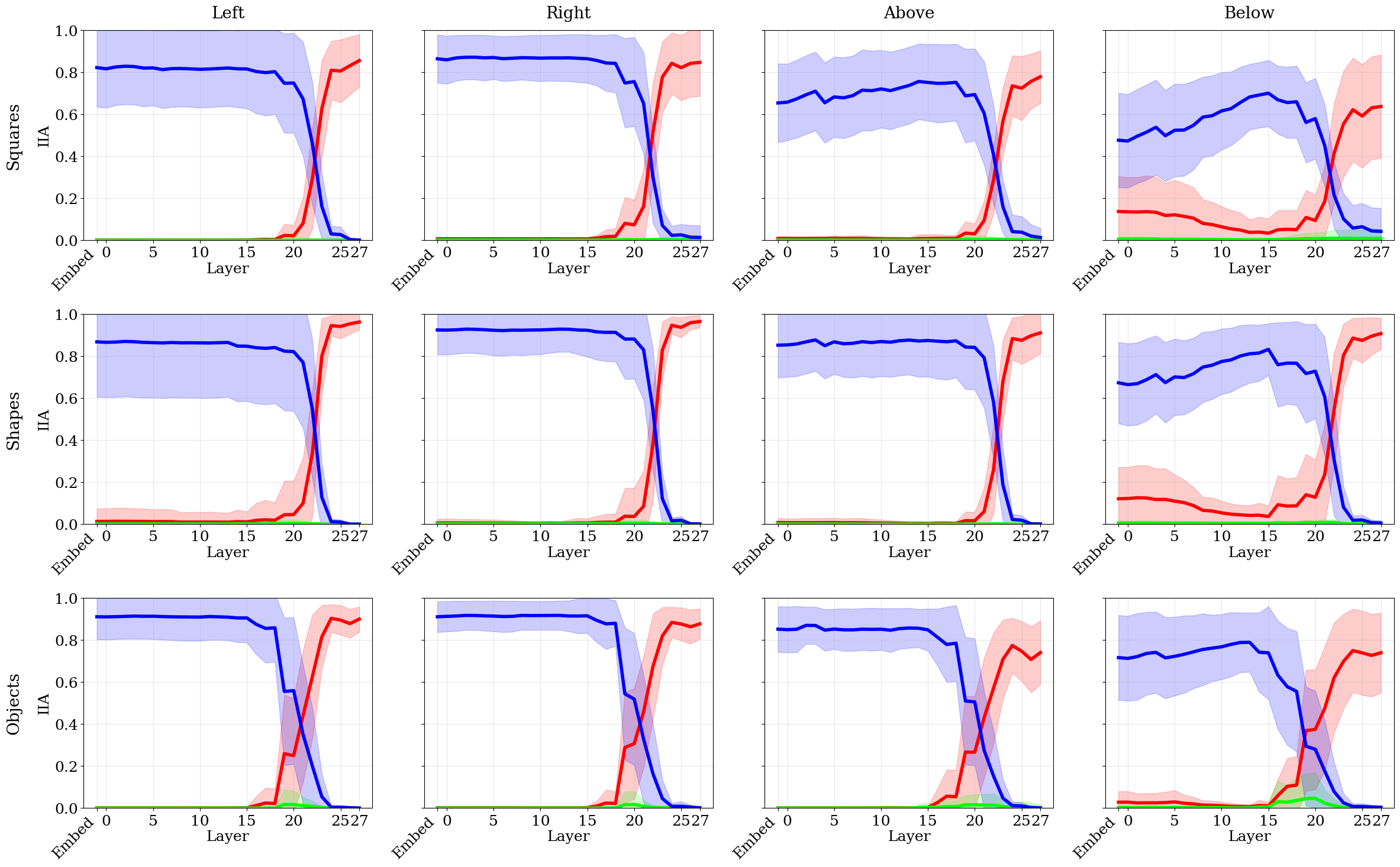}
    \caption{Interchange interventions applied only to visual tokens corresponding to
the strip in \texttt{Qwen2-VL-7B-Instruct} model, with real background images.}
    \label{fig:qwen_criss_cross_strips_real_backgrounds}
\end{figure}

\begin{figure}[h]
    \centering
    \includegraphics[width=1\linewidth]{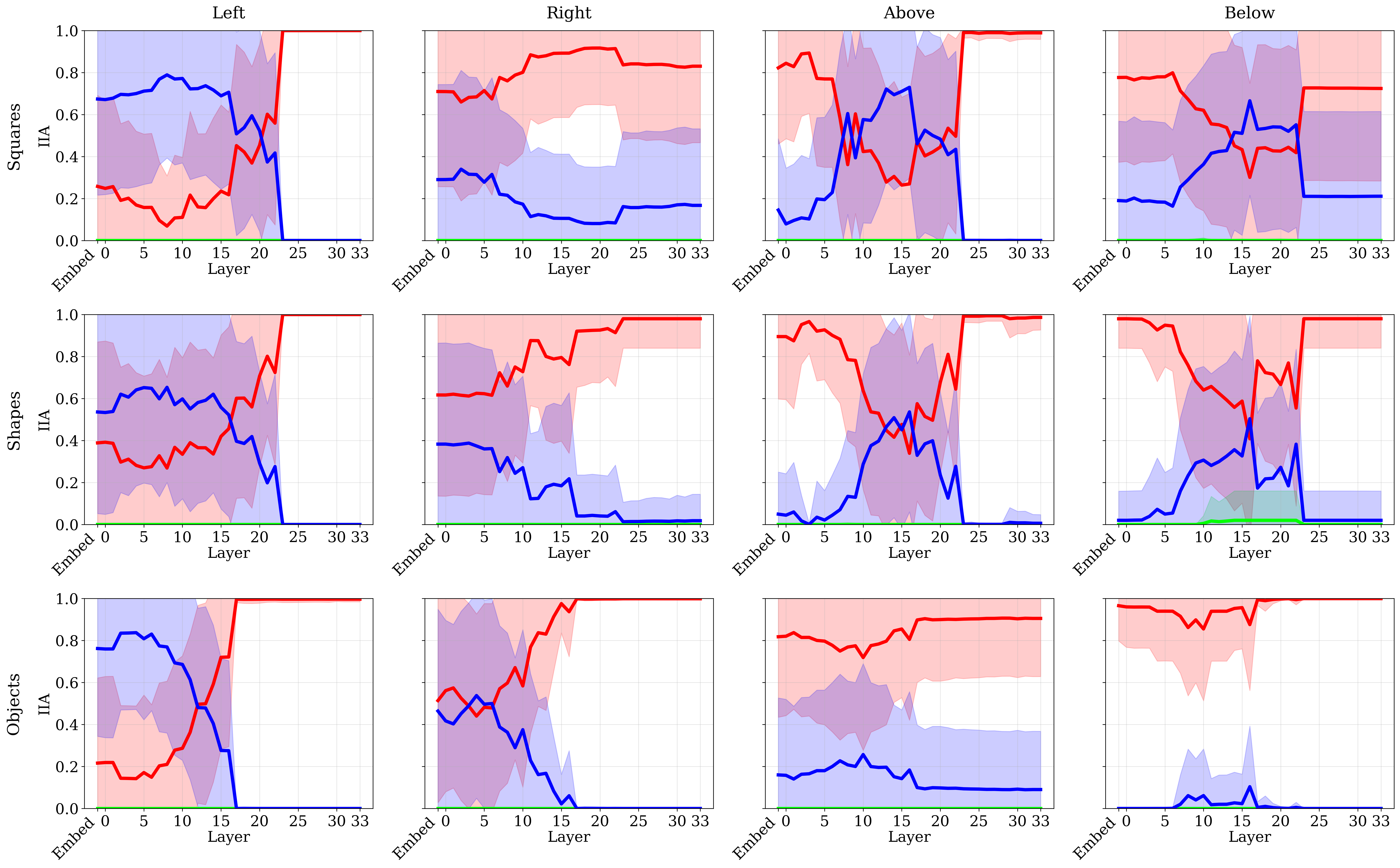}
    \caption{Interchange interventions applied only to visual tokens corresponding to
the squares in \texttt{Gemma-3-4b-it} model, with real background images.}
    \label{fig:gemma_criss_cross_squares_real_backgrounds}
\end{figure}

\begin{figure}[h]
    \centering
    \includegraphics[width=1\linewidth]{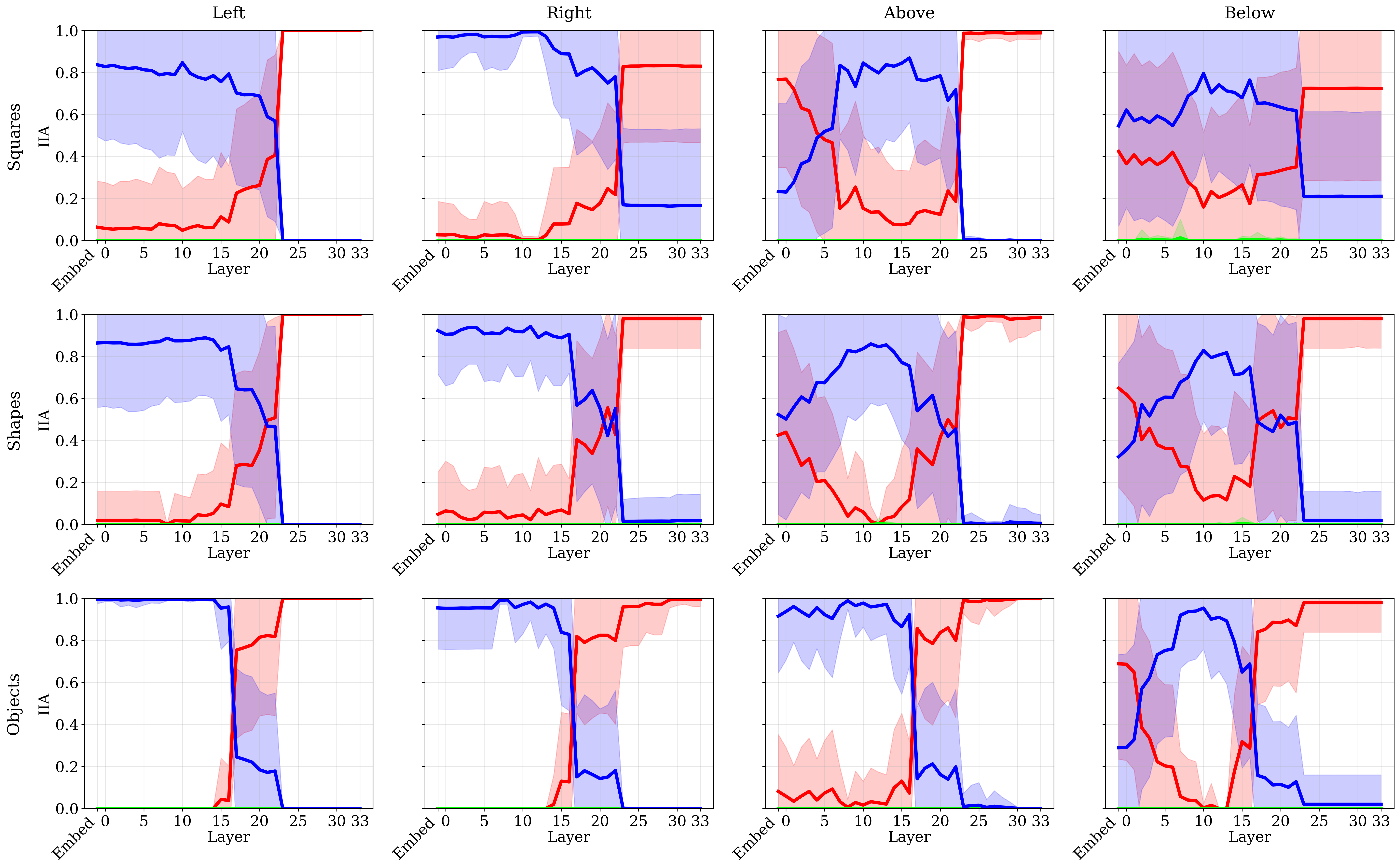}
    \caption{Interchange interventions applied only to visual tokens corresponding to
the strip in \texttt{Gemma-3-4b-it} model, with real background images.}
    \label{fig:gemma_criss_cross_strips_real_backgrounds}
\end{figure}

Figures~\ref{fig:probe_qwen_square_horizonatal_real_background}–\ref{fig:probe_gemma_objects_vertical_real_background} present the probing results, indicating that ordering information is spread across background tokens even when using real background images, rather than a white background. Additionally, the results in figures \ref{fig:qwen_criss_cross_squares_real_backgrounds}-\ref{fig:gemma_criss_cross_strips_real_backgrounds} confirm that patching the strip is required to alter the ordering information generated by the vision encoder.

\end{document}